%% file: main.tex
\pdfoutput=1
\documentclass[11pt]{article}

\usepackage[final]{acl}

\usepackage{times}
\usepackage{latexsym}

\usepackage[T1]{fontenc}
\usepackage[utf8]{inputenc}

\usepackage{microtype}

\usepackage{inconsolata}

\usepackage{graphicx}
\usepackage{booktabs}
\usepackage{multirow}
\usepackage{colortbl}
\usepackage{amssymb}
\usepackage{pifont}% http://ctan.org/pkg/pifont
\newcommand{\cmark}{\ding{51}}%
\newcommand{\xmark}{\ding{55}}%

\usepackage{fontawesome}
\usepackage{soul}

\newif\ifcomments
\commentstrue
\ifcomments
    \newcommand{\todo}[1]{{\color{red} todo: #1}}
    \newcommand{\safi}[1]{{\color{teal} Safi: #1}}
    \newcommand{\sd}[1]{{\color{blue} SD: #1}}
    \newcommand{\mk}[1]{{\color{green} MK: #1}}
\else
    \newcommand{\todo}[1]{}
    \newcommand{\safi}[1]{}
    \newcommand{\sd}[1]{}
    \newcommand{\mk}[1]{}
\fi

\input{defs}

\title{Finding Blind Spots in Evaluator LLMs with Interpretable Checklists}

\author{
 \textbf{Sumanth Doddapaneni\thanks{Equal Contribution.}\textsuperscript{1,2}} \quad
 \textbf{Mohammed Safi Ur Rahman Khan\footnotemark[1]\textsuperscript{1,2}}
\\
 \textbf{Sshubam Verma\textsuperscript{1}} \quad
 \textbf{Mitesh M. Khapra\textsuperscript{1,2}}
\\
\\
 \textsuperscript{1}Nilekani Centre at AI4Bharat \quad
 \textsuperscript{2}Indian Institute of Technology, Madras
\\
 \small{
   \textbf{Correspondence:} \texttt{\{sumanthd, miteshk\}@cse.iitm.ac.in, safikhan@ai4bharat.org}
 }
 \\
 \includegraphics[height=1.1em]{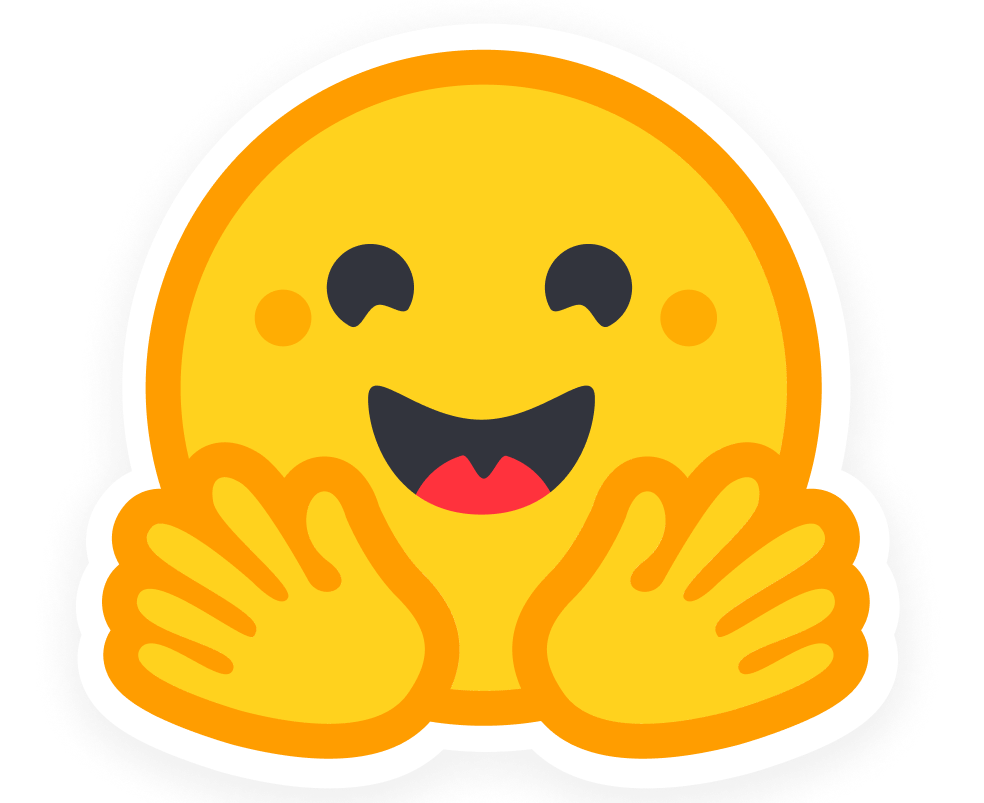}~\small{\url{https://huggingface.co/datasets/ai4bharat/FBI}}
 \\
 \includegraphics[height=1em]{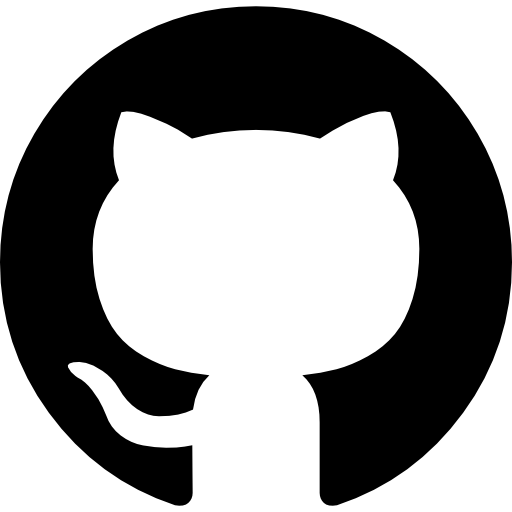}~\small{\url{https://github.com/AI4Bharat/FBI}}
}

\begin{document}
\maketitle

\input{sections/abstarct}
\input{sections/introduction}
\input{sections/related}
\input{sections/method}
\input{sections/experiments}

\input{sections/results}
\input{sections/conclusion}

\input{sections/limitations}
\input{sections/ethics}

\input{sections/acknowledgements}

\bibliography{main}

\appendix

% \section{Example Appendix}
% \label{sec:appendix}

% This is an appendix.
\input{sections/appendix}
\input{sections/full_tables}

\end{document}

%% file: defs.tex
\newcommand{\gpt}{\textsc{GPT-4-turbo}}
\newcommand{\gemini}{\textsc{Gemini-1.5-Pro}}
\newcommand{\llama}{\textsc{Llama-3-70B-Instruct}}
\newcommand{\claude}{\textsc{Claude-3-Opus}}
\newcommand{\prometheus}{\textsc{Prometheus 2}}
\newcommand{\oldg}{\textsc{GPT-3.5-turbo}}

\newcommand{\fbi}{\textsc{FBI}}

\definecolor{mygreen}{HTML}{d9ead3}
\definecolor{myorange}{HTML}{fce5cd}
\definecolor{myred}{HTML}{f4cccc}
\definecolor{mymagenta}{HTML}{ead1dc}
\definecolor{myblue}{HTML}{cfe2f3}
\definecolor{mygray}{HTML}{efefef}

\definecolor{gpt4}{HTML}{ea9999}
\definecolor{gemini}{HTML}{a2c4c9}
\definecolor{llama3}{HTML}{f9cb9c}

\definecolor{pert_green}{HTML}{02a61b}
\definecolor{pert_red}{HTML}{cc0000}

\newcommand{\total}{\textsc{\textbf{Total}}}

\newcommand{\contextual}{\textsc{Contextual}}
\newcommand{\entity}{\textsc{Entity}}
\newcommand{\incorrect}{\textsc{Incorrect Fact}}
\newcommand{\numerr}{\textsc{Number Errors}}
\newcommand{\opposite}{\textsc{Opposite Fact}}
\newcommand{\remove}{\textsc{Remove Fact}}

\newcommand{\domore}{\textsc{Do More}}
\newcommand{\doless}{\textsc{Do Less}}
\newcommand{\format}{\textsc{Ignore Format}}
\newcommand{\sequence}{\textsc{Sequence Errors}}
\newcommand{\assumptions}{\textsc{Assumptions}}

\newcommand{\calculation}{\textsc{Calculations}}
\newcommand{\copying}{\textsc{Copying Numbers}}
\newcommand{\final}{\textsc{Final Errors}}
\newcommand{\units}{\textsc{Incorrect Units}}

\newcommand{\formula}{\textsc{Wrong Formula}}

\newcommand{\grammar}{\textsc{Grammar}}
\newcommand{\spelling}{\textsc{Spelling}}
\newcommand{\consistency}{\textsc{Consistency}}
\newcommand{\chronology}{\textsc{Chronology}}
\newcommand{\coherence}{\textsc{Coherence}}

\newcommand{\comprehensive}{\textsc{Comprehensiveness}}

\newcommand{\scoreinv}{\textsc{Score Invariant}}

%% file: sections/abstarct.tex
\begin{abstract}

Large Language Models (LLMs) are increasingly relied upon to evaluate text outputs of other LLMs, thereby influencing leaderboards and development decisions. However, concerns persist over the accuracy of these assessments and the potential for misleading conclusions. In this work, we investigate the effectiveness of LLMs as evaluators for text generation tasks. We propose \fbi{}, a novel framework designed to examine the proficiency of Evaluator LLMs in assessing four critical abilities in other LLMs: factual accuracy, instruction following, coherence in long-form writing, and reasoning proficiency. By introducing targeted perturbations in answers generated by LLMs, that clearly impact one of these key capabilities, we test whether an Evaluator LLM can detect these quality drops. By creating a total of 2400 perturbed answers covering 22 perturbation categories, we conduct a comprehensive study using different evaluation strategies on five prominent LLMs commonly used as evaluators in the literature. Our findings reveal significant shortcomings in current Evaluator LLMs, which failed to identify quality drops in over 50\% of cases on average. Single-answer and pairwise evaluations demonstrated notable limitations, whereas reference-based evaluations showed comparatively better performance. \textit{These results underscore the unreliable nature of current Evaluator LLMs and advocate for cautious implementation in practical applications.} Code and data are available at \url{https://github.com/AI4Bharat/FBI}.
\end{abstract}

%% file: sections/introduction.tex
\section{Introduction}

\begin{figure}
    \centering
    \includegraphics[width=\columnwidth]{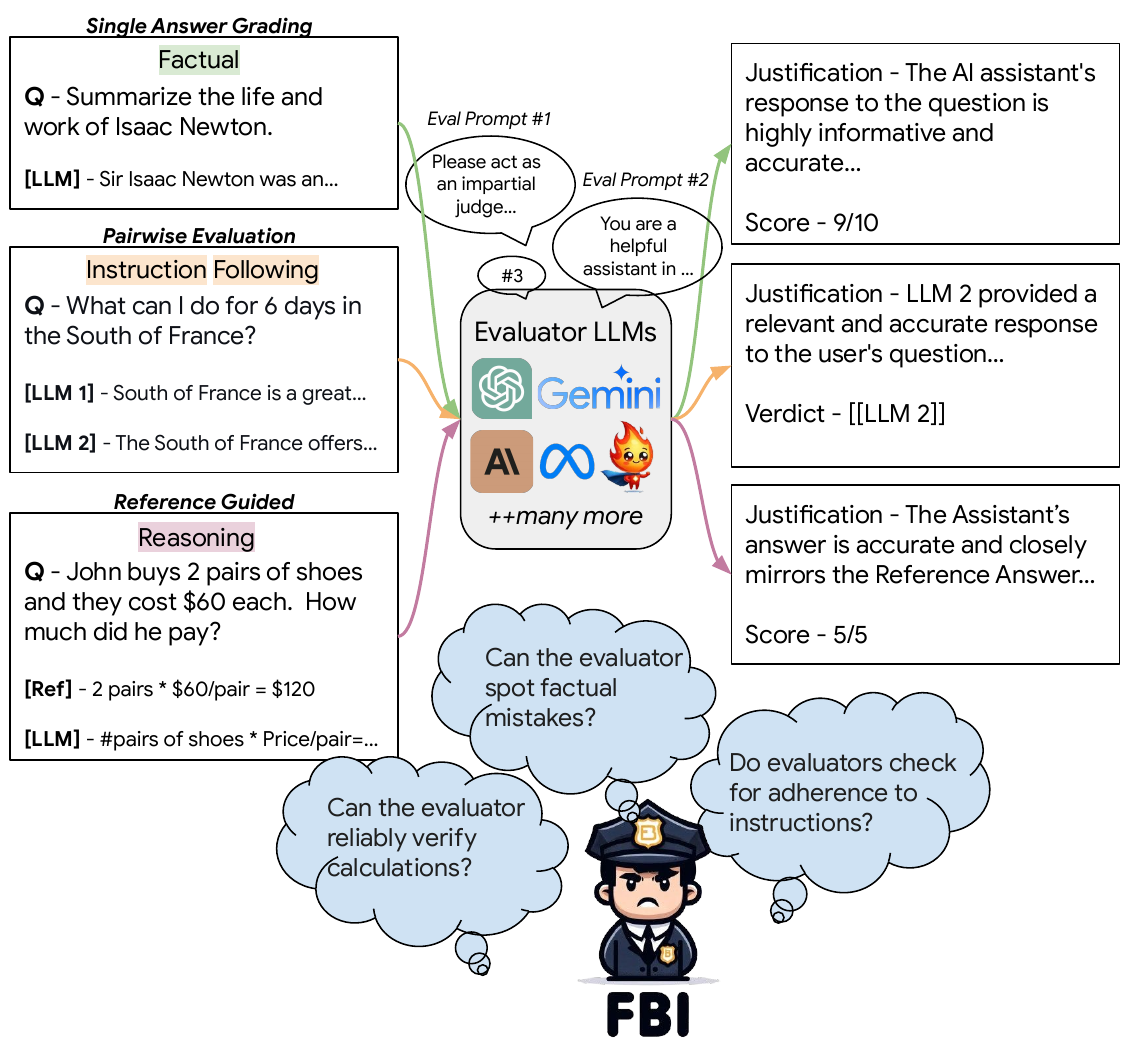}
    \caption{We present \fbi, our novel meta-evaluation framework designed to assess the robustness of evaluator LLMs across diverse tasks and evaluation strategies.}
    \label{fig:motivation}
\end{figure}

%\todo{LLM Evaluator $\rightarrow$ Evaluator LLM}
Large Language Models (LLMs) are gaining widespread acceptance as the gold standard for evaluation in numerous applications, thanks to their efficiency and significant reductions in cost \& time compared to human evaluators~\cite{prometheus, Kim2024TheBB, DBLP:conf/acl/ChiangL23, DBLP:journals/corr/abs-2304-00723, AlpacaFarm}. Furthermore, Evaluator LLMs are increasingly being utilized in the creation and maintenance of leaderboards for benchmarking various AI models~\cite{watts2024pariksha, llm-judge}. While this reliance on LLMs offers significant advantages, it also presents potential drawbacks that warrant careful consideration. If LLMs are not effective evaluators, the resulting rankings and assessments could be fundamentally flawed, leading to inaccurate conclusions and misguided decisions. Therefore, it is crucial to pause and rigorously assess the evaluation capabilities of LLMs.% to ensure the accuracy and reliability of their assessments. 

Recent studies have explored the effectiveness of LLMs as evaluators and have reported strong correlations with human evaluations~\cite{AlpacaFarm, llm-judge}. While these findings are promising, accepting LLMs as reliable evaluators necessitates more nuanced assessments~\cite{llm-bar}. 
As LLMs become integral in a diverse range of tasks, they are expected to demonstrate a wide array of abilities, including factual accuracy, instruction following, coherence in long-form writing, and reasoning proficiency.
Consequently, it is crucial to determine if Evaluator LLMs can indeed do a fine grained assessment of these varied abilities. 
Specifically, can they evaluate factual correctness, grammar, spelling, mathematical proficiency, and adherence to instructions in answers generated by other LLMs? (ref. Fig.~\ref{fig:motivation})
%More specifically, while evaluating answers generated by other LLMs, can Evaluator LLMs accurately evaluate factual correctness, adherence to grammar and spelling rules, proficiency in handling mathematical calculations and formulas, and competence in following complex instructions, among other finer aspects of evaluation. 
The necessity for such thorough fine-grained assessments is underscored by the Checklist~\cite{ribeiro-etal-2020-beyond} approach, initially applied to BERT~\cite{devlin-etal-2019-bert} and subsequently adapted in studies across various tasks and models~\cite{sai-etal-2021-perturbation}.

In this work, we introduce \textbf{\fbi{}}, a comprehensive framework designed to \textbf{F}ind \textbf{B}lind spots in evaluator LLMs  using an \textbf{I}nterpretable checklist across four fundamental text generation abilities: (a) factual accuracy, (b) instruction following, (c) coherence in long-form writing, and (d) reasoning proficiency. To rigorously assess an Evaluator LLM's ability to grade answers along these dimensions, we introduce perturbations that degrade the quality of the answer in one of these areas, expecting that good Evaluator LLMs will detect these quality drops and adjust their scores accordingly. Additionally, we develop quality-preserving perturbations where an Evaluator LLM should maintain consistent scoring. A detailed description of the 22 perturbation categories that we used is provided in Table~\ref{tab:perturbations}. Starting with 500 prompts, we first generate long-form responses using \gpt. We then use a human-in-the-loop approach, to systematically perturb these responses, resulting in a dataset of 2400 tuples, where each tuple contains a prompt, response, and perturbed response. %, \todo{resulting in ...}. %Crucially, our framework is designed for adaptability, allowing for its extension to assess Evaluator LLM's performance across a broader spectrum of abilities in the future.

Using the generated perturbations, we employed three evaluation paradigms (a) single-answer evaluation, (b) pairwise evaluation, and (c) reference-guided evaluation. Within each paradigm, we try multiple popular strategies of using Evaluator LLMs, such as, providing a rubric, asking for a justification, specifying the axis of evaluation, etc. Using these strategies, we assess the evaluation capabilities of five widely-used Evaluator LLMs. \textit{Our findings indicate that LLMs are currently far from being reliable evaluators for text generation tasks}. Even with the best models and evaluation strategies, Evaluator LLMs failed to identify errors in over 50\% of cases, on average. Interestingly, across all evaluation strategies, we observed that all popular Evaluator LLMs consistently performed poorly. Notably, even basic perturbation categories, such as, fluency perturbations (e.g. spellings and grammar) posed challenges for the evaluators. We also observed cases where Evaluator LLMs did not adjust their scores for perturbed responses despite correctly identifying the perturbations in their explanations. When used for single-answer grading and pairwise evaluation, Evaluator LLMs showed significant limitations, suggesting they are not reliable in these setups. In contrast, when used for reference-based evaluation, they demonstrated relatively better performance. Overall, our experiments uncovered significant blind spots in Evaluator LLMs, warranting caution in their direct application in practical settings.

%% file: sections/related.tex
\section{Related Work}
\label{sec:related}

% \subsection{Tasks} 
% Traditionally, models have been evaluated on classification tasks and single-sentence generation tasks such as translation and summarization, where inputs have dedicated ground truths and evaluations are conducted using task-specific metrics like accuracy, F1, BLEU~\cite{papineni-etal-2002-bleu}, ROUGE~\cite{lin-2004-rouge}, and chrF~\cite{popovic-2015-chrf}. However, these metrics fall short in capturing the semantic meaning beyond token-level accuracy. To address this, model-based evaluations such as COMET~\cite{comet}, BERTScore~\cite{bertscore} and BARTScore~\cite{bartscore} were introduced. In the current landscape of LLM generations, where benchmarks often feature open-ended questions, it has become increasingly challenging to use existing metrics, leading to a reliance on using other powerful LLMs as evaluators~\cite{llm-judge, AlpacaFarm}.

\noindent\textbf{LLMs as Evaluators. }
% LLMs have been increasingly used for automated evaluation for various NLG tasks~\cite{llm-judge, gpt-good-evaluator, llms-alt-humans}, given their capability to understand and generate open-ended text. The primary strategies for this include: (i) single answer grading~\cite{}, (ii) pairwise comparison \cite{}, and (iii) reference-guided evaluation~\cite{}. While these methods were initially applied using proprietary LLM APIs, open-source initiatives have focused on training evaluation-specific models. For instance, PandaLM~\cite{PandaLM} used pairwise comparison data to train a preference-based evaluation LLM, while Prometheus~\cite{prometheus} utilized reference answers and rubrics to develop a reference-guided evaluation system. More recent approaches to auto-eval is multi-agent debate/agreement, where multiple LLMs evaluate responses and interact to select the most suitable answer among the presented options~\cite{chateval}.
LLMs have been increasingly used for automated evaluation for various NLG tasks~\cite{gpt-good-evaluator, llms-alt-humans, gemba}. We broadly classify this into two paradigms - (i) reference-driven evaluations~\cite{fu2023gptscore, prometheus}, and (ii) reference-free evaluations~\cite{liu2023geval, llm-judge}. The evaluator is either asked for a score (score-based evaluation)~\cite{liu2023geval, llm-judge, hada2023large} or to choose the best amongst two given responses (pairwise comparison evaluation)~\cite{llm-judge, faireval, liusie2023llm}. Additionally, various open-source evaluation-specific trained models have also been proposed~\cite{PandaLM, prometheus, zhu2023judgelm}. Further, advanced ensemble approaches include evaluation via multi-agent interactions~\cite{chateval, wider-and-deeper} or with external agents~\cite{factscore, hasanbeig2023allure}.

% \subsection{Problems in AutoEval}
\noindent\textbf{Biases in Evalautor LLMs. }
% Issues with automated evaluation have been highlighted, such as position bias, self-enhancement, verbosity bias, and inadequate mathematical capability~\cite{llm-judge}. They suggested remedies including (i) swapping positions, (ii) few-shot judging, (iii) chain-of-thought reasoning, and (iv) training a judge model. Further emphasizing positional bias, \cite{llm-not-fair-eval} recommended calibration to mitigate this issue. Other research suggests incorporating rubrics during evaluation, providing the LLM with specific axes and rules to guide the evaluation process \cite{DBLP:journals/corr/abs-2310-19740, llm-bar, bsm}. \todo{add more works}
Studies around Evaluator LLMs have highlighted the various biases - position bias~\cite{llm-judge, llm-not-fair-eval}, self preference bias~\cite{panickssery2024llm, liu2023geval}, verbosity bias~\cite{wu2023style, llm-bar}, etc. Various approaches, including chain-of-thought reasoning~\cite{llm-judge, llm-bar}, position-swapping~\cite{llm-bar}, among others, have been suggested to mitigate some of these. Recent studies~\cite{hada2023large, bsm} also show the effectiveness of the evaluators can be increased by evaluating specific axes and providing detailed rubrics/rules~\cite{DBLP:journals/corr/abs-2307-10928, Kim2024TheBB}.

% \subsection{MetaEval of LLM}
% Critiquing evaluation metrics has a long history in NLP, with prior works focusing on text classification, machine translation, summarization, dialogue generation, and more. Current research tests the efficacy of LLMs as evaluators by comparing their performance to that of human evaluators, assessing agreement levels~\cite{DBLP:conf/acl/ChiangL23}. 
\noindent\textbf{Evaluation of Evaluator LLMs.}
Critically analysing evaluation metrics and suggesting methods to improve their robustness has always been of interest to the NLP community~\cite{indicmteval, mathur-etal-2020-tangled}. Recent studies have evaluated the efficacy of LLMs as evaluators for specific types of tasks~\cite{hada2024metal, shen-etal-2023-large} and evaluation paradigms~\cite{faireval, gpt-good-evaluator} by assessing their agreement with human evaluations~\cite{hada2023large,DBLP:conf/acl/ChiangL23,llm-judge}. Additionally, the robustness of these evaluators has been tested using adversarial examples~\cite{he-etal-2023-blind, kamoi2024evaluating, chen2024unveiling, wu2023style}, further showing their strengths and weaknesses.

Our proposed framework represents a significant departure from these existing approaches %~\cite{llm-bar, wu2023style, llm-judge, kamoi2024evaluating, chen2024unveiling}
in several key aspects. First, we focus on a broader set of essential abilities: factual understanding, instruction following, long-form writing, and reasoning. %This breadth ensures a robust evaluation across diverse cognitive domains. 
Second, all prompts and the 2400 perturbed answers in our framework are carefully crafted and/or validated by humans, ensuring high quality and relevance to the abilities being evaluated. Third, our framework offers finer granularity in perturbation types, allowing us to finely identify and isolate the capabilities and limitations of Evaluator LLMs. This detailed analysis assists in making more knowledgeable choices about when to utilize LLMs as evaluators. Lastly, we focus on three popular evaluation paradigms, viz., reference-less single answer scoring, reference-less pairwise comparison, and reference based scoring, thereby providing a comprehensive toolkit for evaluating LLM performance across different dimensions.

%% file: sections/method.tex
\section{\fbi: Meta-Evaluation Checklist}
\label{sec:fbi}

We introduce \fbi, a meta-evaluation benchmark designed to assess the capabilities of Evaluator LLMs in examining the outputs of other LLMs across four distinct task abilities: (i) \textit{Factual} accuracy, (ii) \textit{Reasoning} ability, (iii) \textit{instruction} following, and (iv) proficiency in \textit{long-form} writing. Each instance within the benchmark comprises a tuple ($I$, $A_{gold}$, $A_{perturb}$), where $I$ represents the input instruction or prompt given to the model, $A_{gold}$ denotes the correct or \textit{gold} answer, and $A_{perturb}$ signifies a \textit{perturbed} version of the gold answer.
% The perturbed answers, $A_{perturb}$, are generated by introducing specific types of errors across each of the four task abilities. The primary objective is to evaluate whether LLM evaluators can accurately identify and account for these errors while evaluating the perturbed answers. This benchmark aims to provide a rigorous framework for testing the robustness and reliability of Evaluator LLMs in discerning subtle and complex errors across diverse types of tasks.
% \fbi{} encompasses four task abilities, with perturbation axes meticulously crafted by humans, informed by the failure modes prevalent in current LLMs. Human oversight is pivotal throughout the benchmark's development, from prompt selection and task categorization to the definition of perturbation axes and the creation of the perturbations themselves. To ensure the highest standards of accuracy and reliability, all perturbations within \fbi~ undergo rigorous manual vetting. The statistical overview of \fbi~ is provided in Table~\ref{tab:fbi_stats}, and the detailed generation process is discussed in the following sections.
The perturbed answers, $A_{perturb}$, are generated by introducing specific types of errors across each of the four task abilities (Table~\ref{tab:perturbations}) to evaluate whether LLM evaluators can accurately identify and account for these errors in the perturbed answers.

%\fbi{} provides a rigorous framework for testing the robustness and reliability of Evaluator LLMs in discerning subtle and complex errors in the outputs of other LLMs. 
The perturbations are based on perturbation categories carefully crafted by human annotators, informed by the prevalent failure modes in current LLMs~\cite{factscore, wu2023reasoning, ifeval}. These human annotators are graduate students who are well aware of the typical errors made by LLMs. Such human oversight is used throughout the benchmark's development, from prompt selection ($\S$~\ref{subsec:prompt_selection}) to defining perturbation categories ($\S$~\ref{subsec:perturb_categories}) and creating the perturbations ($\S$~\ref{subsec:perturb_gen}). To ensure a high standard of accuracy and reliability, all perturbations within \fbi{} undergo rigorous manual vetting($\S$~\ref{subsec:human_checks}). %The statistical overview of \fbi{} is provided in 
Table~\ref{tab:fbi_stats} presents some statistics about \fbi, and the detailed generation process is discussed in the following sub-sections.

\begin{table}[t]
\centering
\small
\begin{tabular}{lc}
\toprule
\textbf{Category} & \textbf{\# Instances} \\
\midrule

\rowcolor{myblue}
Long Form (LF) &  \textbf{528} \\
\quad\quad \grammar &  92\\
\quad\quad \spelling &  100\\
\quad\quad \consistency &  84\\
\quad\quad \chronology &  71\\
\quad\quad \coherence &  91\\
\quad\quad \comprehensive &  90\\
\midrule

\rowcolor{mygreen}
Factual (F) & \textbf{483} \\
\quad\quad \contextual & 94 \\
\quad\quad \entity & 87 \\
\quad\quad \incorrect & 68 \\
\quad\quad \numerr & 74 \\
\quad\quad \opposite & 91 \\
\quad\quad \remove & 69 \\
\midrule
% \begin{tabular}[c]{@{}l@{}}Instruction\\ Following\end{tabular} & \\

\rowcolor{myorange}
Instruction Following (IF) & \textbf{379} \\
\quad\quad \domore & 50 \\
\quad\quad \doless & 100 \\
\quad\quad \format & 99 \\
\quad\quad \sequence & 49 \\
\quad\quad \assumptions & 81 \\
\midrule

\rowcolor{mymagenta}
Reasoning (R) & \textbf{494} \\
\quad\quad \calculation & 149 \\
\quad\quad \copying & 83 \\
\quad\quad \final & 97 \\
\quad\quad \units & 77 \\
% \quad\quad \bodmas & 57 \\
\quad\quad \formula & 88 \\
\midrule

\rowcolor{mygray}
Score Invariant (SI) & \textbf{516} \\

\midrule
Total & \textbf{2400} \\
\bottomrule
\end{tabular}
% \caption{Statistics of the created perturbations in \fbi. We list down the different types of perturbations created under each category along with the number of instances.}
\caption{Statistics of perturbations across all the 4 task abilities and each of the perturbation categories.}
\label{tab:fbi_stats}
\end{table}

\subsection{Prompt Selection}
\label{subsec:prompt_selection}
We selected six test sets containing prompts in English, viz., WizardLM~\cite{wizardlm}, MT Bench~\cite{llm-judge}, UltraChat~\cite{ultrachat}, LIMA~\cite{lima}, LLMBar~\cite{llm-bar}, and IFEval~\cite{ifeval}. 
These test sets were selected for their recency and because they contain prompts for long-form generation, creativity, and open-ended tasks that require instruction-following.
%These test sets were selected based on criteria such as recency, long-form generation, creativity, instruction-following nature, and open-endedness to form the base of our benchmark. 
Collectively, these test sets comprise of 1809 prompts. We manually categorized each prompt into one of the 4 task categories:

% \begin{enumerate}
%     \item \colorbox{myblue}{Factual}: These prompts seek objective information or facts. For example, \textit{What is the primary function of a capacitor in an electrical circuit?}
%     \item \colorbox{mygreen}{Instruction Following}: These prompts require executing specific steps or guidelines to achieve a particular outcome or answer. For example, \textit{Write a poem with \textbf{four} lines and the following words: peace, sky, race, ground.}
%     \item \colorbox{myorange}{Reasoning}: These prompts necessitate the application of logic, mathematics, and critical thinking to analyze information and draw conclusions. For example, \textit{A bat and a ball together cost \$1.10. The bat costs \$1.00 more than the ball. How much does the ball cost?}
%     \item \colorbox{mymagenta}{Long Form Writing}: These prompts require crafting extended pieces of text that explore generic topics, often including detailed analysis and storytelling. For example, \textit{How can I improve my time management skills?}
% \end{enumerate}

\setlength\fboxsep{1pt}
\noindent \colorbox{myblue}{Long Form Writing (LF):}~These prompts require generating long pieces of text and explore generic topics, often including detailed analysis and storytelling. For example, \textit{How can I improve my time management skills?}

\noindent \colorbox{mygreen}{Factual (F):}~These prompts seek objective information or facts. For example, \textit{What is the primary function of a capacitor in an electrical circuit?}

\noindent \colorbox{myorange}{Instruction Following (IF):}~These prompts require executing specific steps or guidelines to achieve a particular outcome or answer. For example, \textit{Write a poem with \textbf{four} lines and the following words: peace, sky, race, ground.}

\noindent \colorbox{mymagenta}{Reasoning (R):}~These prompts necessitate the application of logic, mathematics, and critical thinking to analyze information and draw conclusions. For example, \textit{A bat and a ball together cost \$1.10. The bat costs \$1.00 more than the ball. How much does the ball cost?}

We sampled 100 questions from each of the four abilities, supplementing prompts requiring reasoning ability from the GSM8k~\cite{gsm8k} and MATH~\cite{math} benchmarks. Additionally, we created 200 prompts tailored to instruction following to address specific perturbation categories\footnote{Based on our categorization, we were unable to find a sufficient number of prompts in existing test sets to fit the perturbation categories.}. The gold answers ($A_{gold}$) for all prompts were generated using the \gpt{} model. To ensure the quality and accuracy of $A_{gold}$, we conducted manual verification by randomly sampling 25\% instances from each category and found that the gold answers maintain a high level of correctness. \textit{Importantly, we emphasize that the quality of gold answers is not critical in our study, as our primary focus is on directional score changes} (i.e., we are interested in knowing if a perturbed answer with clear errors scores \textit{relatively lower} than the original answer which did not have these errors).

\setlength\fboxsep{3pt}
\begin{table*}[t!]
\centering
\small
\begin{tabular}{cll}
\toprule
Task & Perturbation Axis & Description \\
\midrule
\multirow{7}{*}{\colorbox{myblue}{LF}} & \grammar & Introducing grammatical errors in the answer. Eg: \textcolor{pert_green}{This is good} $\rightarrow$ \textcolor{pert_red}{This are good}. \\
 & \spelling & Introducing ``valid'' spelling errors in the answer. Eg: \textcolor{pert_green}{Toxicity} $\rightarrow$ \textcolor{pert_red}{Tocixity}. \\
 & \consistency &  Introducing  errors in the ``consistency'' of the answer (like tone, terminology, etc.) \\
  & \chronology & Introducing errors in the chronological or the logical flow of the answer.    \\
 & \coherence & Introducing errors that affect the coherence of the answer.   \\
 & \comprehensive & Introducing vagueness, irrelevance or lack of context in the answer. \\
\midrule
\multirow{6}{*}{\colorbox{mygreen}{F}} & \contextual & Replacing fact with a contextually similar incorrect fact. Eg: \textcolor{pert_green}{electricity} $\rightarrow$ \textcolor{pert_red}{magnetism}.   \\
 & \entity & Replacing a named entity with an incorrect entity. Eg: \textcolor{pert_green}{Poland} $\rightarrow$ \textcolor{pert_red}{London}.\\
 & \incorrect & Adding a new contextually relevant incorrect fact in the answer.\\
  & \numerr & Introducing errors in the various numbers reported in the answer. Eg: \textcolor{pert_green}{1987} $\rightarrow$ \textcolor{pert_red}{1887}.   \\
 & \opposite & Replacing a fact in the answer with its negation. Eg: \textcolor{pert_green}{... will have ...} $\rightarrow$ \textcolor{pert_red}{... wont have ...}.  \\
 & \remove & Removing a fact critical to the correctness and completeness of the answer.\\
 \midrule
% \begin{tabular}[c]{@{}l@{}}Instruction\\ Following\end{tabular}
\multirow{5}{*}{\colorbox{myorange}{IF}} & \doless & Doing less than what is \textit{explicitly} requested in the question.\\
 & \domore & Doing more than what is \textit{explicitly} requested in the question.   \\
 & \format & Ignoring the formatting and other constraints mentioned in the question.   \\
  & \sequence & Ignoring the sequence in the response when \textit{explicitly} requested in the instruction. \\
 & \assumptions & Making new incorrect assumptions about the instruction.\\
\midrule
\multirow{5}{*}{\colorbox{mymagenta}{R}} & \calculation & Introducing  calculation errors in the answer. Eg: \textbf{\textcolor{pert_green}{$2+3=5$} $\rightarrow$ \textcolor{pert_red}{$2+3=6$}}   \\
 & \copying & Introducing errors while considering the numbers mentioned in the instruction.   \\
 & \final & Introducing errors only the final reported answer while retaining the correct solution.  \\
  & \units & Introducing errors in the units reported and considered in the answer.  \\
 & \formula & Introducing errors in the formula used in the answer. Eg: \textcolor{pert_green}{$\pi r^{2}$} $\rightarrow$ \textcolor{pert_red}{$2\pi r$} \\
 \midrule
 \colorbox{mygray}{SI} & \scoreinv & Introducing modifications in the answer which would not result in a score penalty.\\
\bottomrule
\end{tabular}
\caption{Perturbation categories across each of the task abilities. The \textcolor{pert_green}{green} highlights indicate the original text and the \textcolor{pert_red}{red} highlights indicated the perturbed text. Complete examples of each perturbation can be found in supplementary material.}
\label{tab:perturbations}
\end{table*}

\subsection{Perturbation Categories}
\label{subsec:perturb_categories}
LLMs exhibit numerous failure modes, encompassing shortcomings in reasoning~\cite{wu2023reasoning, Wei2022ChainOT}, factuality~\cite{hu2024towards, factscore}, instruction-following~\cite{ifeval, li2023evaluating}, and, in some instances, coherence and consistency~\cite{naismith-etal-2023-automated, shen-etal-2023-large} in generated text. Given that we utilize Evaluator LLMs to assess responses in one or more of these abilities, it is imperative for the evaluator to excel in them. Our perturbations across each task ability are crafted keeping these failure modes in mind, as presented in Table~\ref{tab:perturbations}. While our perturbations are primarily designed to decrease scores, we also develop score-invariant perturbations ($\S$~\ref{subsec:score_inv}), which are intended not to affect the score relative to the gold answer.%which should not affect the score compared to the gold answer.

\subsection{Perturbation Generation}
\label{subsec:perturb_gen}
To generate perturbed answers ($A_{perturb}$) along each of the defined categories ($\S$~\ref{subsec:perturb_categories}), we use \gpt{} by prompting it with specific instructions tailored to each perturbation category. The model was tasked with producing perturbed answers and explaining the reasoning behind each perturbation. We iteratively refined the instructions by manually reviewing a sample of 25\% of perturbed answers for each category, till we were satisfied with the generated perturbations. %The prompts used for each perturbation axis can be found in \todo{Appendix}.

\subsection{Human-In-The-Loop}
\label{subsec:human_checks}
While GPT generally succeeds in generating the expected perturbations, we observed instances where the model (i) deviates from the intended perturbation, (ii) produces the incorrect style of perturbation, or (iii) accurately generates the reasoning but fails to reflect it in $A_{perturb}$. To address these inconsistencies, we meticulously vet all generated perturbations through a manual review process. Each perturbed answer produced by \gpt{} is examined against $A_{gold}$, and then categorized as valid, invalid, or score invariant. A perturbation is considered valid only if it should logically result in a scoring penalty as determined by human annotators. The vetting is carried out by students who possess a comprehensive understanding of LLM literature, holding at least a bachelor's or master's degree. To aid in  validating perturbations, we developed a tool, the details of which are outlined in Appendix~\ref{apx:perturb_ui}.

\subsection{Score-Invariant Perturbations}
\label{subsec:score_inv}
Score-invariant perturbations are those modifications that do not warrant a scoring penalty. These are collected in two ways: (i) human annotators categorize specific instances from our initial list as invariant ($\S$~\ref{subsec:human_checks}), and (ii) prompting \gemini{} model to paraphrase $A_{gold}$ ensuring retention of all original facts and details followed by human verification on a sample. We collect 516 score invariant perturbations in total.

%% file: sections/experiments.tex
\section{Strategies for using Evaluator LLMs}
\label{sec:eval_setup}

In this section, we outline the prompting strategies employed by Evaluator LLMs benchmarked on \fbi{}. 
An Evaluator LLM, $f(\cdot)$, takes the input instruction, LLM generated response and an evaluation prompt, $P_{eval}$, as input, and is required to generate a score and an optional explanation.
% An Evaluator LLM, $f(\cdot)$, takes the original instruction and a response generated by another LLM as input. The Evaluator LLM is also given an evaluation prompt, $P_{eval}$, and is required to generate a score and an optional explanation. 
To make the evaluation more robust, the evaluator may also be provided with additional information specifying the axes of evaluation, rubrics, rules, and other criteria. Our study focuses on 3 evaluation paradigms: (i) Single-answer scoring ($\S$\ref{subsec:single}), (ii) Pairwise comparison ($\S$\ref{subsec:pairwise}), and (iii) Reference-guided evaluation ($\S$\ref{subsec:reference}). For all the strategies evaluation prompts $P_{eval}$ are adapted from~\citet{llm-judge, llm-bar, hada2023large}.

% An Evaluator LLM consists of a robust LLM ($M$) coupled with an evaluation prompt $P_{eval}$, denoted as $f(\cdot)$. The evaluator can be supplemented with additional axis, rubric, rules, etc. to further refine the evaluation robustness.

% In this section, we outline our evaluation methodologies. Fundamentally, an LLM evaluator is a combination of a strong LLM ($M$), and a prompting strategy ($P_{eval}$)~\cite{llm-bar}. We denote this combination with the function $f(\cdot)$. We consider three evaluation paradigms in our study: (i) reference-less single-answer scoring, (ii) reference-less pairwise comparison, and (iii) reference-based scoring evaluation. Details of each paradigm, along with various evaluation strategies is discussed below. By default, we consider the Chain-of-Thought (CoT) prompting~\cite{Wei2022ChainOT} for all, unless mentioned otherwise.

\subsection{Single Answer Scoring}
\label{subsec:single}
In this paradigm, evaluator $f(\cdot)$ is tasked with scoring a model response based solely on its parameterized knowledge.

\paragraph{Vanilla$^*$}\cite{llm-judge}: In this strategy, the evaluator $f(\cdot)$ is presented with only the input instruction $I$ and a model response $A_{model}$. The role of $f(\cdot)$ is to evaluate $A_{model}$ and assign a score, denoted as $f(P_{eval}, I, A_{model}) \rightarrow (score)$. 
% We adopt the prompt, $P_{eval}$, given to the Evalauator LLM from \cite{llm-judge}.

\paragraph{Vanilla}\cite{llm-judge}: This strategy extends ``Vanilla$^*$'', where the evaluator $f(\cdot)$ is tasked not only with scoring the model response $A_{model}$ but also providing an explanation for the score - represented as $f(P_{eval}, I, A_{model}) \rightarrow (exp, score)$. 
% We again adopt the prompt, $P_{eval}$, recommended by \citet{llm-judge}.

\paragraph{Rubric}\cite{llm-bar}: In this strategy, in addition to the instruction $I$ and the model response $A_{model}$, we also provide a grading rubric $R$. The evaluator $f(\cdot)$ is prompted to first generate an explanation followed by a score - represented as $f(P_{eval}, R, I, A_{model}) \rightarrow (exp, score)$. 
% For this, we adapt and extend the prompt, $P_{eval}$, recommended by \citet{llm-bar}.

\paragraph{Axis}\cite{hada2023large}: In this strategy, the evaluator $f(\cdot)$ is prompted to assess the model response, $A_{model}$, along a designated axis, $Ax$, aligning with the category of the instruction ($\S$~\ref{subsec:prompt_selection}). For instance, factual questions are evaluated along the $hallucination$ axis to determine the presence of fabricated content. This process is formally represented as $f(P_{eval}, Ax, I, A_{model}) \rightarrow (exp, score)$. 
% We extend and adapt the axes and prompts recommended by \citet{hada2023large}.

\paragraph{Axis+Rubric}\cite{hada2023large}: In this strategy, the evaluator $f(\cdot)$ is provided with both a specific evaluation axis $Ax$ and detailed scoring rubrics $R$ for that axis. The is formally represented as $f(P_{eval}, Ax, R, I, A_{model}) \rightarrow (exp, score)$. 
% We again extend the axes and rubrics recommended by \citet{hada2023large}.

\subsection{Pairwise Comparison}
\label{subsec:pairwise}
In this paradigm, evaluator $f(\cdot)$ is tasked to choose the better response from the two given options by again relying on its parameterized knowledge. 
% Below, we briefly describe various evaluation strategies considered under this paradigm.

\paragraph{Pairwise$^*$}\cite{llm-judge}: The evaluator $f(\cdot)$ here is given only an instruction $I$ and two model responses $A_1$ and $A_2$ and is tasked to determine the better response or mark both as equally valid. This is formally represented as $f(P_{eval}, I, A_1, A_2) \rightarrow (verdict)$. 
% We adopt the prompt, $P_{eval}$,  recommended by \citet{llm-judge}.

\paragraph{Pairwise}\cite{llm-judge}: This strategy extends ``Pairwise$^*$'', where the evaluator is tasked not only with choosing the better response but also providing an explanation for the verdict - represented as $f(P_{eval}, I, A_1, A_2) \rightarrow (exp, verdict)$. 
% We again adopt the prompt, $P_{eval}$,   recommended by \citet{llm-judge}.

\paragraph{Rules}\cite{llm-bar}: In this strategy, in addition to the instruction $I$ and the two model responses $A_1$, $A_2$, the evaluator $f(\cdot)$ is given detailed rules for evaluation and is asked to generate an explanation followed by the verdict. This process is formally represented as $f(P_{eval}, R, I, A_1, A_2) \rightarrow (exp, verdict)$. 
% We extend the prompt, $P_{eval}$,   recommended by \citet{llm-bar}.

\paragraph{Axis}\cite{hada2023large}: Extending the Axis strategy defined in Sec $\S$\ref{subsec:single}, the evaluator $f(\cdot)$ is asked to choose the better response along a designated axis $Ax$. The evaluator is prompted with the instruction $I$, two model responses $A_1$, $A_2$, and the description of the axis $Ax$ - represented as $f(P_{eval}, Ax, R, I, A_1, A_2) \rightarrow (exp, verdict)$. 
% We adapt and extend the prompt, $P_{eval}$,   recommended by \citet{hada2023large}.

\paragraph{Axis+Rules}\cite{llm-bar, hada2023large}: Extending the Axis+Rubric strategy defined in Sec $\S$\ref{subsec:single}, this strategy involves choosing the better response along the designated axis $Ax$. The evaluator is prompted with the instruction $I$, two model responses $A_1$, $A_2$, details about the axis $Ax$, and detailed rules for evaluation - represented as $f(P_{eval}, Ax, R, I, A_1, A_2) \rightarrow (exp, verdict)$. 
% We adapt and extend the prompts and axes recommended by \citet{llm-bar} and \citet{hada2023large}.

% \noindent \paragraph{ChatEval:} Multiple LLM evaluators participate in a collaborative multi-agentic debate to autonomously discuss and evaluate the quality of model responses. Each evaluator takes turns to provide their final response based on the context of the preceding discussion. This process is described as $f(M_{1\cdots N}, I, A_1, A_2) \rightarrow (exp, score)$.

% \noindent \paragraph{PandaLM:} While all evaluations are conducted via APIs without access to model weights, this approach utilizes a trained LLM Evaluator to obtain preference ratings. Specifically, PandaLM, which is trained on top of the LLaMa model, is used for this purpose. This process is described as $f(M_{pandaLM}, I, A_1, A_2) \rightarrow (exp, score)$.

\subsection{Reference-guided Single Answer Scoring}
\label{subsec:reference}
In this paradigm, the evaluator $f(\cdot)$ is tasked to score a response by comparing against a reference. \textit{It is important to note that this approach may not be feasible for many open-ended questions}.

% \noindent \textbf{Reference:} Given an instruction $I$, a model response $A_{model}$, and a ground truth answer $A_{gold}$, the LLM Evaluator $M$ is tasked with grading the model response on a scale of 1 to 10, providing an accompanying explanation. This process is represented as $f(M, I, A_{gold}, A_{model}) \rightarrow (exp, score)$.

\paragraph{Reference}\cite{llm-judge}: In this strategy, given an instruction $I$, a model response $A_{model}$, and a ground truth reference answer $A_{gold}$, the evaluator $f(\cdot)$ is tasked with scoring the model response, along with giving an explanation. This is formally represented as $f(P_{eval}, I, A_{gold}, A_{model}) \rightarrow (exp, score)$. 
% We adopt the prompt suggested by \citet{llm-judge}.

% \noindent \paragraph{Prometheus:} Here Prometheus, adapted from Llama-2, is trained to evaluate model responses when paired with a reference answer. The LLM Evaluator is provided with an instruction $I$, a model response $A_{model}$, a reference answer $A_{gold}$, and a customized rubric $R$. The evaluator is tasked with generating a score and an explanation. This process is represented as $f(M_{prometheus}, I, A_{gold}, R, A_{model}) \rightarrow (exp, score)$.

% \noindent \textbf{Prometheus}~\cite{kim2024prometheus2}: In this strategy, we use Prometheus2, a trained LLM response evaluator. The model expects an instruction $I$, a model response $A_{model}$, a reference answer $A_{gold}$, and a custom scoring rubric $R$ and generates an explanation and a score - represented as $f_{prom}(P_{eval},  A_{gold}, R, I, A_{model}) \rightarrow (exp, score)$. %Prometheus2 is also capable of performing reference-based pairwise evaluations, which we leave for future work.

%% file: sections/results.tex
% \section{Results \& Discussion}
% \subsection{Experimental Setup}
\section{Experiments}

We use \gpt{ \raisebox{-0.15em}{\includegraphics[height=1em]{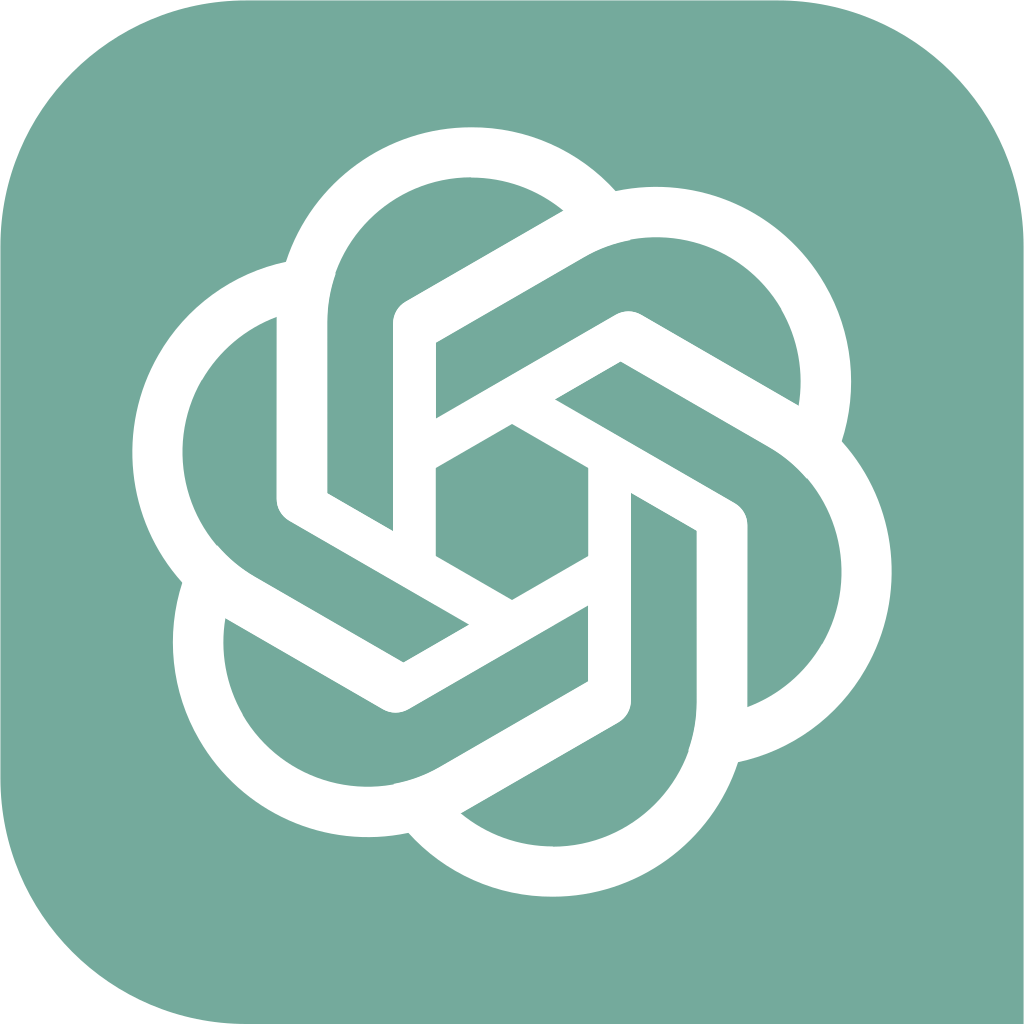}}}  as our primary evaluation model, given its widespread adoption~\cite{llm-bar, hada2024metal, factscore}. We also extend our analysis to other proprietary models - \gemini{ \raisebox{-0.3em}{\includegraphics[height=1.2em]{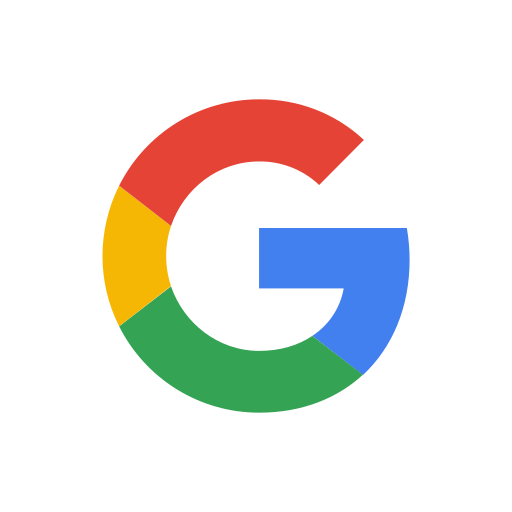}}} ~\cite{team2024gemini} and \claude{ \raisebox{-0.15em}{\includegraphics[height=0.9em]{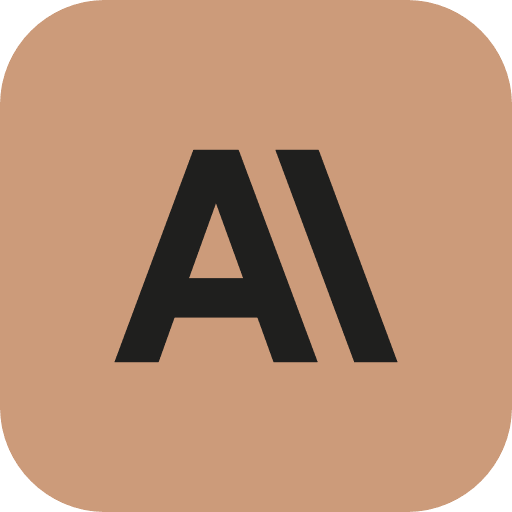}}} ~\cite{claude}, open-source models like \llama{\raisebox{-0.4em}{\includegraphics[height=1.4em]{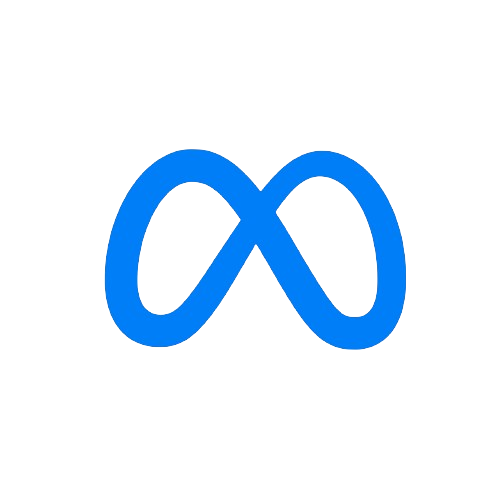}}} ~\cite{llama3}, and trained evaluator models like \prometheus{\raisebox{-0.4em}{\includegraphics[height=1.4em]{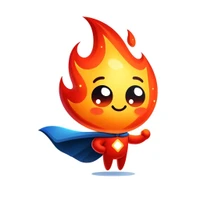}}} ~\cite{kim2024prometheus2}\footnote{~We reuse the axes and rubrics defined in Section $\S$\ref{subsec:single} as the evaluation rubrics for \prometheus. }. All evaluations are conducted at a temperature of zero to ensure reproducibility.

\noindent In single answer scoring ($\S$~\ref{subsec:single}) paradigm, we measure the percentage of instances where the score remains unchanged by the perturbation as our metric. Ideally, except for score-invariant perturbations, the evaluator should penalize the score of the perturbed answer. For pairwise comparison paradigm ($\S$~\ref{subsec:pairwise}), we include our ``gold'' answer as one of the responses, requiring the evaluator to select the best response between the ``gold'' and the ``perturbed'' answer.  Here, we measure the percentage of times the evaluator does not choose the gold answer as our metric. To mitigate position bias~\cite{llm-not-fair-eval}, we conduct each evaluation twice, swapping the order of the gold and perturbed responses. 

\noindent For reference-guided single answer scoring paradigm ($\S$~\ref{subsec:reference}), the gold answer serves as the reference. Here, we measure the percentage of times the evaluator awards a perfect score to the perturbed answer as our metric.

% \subsection{Models}
% We use \gpt~as our primary evaluation model, given its widespread adoption in contemporary research~\cite{llm-bar, hada2024metal, factscore}. To strengthen our findings, we extend our analysis to other proprietary models, such as \gemini~\cite{team2024gemini} and \claude~\cite{claude}, open-source models like \llama~\cite{llama3}, and trained evaluator models like \prometheus~\cite{kim2024prometheus2}. To ensure reproducibility, we conduct all our experiments with a temperature of zero.

\begin{table}[]
\centering
\small
\begin{tabular}{lccccc}
\toprule
\textbf{Strategy} & \colorbox{myblue}{\textbf{LF}$\downarrow$} & \colorbox{mygreen}{\textbf{F}$\downarrow$} & \colorbox{myorange}{\textbf{IF}$\downarrow$} & \colorbox{mymagenta}{\textbf{R}$\downarrow$} & \colorbox{mygray}{\textbf{SI}$\uparrow$}\\
\midrule

\multicolumn{6}{c}{\textit{Single Answer Scoring}} \\
Vanilla$^*$ & 0.73 & 0.67 & 0.71 &  \textbf{0.22} & 0.83\\
Vanilla & \textbf{0.57} & \textbf{0.54} & \textbf{0.57} & 0.25 & 0.71\\
Rubric & 0.85 & 0.73 & 0.80 & 0.33 & 0.96\\
Axis & 0.83 & 0.74 & 0.75 & 0.43 & 0.96\\
Axis+Rubric & 0.86 & 0.76 & 0.77 & 0.37 & \textbf{0.97}\\
\midrule

\multicolumn{6}{c}{\textit{Pairwise Comparison}} \\
Pairwise$^*$ & 0.73 & 0.52 & 0.83 & 0.36 & \textbf{0.93}\\
Pairwise & 0.77 & 0.46 & 0.67 & 0.35 & 0.74\\
Rules & 0.75 & 0.63 & 0.68 & 0.41 & 0.74\\
Axis & 0.64 & 0.44  & \textbf{0.59} & \textbf{0.27} & 0.71\\
Axis+Rules & \textbf{0.64} & \textbf{0.42} & 0.61 & 0.32 & 0.72\\
%ChatEval &  &  &  &  & \textbf{-}\\
% PandaLM &  &  &  &  \\
\midrule

\multicolumn{6}{c}{\textit{Reference-guided Single Answer Scoring}} \\
Reference & 0.26 & 0.11 & 0.49 & 0.04 & 0.63\\
% Prometheus &  &  &  &  & \\
% \quad \textit{---generic} & 0.41 & 0.46 & 0.46 & 0.18 & 0.41\\
% \quad \textit{---specific} & 0.51 & 0.62 & 0.53 & 0.12 & 0.38\\
\bottomrule
\end{tabular}
\caption{Comparison of different evaluation strategies using \gpt{}. The numbers indicate the percentage of instances where the score/verdict generated by the LLM evaluator is \textbf{not affected} by the perturbation. Lower values ($\downarrow$) indicate better performance in all categories except \colorbox{mygray}{SI}. \textbf{*} denotes evaluators that only give a score without any justification.}
\label{tab:main_results}
\end{table}

\subsection{Is GPT-4-Turbo a good evaluator?}
% \noindent \textbf{Is GPT-4-Turbo a good evaluator?}\quad
Referring to the first section of Table \ref{tab:main_results}, we observe that in the case of single answer scoring, \gpt{} fails to lower its score for the perturbed answer in a majority of the cases, except for Reasoning tasks. Further, the performance of \gpt{} is better when using simpler strategies, such as, Vanilla$^*$ and Vanilla, as compared to the more advanced strategies with explicit rubrics and/or specified axis of evaluation. This could imply that while adding additional rubrics and criteria may increase the overall thoroughness, it may not necessarily enhance the model's ability to detect subtler errors. 

%As shown by the results in Table \ref{tab:main_results}, \gpt~ fails to detect perturbations in as high as demonstrates varying performance levels across various categories and evaluation paradigms. In the reference-less scoring paradigm, simpler strategies including Vanilla$*$ and Vanilla show strong performance across all categories. Surprisingly, the advanced strategies show a decline in performance, with consistently higher scores. This could imply that while adding additional rubrics and criteria may increase the overall thoroughness, it may not necessarily enhance the model's ability to detect subtler errors.

Now, referring to the second section of Table \ref{tab:main_results}, we observe that in the case of pairwise comparison, \gpt{} fails to detect the perturbed answer in majority of the cases, except for Reasoning tasks. Further, in contrast to the above, in this case, advanced strategies perform better than the basic strategies. This indicates that for comparative evaluations, having detailed specific rules can help improve the reliability of the models. Lastly, referring to the first row of the last section of Table \ref{tab:main_results}, we observe that when a reference is provided, \gpt{} performs much better but there are still a notable number of failures. The evaluator, despite being presented with the gold answer marked as a reference answer, fails to recognize the perturbations in many cases, except for reasoning tasks where it performs very well. % Note that this test is conducted in a very ideal scenario wherein the difference between the reference and the perturbed a%Ideally, our setting is akin to a text-diff operation, and while it performs effectively within the reasoning category, it shows limitations across others. 
\textit{Our overall verdict is that \gpt{} is not a good evaluator as it fails to detect perturbations which cause a drop in the quality of the answer.}

\begin{table}[t!]
\setlength{\tabcolsep}{4pt}
\renewcommand{\arraystretch}{1}
\centering
\small
% \resizebox{\textwidth}{!}{
\begin{tabular}{lcccccc}
\toprule
\textbf{Strategy} & \textbf{Model} & \colorbox{myblue}{\textbf{LF$\downarrow$}} & \colorbox{mygreen}{\textbf{F$\downarrow$}} & \colorbox{myorange}{\textbf{IF$\downarrow$}} & \colorbox{mymagenta}{\textbf{R$\downarrow$}} & \colorbox{mygray}{\textbf{SI$\uparrow$}}\\
\midrule

\multirow{4}{*}{Vanilla} & \includegraphics[width=0.3cm]{figures/chatgpt.png} & \textbf{0.57} & \textbf{0.54} & 0.57 & \textbf{0.25} & 0.71 \\
 & \includegraphics[width=0.4cm]{figures/google.png} & 0.61 & 0.73 & \textbf{0.54} & 0.41 & 0.71\\
 & \includegraphics[width=0.3cm]{figures/anthropic.png} & 0.74 & 0.84 & 0.75 & 0.47 & \textbf{-}\\
 & \includegraphics[width=0.4cm]{figures/meta.png} & 0.86 & 0.95 & 0.90 & 0.71 & \textbf{0.75}\\
 \midrule

\multirow{3}{*}{Axis+Rules} & \includegraphics[width=0.3cm]{figures/chatgpt.png} & \textbf{0.64} & \textbf{0.42} & \textbf{0.61} & \textbf{0.32} & \textbf{0.72}\\
 & \includegraphics[width=0.4cm]{figures/google.png} & 0.72 & 0.58 & 0.70 & 0.39 & 0.65\\
 % & Claude &  &  &  &  \\
 & \includegraphics[width=0.4cm]{figures/meta.png} & 0.75 & 0.69 & 0.70 & 0.60 & 0.64\\
\midrule

\multirow{4}{*}{Reference} & \includegraphics[width=0.3cm]{figures/chatgpt.png} & 0.26 & 0.11 & 0.49 & 0.04 & \textbf{0.63}\\
 & \includegraphics[width=0.4cm]{figures/google.png} & 0.25 & 0.07 & 0.17 & 0.03 & 0.33\\
 % & Claude &  &  &  &  \\
 & \includegraphics[width=0.4cm]{figures/meta.png} & \textbf{0.03} & \textbf{0.01} & \textbf{0.05} & \textbf{0.05} & 0.13\\
 & \includegraphics[width=0.4cm]{figures/prometheus.png} & 0.51 & 0.62 & 0.53 & 0.12 & 0.38\\
 
 \bottomrule
\end{tabular}
% }
\caption{Comparison of the performance of different models across the best-observed evaluation strategies. Lower values ($\downarrow$) indicate better performance in all categories except \colorbox{mygray}{SI}.}
\label{tab:other_models}
\end{table}

\subsection{How do other popular Evaluator LLMs perform?}
% \noindent \textbf{How do other popular Evaluator LLMs perform?} \quad
We extend our evaluation to other models and compare their performance when using the 3 best strategies identified in Table \ref{tab:main_results}. Table \ref{tab:other_models} shows that \gpt{} consistently outperforms other models in both the reference-less paradigms. Due to the high API cost of using the \claude{} model, we restrict its evaluation to only the Vanilla strategy, and note that it performed poorly as an Evaluator LLM.

In the reference-based paradigm, \llama{} model surprisingly outperforms all others. Upon manually reviewing few instances, we observe that \llama{} is a stringent evaluator and rarely awards perfect scores to even very well-formed answers when presented with a reference answer. While this may suggest that \llama{} has a high evaluation standard, it also raises concerns about overlyrelying on the reference answer, which is typically not available in most practical scenarios. To further investigate this, we evaluate all the models on Score Invariant perturbations (Section $\S$\ref{subsec:score_inv}) using the Reference evaluation strategy. Consistent with our prior observations, \llama{} seldom awards perfect scores, doing so only in 13\% of the cases as shown in Table \ref{tab:other_models}. Lastly, looking at the last row of Table \ref{tab:other_models}, we observe that even trained Evaluator LLMs like \prometheus{} are worse than other general Evaluator LLMs.

% \begin{table}[t!]
% \centering
% \small
% % \resizebox{\textwidth}{!}{
% \begin{tabular}{llc}
% \toprule
% \textbf{Strategy} & \textbf{Model} & \colorbox{mygray}{\textbf{SI}} \\
% \midrule

% \multirow{3}{*}{Reference} & GPT4-T & \textbf{0.63}\\
%  & Gemini & 0.33 \\
%  & Llama3 & 0.13 \\
%  \bottomrule
% \end{tabular}
% % }
% \caption{Here we compare the performance of models across the different eval strategies}
% \end{table}

% \begin{table}[]
%     \centering
%     \small
%     \begin{tabular}{lccc}
%     \toprule
%         \textbf{Category} & \colorbox{gpt4}{GPT4-Turbo} & \colorbox{gemini}{Gemini}& \colorbox{llama3}{Llama3}\\
%         \midrule
%          \colorbox{mygray}{Score Invariant} & \textbf{0.63} & 0.33 & 0.13\\
%     \bottomrule
%     \end{tabular}
%     \caption{Evaluating models on the Score Invariant Perturbations using the Reference evaluator. Higher $(\uparrow)$ is better.}
%     \label{tab:inv_ref}
% \end{table}

\begin{figure}
    \centering
    \includegraphics[width=\columnwidth]{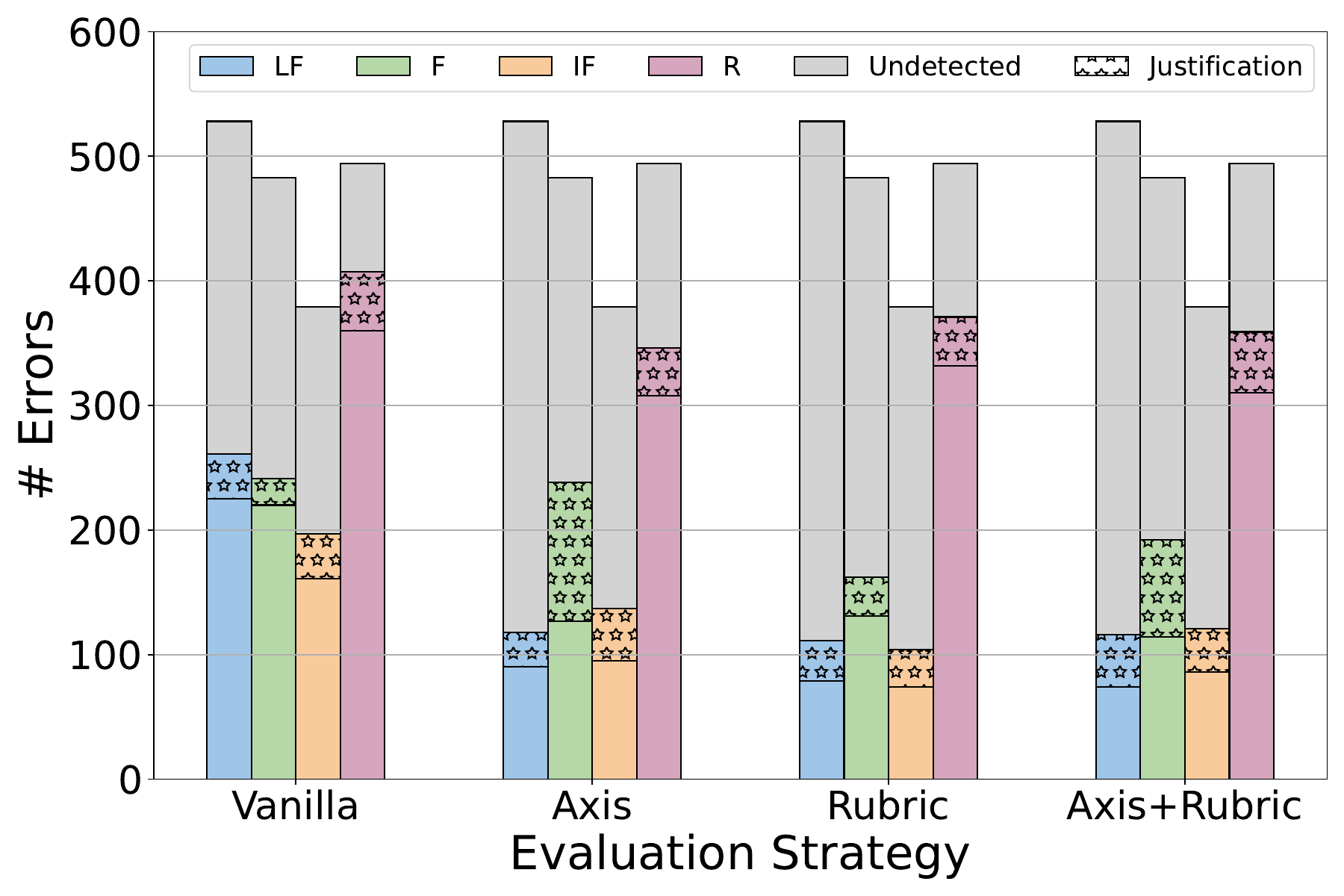}
    \caption{Comparison of perturbations detected solely by score analysis versus those identified with explanations. The highlighted region marked with stars denotes perturbations detected in explanations but not reflected in scores. Despite this, a significant proportion of perturbations remain undetected.}
    \label{fig:justification}
\end{figure}

\begingroup
\setlength{\tabcolsep}{4pt}
\begin{table}[t]
\small
\centering
\begin{tabular}{lcccccccc}
\toprule
\multirow{2}{*}{\textbf{}} & \multicolumn{2}{c}{\colorbox{myblue}{\textbf{LF$\downarrow$}}}& \multicolumn{2}{c}{\colorbox{mygreen}{\textbf{F$\downarrow$}}} & \multicolumn{2}{c}{\colorbox{myorange}{\textbf{IF$\downarrow$}}}& \multicolumn{2}{c}{\colorbox{mymagenta}{\textbf{R$\downarrow$}}}\\ 
\cmidrule(lr){2-3} \cmidrule(lr){4-5} \cmidrule(lr){6-7} \cmidrule(lr){8-9}

& 1-3 & \multicolumn{1}{c}{1-5} & 1-3 & \multicolumn{1}{c}{1-5} & 1-3 & \multicolumn{1}{c}{1-5} & 1-3 & \multicolumn{1}{c}{1-5} \\ 
\midrule

R  & \multicolumn{1}{r}{0.85} & \multicolumn{1}{r}{\textbf{0.76}} & \multicolumn{1}{r}{0.73} & \multicolumn{1}{r}{\textbf{0.69}} & \multicolumn{1}{r}{0.80}  & \multicolumn{1}{r}{\textbf{0.72}} & \multicolumn{1}{r}{0.33} & \textbf{0.30} \\

A+R  & \multicolumn{1}{r}{0.86} & \multicolumn{1}{r}{\textbf{0.73}} & \multicolumn{1}{r}{0.76} & \multicolumn{1}{r}{\textbf{0.74}} & \multicolumn{1}{r}{0.77} & \multicolumn{1}{r}{\textbf{0.74}} & \multicolumn{1}{r}{\textbf{0.37}} & 0.38 \\
\bottomrule
\end{tabular}
\caption{Comparing performance of Rubrics and Axis+Rubrics strategies with score range of 1-3 and 1-5. The numbers indicate the percentage of instances where the score generated by the LLM evaluator is not affected by the perturbation. Lower values $(\downarrow)$ indicate better performance in all categories.}
\label{tab:inc_range}
\end{table}
\endgroup

\subsection{Does it help to look beyond scores?} 
% \noindent \textbf{Does it help to look beyond scores?} \quad
In addition to scoring, our evaluators also generate explanations that provide a justification for each score. We investigate whether these explanations detect the perturbations, even though this is not reflected in the scores. We prompt \oldg~model with explanations from the instances where the evaluator rated the perturbed answer as equal to the gold answer, asking it to identify if any mistake or error has been reported in the explanation. Figure \ref{fig:justification} reveals that explanations are only marginally helpful. Although perturbations are sometimes identified, they are overlooked or not considered significant enough to penalize the score. It is important to note that all the perturbations here were intended to incur a scoring penalty. 
Thus, while explicitly considering the explanations offers a slight improvement in the evaluator's performance, the overall performance is still poor.

\subsection{What about score-invariant perturbations?}
% \noindent \textbf{What about score-invariant perturbations?} \quad
We evaluate different Evaluator LLMs using  score-invariant perturbations ($\S$~\ref{subsec:score_inv}). Ideally, the evaluator should not reduce its score for these perturbations in score-based evaluations and should deem both responses correct in pairwise evaluations. Referring to Table \ref{tab:main_results} 
%shows trends opposite to those observed in the other categories. 
, in reference-less scoring, \gpt{} performs better when using non-vanilla evaluating strategies, while in pairwise comparison, it performs better when using simpler evaluation strategies. Similarly, as shown in Table \ref{tab:other_models}, we observe that other Evaluator LLMs also perform well in a majority of cases. However, there is still a significant number of responses with score-invariant perturbations that they rate poorly. %This suggests that detailed evaluators are better at scoring good answers but struggle with scoring bad answers, while for comparative evaluations, the reverse is true.

\subsection{Does increasing the range help in scoring?}
% \noindent \textbf{Does increasing the range help in scoring?}\quad

Based on recommendations from \citet{hada2023large}, our initial set-up for the Rubrics and Axis+Rubrics evaluators used a scoring range of 1 to 3. To explore whether a wider scoring range could enhance the evaluators' ability to identify and account for the perturbations, we extended the range to 1 to 5. Results presented in Table~\ref{tab:inc_range} suggest that this broader range slightly improves the evaluators' performance, perhaps due to the availability of more flexibility in scoring decisions.

%% file: sections/conclusion.tex
\section{Conclusion}
We propose \fbi, a novel framework designed to evaluate the proficiency of Evaluator LLMs in assessing four critical abilities: factual accuracy, instruction adherence, coherence in long-form writing, and reasoning proficiency, through targeted perturbations. Our comprehensive study, involving 2400 perturbed answers across 22 categories and using three evaluation paradigms (single-answer, pairwise, and reference-guided evaluation), reveals significant shortcomings in current Evaluator LLMs. Our findings show that even the most advanced models failed to identify quality drops in over 50\% of cases on average. While reference-based evaluations performed relatively better, single-answer and pairwise evaluations demonstrated notable limitations. These results underscore the unreliable nature of current Evaluator LLMs and advocate for cautious implementation in practical applications. We hope that the \fbi{} framework will be further extended and used for continued meta-evaluation of Evaluator LLMs.

%We introduce \fbi, a comprehensive framework designed to evaluate Evaluator LLMs across four critical text generation abilities: factual accuracy, instruction following, coherence in long-form writing, and reasoning proficiency. Using a set of quality-degrading and quality-preserving perturbations, we examine the ability of Evaluator LLMs to correctly identify such perturbations. Our findings show that even state-of-the-art Evaluator LLMs fail to detect and correctly score errors in over 50\% of cases, with even popular and powerful LLMs like \gpt{} falling short of expectations.

%% file: sections/limitations.tex
\section*{Limitations}
In our evaluation setup, detailed in Section~\ref{sec:eval_setup}, we concentrate on three primary evaluation paradigms: single-answer assessment, pairwise comparison, and reference-guided evaluation within a single model context and leave out multi-agent meta-evaluation and for future work. 
While we have compiled a list of perturbation categories, we believe it is not exhaustive and there is room for further expansion. Our evaluation framework encompasses four fundamental task abilities, with plans to explore more advanced capabilities such as multilingual generation, tool usage, and planning in future work.

%% file: sections/ethics.tex
\section*{Ethics}

All annotations described in Section~\ref{sec:fbi} were done by students from our research group, all of whom hold at least a bachelor's or master's degree. This annotation was done as a part of their routine research work. The datasets used in this paper are all available under permissible licenses, and we adhere strictly to their intended usage, maintaining compliance with licensing requirements. Additionally, the code used for our evaluations and perturbation generation will be made publicly available under the MIT License\footnote{\url{https://opensource.org/licenses/MIT}}. We only used ChatGPT\footnote{\url{https://chatgpt.com}} for assistance purely with the language of the paper, e.g., paraphrasing, spell-checking, or polishing the author’s original content, without suggesting new content. %This does not have any bearing on our research and thus we did not mention it.

% and the \fbi{} dataset will be released under a CC-0 License\footnote{\url{https://creativecommons.org/share-your-work/public-domain/cc0/}}.

% \todo{add license of datasets}

%% file: sections/acknowledgements.tex
\section*{Acknowledgements}
We would like to thank EkStep Foundation and Nilekani Philanthropies for their generous grant, which supported this research. We extend our gratitude to Ananth, Devilal, Niharika, Nikhil, Sakshi, Sparsh, and Suhaas, Suriya for their invaluable assistance with manual audits. We also thank Raj Dabre and Anoop Kunchukuttan for their insightful discussions. We thank Google for supporting Sumanth's work through the Google Ph.D. Fellowship.

%% file: sections/appendix.tex
% \section*{Appendix}

\section{Manual Verication Process of the Perturbations}
\label{apx:perturb_ui}
We engaged 17 graduate student volunteers with a good understanding of Large Language Models to manually verify the perturbations. Each annotator was provided with the instruction, the original gold answer, and the \gpt~ generated perturbed answer. They were tasked with classifying each perturbation into one of five categories: (i) Valid Perturbation, (ii) Invalid Perturbation, (iii) Score Invariant Perturbation, (iv) Not Relevant, and (v) Not Sure. Additionally, annotators were given explanations of the expected perturbations and the reasons why \gpt~ considered them valid.

\begin{figure*}
    \includegraphics[width=16cm]{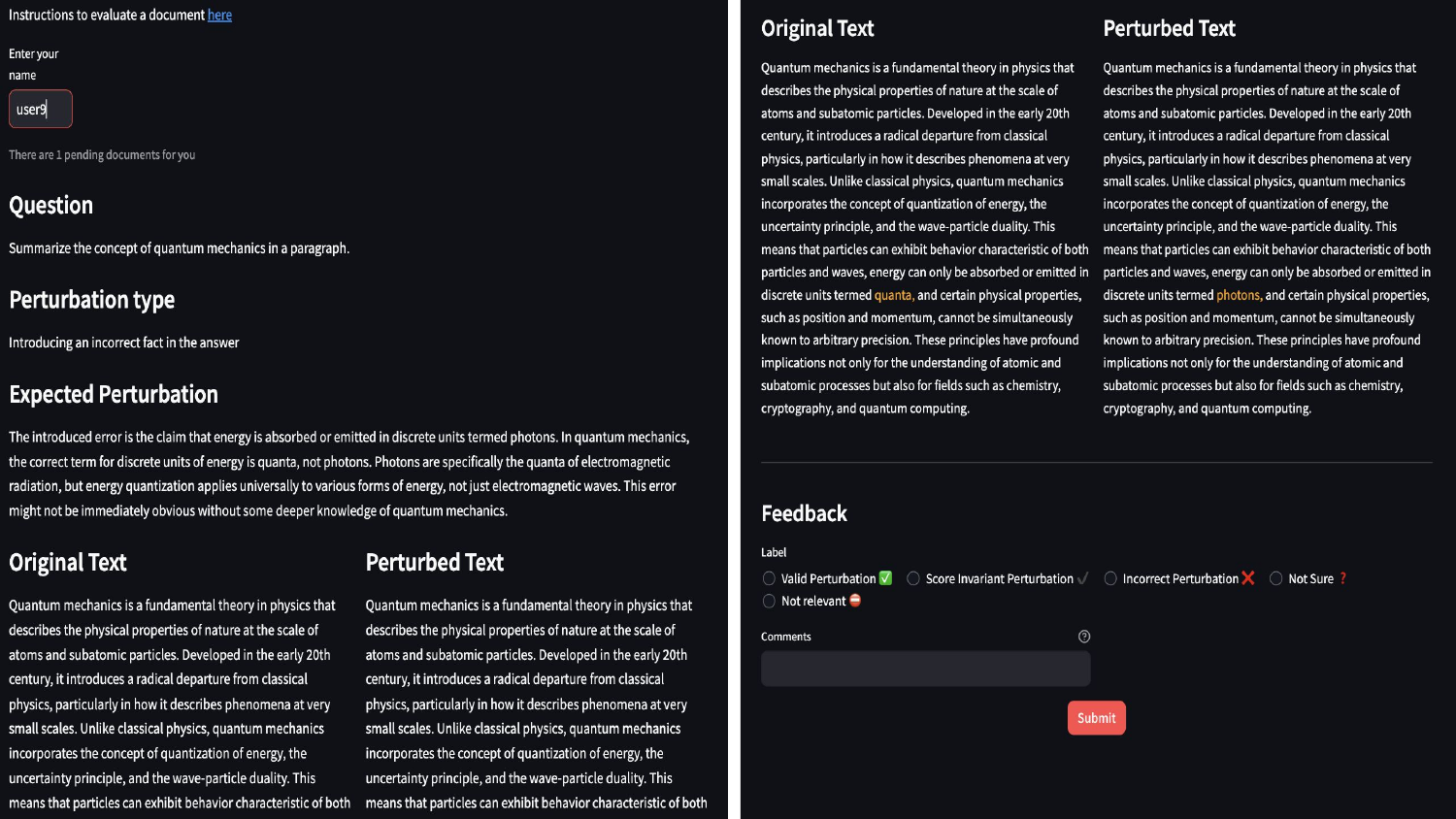}
    \caption{Screenshot of the User Application developed for validating perturbations.}
    \label{fig:UI}
\end{figure*}

To facilitate this process, we developed a straightforward application, the interface of which is depicted in Figure \ref{fig:UI}. This tool highlights the differences between the original and perturbed answers to aid easy identification.

Annotators were instructed to label an answer as ``Valid Perturbation'' only if they believed the perturbation warranted a score penalty relative to the gold answer. Perturbations not affecting the score were to be labeled ``Score Invariant''. If a perturbation was deemed incorrect or not reflected in the perturbed answer, annotators were asked to adjust the perturbation manually. Perturbations irrelevant to the category were to be marked as ``Not Relevant''.

\section{Detailed Results of Single Answer Evaluators}
\label{apx:detailed_single}
Detailed results of Single Answer evaluators can be found in Table \ref{tab:apx:single1}, \ref{tab:apx:single2}, \ref{tab:apx:single3}, \ref{tab:apx:single4}, \ref{tab:apx:single5}.

\section{Detailed Results of Pairwise Evaluators}
\label{apx:detailed_pair}
Detailed results of Pairwise Evaluators can be found in Table \ref{tab:apx:pair1}, \ref{tab:apx:pair2}, \ref{tab:apx:pair3}, \ref{tab:apx:pair4}, \ref{tab:apx:pair5}.

\section{Detailed Results of Reference-Guided Evaluators}
\label{apx:detailed_ref}
Detailed results of Reference-guided Evaluators can be found in Table \ref{tab:apx:ref1}, \ref{tab:apx:ref2}

%% file: sections/full_tables.tex
\begin{table*}[t]
\centering
\small
\begin{tabular}{clcccc}
\toprule
\multicolumn{1}{l}{} & \textbf{Perturbation Type} & \multicolumn{1}{c}{\textbf{\begin{tabular}[c]{@{}l@{}}\total\\ Errors\end{tabular}}} & \multicolumn{1}{c}{\textbf{\begin{tabular}[c]{@{}l@{}}Detected\\ Errors\end{tabular}}} & \multicolumn{1}{c}{\textbf{\begin{tabular}[c]{@{}l@{}}Undetected\\ Errors\end{tabular}}} & \multicolumn{1}{c}{\textbf{\begin{tabular}[c]{@{}c@{}}\% Undetected\\ Errors\end{tabular}}} \\
\midrule

\multirow{7}{*}{\colorbox{myblue}{LF}} & \coherence & 91 & 78 & 13 & 0.14 \\
 & \comprehensive & 90 & 9 & 82 & 0.91 \\
 & \consistency & 84 & 16 & 68 & 0.81 \\
 & \grammar & 92 & 25 & 67 & 0.73 \\
 & \chronology & 71 & 7 & 64 & 0.90 \\
 & \spelling & 100 & 11 & 89 & 0.89 \\
 \rowcolor{myblue}
 & \textbf{\total} & \textbf{528} & \textbf{146} & \textbf{383} & \textbf{0.73} \\
\midrule

\multirow{7}{*}{\colorbox{mygreen}{F}} & \contextual & 94 & 41 & 53 & 0.56 \\
 & \entity & 87 & 29 & 58 & 0.67 \\
 & \incorrect & 68 & 24 & 44 & 0.65 \\
 & \numerr s & 74 & 22 & 52 & 0.70 \\
 & \opposite & 91 & 39 & 52 & 0.57 \\
 & \remove & 69 & 4 & 65 & 0.94 \\
 \rowcolor{mygreen}
 & \textbf{\total} & \textbf{483} & \textbf{159} & \textbf{324} & \textbf{0.67} \\
\midrule

\multirow{6}{*}{\colorbox{myorange}{IF}} & \assumptions & 81 & 4 & 77 & 0.95 \\
 & \doless & 100 & 32 & 68 & 0.68 \\
 & \domore & 50 & 34 & 16 & 0.32 \\
 & \format & 99 & 36 & 63 & 0.64 \\
 & \sequence & 49 & 4 & 45 & 0.92 \\
\rowcolor{myorange}
 & \textbf{\total} & \textbf{379} & \textbf{110} & \textbf{269} & \textbf{0.71} \\
\midrule

\multirow{6}{*}{\colorbox{mymagenta}{R}} & \calculation & 149 & 121 & 28 & 0.19 \\
 & \copying & 83 & 69 & 14 & 0.17 \\
 & \final & 97 & 54 & 43 & 0.44 \\
 & \units & 77 & 66 & 11 & 0.14 \\
 & \formula & 88 & 73 & 15 & 0.17 \\
\rowcolor{mymagenta}
 & \textbf{\total} & \textbf{494} & \textbf{383} & \textbf{111} & \textbf{0.22} \\
\bottomrule
\end{tabular}
\caption{Results from evaluating \fbi~using \textbf{Vanilla}$^*$ evaluator. An error is said to be detected if the evaluator penalizes the score of the perturbed answer.}
\label{tab:apx:single1}
\end{table*}

% f(i, r) -> (explanation, score)
\begin{table*}[]
\centering
\small
\begin{tabular}{clcccc}
\toprule
\multicolumn{1}{l}{} & \textbf{Perturbation Type} & \multicolumn{1}{c}{\textbf{\begin{tabular}[c]{@{}l@{}}\total\\ Errors\end{tabular}}} & \multicolumn{1}{c}{\textbf{\begin{tabular}[c]{@{}l@{}}Detected\\ Errors\end{tabular}}} & \multicolumn{1}{c}{\textbf{\begin{tabular}[c]{@{}l@{}}Undetected\\ Errors\end{tabular}}} & \multicolumn{1}{c}{\textbf{\begin{tabular}[c]{@{}c@{}}\% Undetected\\ Errors\end{tabular}}} \\
\midrule

\multirow{7}{*}{\colorbox{myblue}{LF}} & \coherence & 91 & 82 & 9 & 0.10 \\
 & \comprehensive & 90 & 30 & 60 & 0.67 \\
 & \consistency & 84 & 35 & 49 & 0.58 \\
 & \grammar & 92 & 40 & 52 & 0.57 \\
 & \chronology & 71 & 18 & 53 & 0.75 \\
 & \spelling & 100 & 20 & 80 & 0.80 \\
 \rowcolor{myblue}
 & \total & \textbf{528} & \textbf{225} & \textbf{303} & \textbf{0.57} \\
\midrule

\multirow{7}{*}{\colorbox{mygreen}{F}} & \contextual & 94 & 45 & 48 & 0.51 \\
 & \entity & 87 & 43 & 44 & 0.51 \\
 & \incorrect & 68 & 29 & 38 & 0.56 \\
 & \numerr & 74 & 30 & 44 & 0.59 \\
 & \opposite & 91 & 48 & 42 & 0.46 \\
 & \remove & 69 & 25 & 44 & 0.64 \\
 \rowcolor{mygreen}
 & \total & \textbf{483} & \textbf{220} & \textbf{260} & \textbf{0.54} \\
\midrule

\multirow{6}{*}{\colorbox{myorange}{IF}} & \assumptions & 81 & 12 & 69 & 0.85 \\
 & \doless & 100 & 57 & 43 & 0.43 \\
 & \domore & 50 & 31 & 19 & 0.38 \\
 & \format & 99 & 41 & 57 & 0.58 \\
 & \sequence & 49 & 20 & 29 & 0.59 \\
 \rowcolor{myorange}
 & \total & \textbf{379} & \textbf{161} & \textbf{217} & \textbf{0.57} \\
\midrule

\multirow{6}{*}{\colorbox{mymagenta}{R}} & \calculation & 149 & 112 & 34 & 0.23 \\
 & \copying & 83 & 69 & 12 & 0.14 \\
 & \final & 97 & 53 & 43 & 0.44 \\
 & \units & 77 & 60 & 16 & 0.21 \\
 & \formula & 88 & 66 & 19 & 0.22 \\
 \rowcolor{mymagenta}
 & \total & \textbf{494} & \textbf{360} & \textbf{124} & \textbf{0.25} \\
\bottomrule
\end{tabular}
\caption{Results from evaluating \fbi~using \textbf{Vanilla} evaluator. An error is said to be detected if the evaluator penalizes the score of the perturbed answer.}
\label{tab:apx:single2}
\end{table*}

% f(i, r, rubrics) -> (explanation, score)					
\begin{table*}[]
\centering
\small
\begin{tabular}{clcccc}
\toprule
\multicolumn{1}{l}{} & \textbf{Perturbation Type} & \multicolumn{1}{c}{\textbf{\begin{tabular}[c]{@{}l@{}}\total\\ Errors\end{tabular}}} & \multicolumn{1}{c}{\textbf{\begin{tabular}[c]{@{}l@{}}Detected\\ Errors\end{tabular}}} & \multicolumn{1}{c}{\textbf{\begin{tabular}[c]{@{}l@{}}Undetected\\ Errors\end{tabular}}} & \multicolumn{1}{c}{\textbf{\begin{tabular}[c]{@{}c@{}}\% Undetected\\ Errors\end{tabular}}} \\
\midrule

\multirow{7}{*}{\colorbox{myblue}{LF}} & \coherence & 91 & 47 & 44 & 0.48 \\
 & \comprehensive & 90 & 2 & 88 & 0.98 \\
 & \consistency & 84 & 11 & 73 & 0.87 \\
 & \grammar & 92 & 15 & 77 & 0.84 \\
 & \chronology& 71 & 0 & 71 & 1.00 \\
 & \spelling & 100 & 4 & 96 & 0.96 \\
 \rowcolor{myblue}
 & \textbf{\total} & \textbf{528} & \textbf{79} & \textbf{449} & \textbf{0.85} \\
 \midrule
 
\multirow{7}{*}{\colorbox{mygreen}{F}} & \contextual & 94 & 34 & 60 & 0.64 \\
 & \entity & 87 & 29 & 58 & 0.67 \\
 & \incorrect& 68 & 18 & 50 & 0.74 \\
 & \numerr & 74 & 17 & 57 & 0.77 \\
 & \opposite & 91 & 32 & 59 & 0.65 \\
 & \remove & 69 & 1 & 68 & 0.99 \\
 \rowcolor{mygreen}
 & \textbf{\total} & \textbf{483} & \textbf{131} & \textbf{352} & \textbf{0.73} \\
 \midrule
 
\multirow{6}{*}{\colorbox{myorange}{IF}} & \assumptions & 81 & 1 & 80 & 0.99 \\
 & \doless & 100 & 8 & 92 & 0.92 \\
 & \domore & 50 & 39 & 11 & 0.22 \\
 & \format & 99 & 26 & 73 & 0.74 \\
 & \sequence & 49 & 0 & 49 & 1.00 \\
 \rowcolor{myorange}
 & \textbf{\total} & \textbf{379} & \textbf{74} & \textbf{305} & \textbf{0.80} \\
 \midrule
 
\multirow{6}{*}{\colorbox{mymagenta}{R}} & \calculation & 149 & 102 & 47 & 0.32 \\
 & \copying & 83 & 64 & 19 & 0.23 \\
 & \final & 97 & 49 & 48 & 0.49 \\
 & \units & 77 & 56 & 21 & 0.27 \\
 & \formula & 88 & 61 & 27 & 0.31 \\
 \rowcolor{mymagenta}
 & \textbf{\total} & \textbf{494} & \textbf{332} & \textbf{162} & \textbf{0.33} \\
 \bottomrule
\end{tabular}
\caption{Results from evaluating \fbi~using \textbf{Rubrics} evaluator. An error is said to be detected if the evaluator penalizes the score of the perturbed answer.}
\label{tab:apx:single3}
\end{table*}

% f(i, r, axes) -> (explanation, score)
\begin{table*}[]
\centering
\small
\begin{tabular}{clcccc}
\toprule
\multicolumn{1}{l}{} & \textbf{Perturbation Type} & \multicolumn{1}{c}{\textbf{\begin{tabular}[c]{@{}l@{}}\total\\ Errors\end{tabular}}} & \multicolumn{1}{c}{\textbf{\begin{tabular}[c]{@{}l@{}}Detected\\ Errors\end{tabular}}} & \multicolumn{1}{c}{\textbf{\begin{tabular}[c]{@{}l@{}}Undetected\\ Errors\end{tabular}}} & \multicolumn{1}{c}{\textbf{\begin{tabular}[c]{@{}c@{}}\% Undetected\\ Errors\end{tabular}}} \\
\midrule

\multirow{7}{*}{\colorbox{myblue}{LF}} & \coherence & 91 & 58 & 33 & 0.36 \\
 & \comprehensive & 90 & 1 & 89 & 0.99 \\
 & \consistency & 84 & 8 & 76 & 0.90 \\
 & \grammar & 92 & 17 & 75 & 0.82 \\
 & \chronology& 71 & 0 & 71 & 1.00 \\
 & \spelling & 100 & 6 & 94 & 0.94 \\
 \rowcolor{myblue}
 & \textbf{\total} & \textbf{528} & \textbf{90} & \textbf{438} & \textbf{0.83} \\
 \midrule
 
\multirow{7}{*}{\colorbox{mygreen}{F}} & \contextual & 94 & 29 & 65 & 0.69 \\
 & \entity & 87 & 30 & 57 & 0.66 \\
 & \incorrect& 68 & 17 & 51 & 0.75 \\
 & \numerr & 74 & 18 & 56 & 0.76 \\
 & \opposite & 91 & 32 & 59 & 0.65 \\
 & \remove & 69 & 1 & 68 & 0.99 \\
 \rowcolor{mygreen}
 & \textbf{\total} & \textbf{483} & \textbf{127} & \textbf{356} & \textbf{0.74} \\
 \midrule
 
\multirow{6}{*}{\colorbox{myorange}{IF}} & \assumptions & 81 & 5 & 76 & 0.94 \\
 & \doless & 100 & 20 & 80 & 0.80 \\
 & \domore & 50 & 40 & 10 & 0.20 \\
 & \format & 99 & 25 & 74 & 0.75 \\
 & \sequence & 49 & 5 & 44 & 0.90 \\
 \rowcolor{myorange}
 & \textbf{\total} & \textbf{379} & \textbf{95} & \textbf{284} & \textbf{0.75} \\
 \midrule
 
\multirow{7}{*}{\colorbox{mymagenta}{R}} & \calculation & 149 &  100 & 49 & 0.53 \\
 & \copying & 83 & 57 & 26 & 0.31 \\
 & \final & 97 & 46 & 51 & 0.53 \\
 & \units & 77 & 42 & 35 & 0.45 \\
 & \formula & 88 & 63 & 25 & 0.28 \\
 \rowcolor{mymagenta}
 & \textbf{\total} & \textbf{494} & \textbf{308} & \textbf{186} & \textbf{0.43} \\
 \bottomrule
\end{tabular}
\caption{Results from evaluating \fbi~using \textbf{Axis} evaluator. An error is said to be detected if the evaluator penalizes the score of the perturbed answer.}
\label{tab:apx:single4}
\end{table*}

% f(i, r, axes, rubrics) -> (explanation, score)
\begin{table*}[]
\centering
\small
\begin{tabular}{clcccc}
\toprule
\multicolumn{1}{l}{} & \textbf{Perturbation Type} & \multicolumn{1}{c}{\textbf{\begin{tabular}[c]{@{}l@{}}\total\\ Errors\end{tabular}}} & \multicolumn{1}{c}{\textbf{\begin{tabular}[c]{@{}l@{}}Detected\\ Errors\end{tabular}}} & \multicolumn{1}{c}{\textbf{\begin{tabular}[c]{@{}l@{}}Undetected\\ Errors\end{tabular}}} & \multicolumn{1}{c}{\textbf{\begin{tabular}[c]{@{}c@{}}\% Undetected\\ Errors\end{tabular}}} \\
\midrule

\multirow{7}{*}{\colorbox{myblue}{LF}} & \coherence & 91 & 45 & 46 & \textbf{0.51} \\
 & \comprehensive & 90 & 0 & 90 & 1.00 \\
 & \consistency & 84 & 6 & 78 & 0.93  \\
 & \grammar & 92 & 16 & 76 & 0.83  \\
 & \chronology& 71 & 0 & 71 & 1.00  \\
 & \spelling & 100 & 7 & 93 & 0.93 \\
 \rowcolor{myblue}
 & \textbf{\total} & \textbf{528} & \textbf{74} & \textbf{454} & \textbf{0.86} \\
 \midrule
 
\multirow{7}{*}{\colorbox{mygreen}{F}} & \contextual & 94 & 28 & 66 & \textbf{0.70} \\
 & \entity & 87 & 27 & 60 & 0.69 \\
 & \incorrect& 68 & 15 & 53 & 0.78 \\
 & \numerr & 74 & 15 & 59 & 0.80 \\
 & \opposite & 91 & 28 & 63 & 0.69 \\
 & \remove & 69 & 1 & 68 & 0.99 \\
 \rowcolor{mygreen}
 & \textbf{\total} & \textbf{483} & \textbf{114} & \textbf{369} & \textbf{0.76} \\
 \midrule
 
\multirow{6}{*}{\colorbox{myorange}{IF}} & \assumptions & 81 & 2 & 79 & 0.98 \\
 & \doless & 100 & 17 & 83 & 0.83 \\
 & \domore & 50 & 39 & 11 & 0.22 \\
 & \format & 99 & 24 & 75 & 0.76 \\
 & \sequence & 49 & 4 & 45 & 0.92 \\
 \rowcolor{myorange}
 & \textbf{\total} & \textbf{379} & \textbf{86} & \textbf{293} & \textbf{0.77} \\
 \midrule
 
\multirow{6}{*}{\colorbox{mymagenta}{R}} & \calculation & 149 & 97 & 52 & 0.35 \\
 & \copying & 83 & 58 & 25 & 0.30 \\
 & \final & 97 & 48 & 49 & 0.51 \\
 & \units & 77 & 44 & 33 & 0.43 \\
 & \formula & 88 & 63 & 25 & 0.37 \\
 \rowcolor{mymagenta}
 & \textbf{\total} & \textbf{494} & \textbf{310} & \textbf{184} & \textbf{0.37} \\
 \bottomrule
\end{tabular}
\caption{Results from evaluating \fbi~using \textbf{Axis+Rubrics} evaluator. An error is said to be detected if the evaluator penalizes the score of the perturbed answer.}
\label{tab:apx:single5}
\end{table*}

%----------------------pairwise comparision----------------------

% f(i, r, gold) -> (score)
\begin{table*}[]
\centering
\small
\begin{tabular}{clccccccc}
\toprule
\multicolumn{1}{l}{} & \textbf{Perturbation Type} & \multicolumn{1}{c}{\textbf{\begin{tabular}[c]{@{}l@{}}\total\\ Errors\end{tabular}}} & \multicolumn{1}{l}{\textbf{G}} & \multicolumn{1}{l}{\textbf{P}} & \multicolumn{1}{l}{\textbf{Both \cmark}} & \multicolumn{1}{l}{\textbf{Both \xmark}} & \multicolumn{1}{l}{\textbf{$\neq$}} & \multicolumn{1}{c}{\textbf{\begin{tabular}[c]{@{}c@{}}\% Undetected\\ Errors\end{tabular}}} \\
\midrule

\multirow{7}{*}{\colorbox{myblue}{LF}} & \coherence & 91 & 73 & 0 & 11 & 0 & 7 & 0.20 \\
 & \comprehensive & 90 & 11 & 0 & 57 & 0 & 22 & 0.88 \\
 & \consistency & 84 & 12 & 0 & 59 & 0 & 13 & 0.86 \\
 & \grammar & 92 & 32 & 0 & 46 & 0 & 14 & 0.65 \\
 & \chronology& 71 & 1 & 0 & 68 & 0 & 2 & 0.99 \\
 & \spelling & 100 & 12 & 0 & 77 & 0 & 11 & 0.88 \\
 \rowcolor{myblue}
 & \textbf{\total} & \textbf{528} & \textbf{141} & \textbf{0} & \textbf{318} & \textbf{0} & \textbf{69} & 0.73 \\
 \midrule
 
\multirow{7}{*}{\colorbox{mygreen}{F}} & \contextual & 94 & 55 & 0 & 12 & 0 & 27 & 0.41 \\
 & \entity & 87 & 51 & 0 & 16 & 0 & 20 & 0.41 \\
 & \incorrect& 68 & 32 & 0 & 12 & 0 & 24 & 0.53 \\
 & \numerr & 74 & 29 & 1 & 22 & 0 & 22 & 0.61 \\
 & \opposite & 91 & 55 & 0 & 12 & 0 & 24 & 0.40 \\
 & \remove & 69 & 12 & 0 & 42 & 0 & 15 & 0.83 \\
 \rowcolor{mygreen}
 & \textbf{\total} & \textbf{483} & \textbf{234} & \textbf{1} & \textbf{116} & \textbf{0} & \textbf{132} & 0.52 \\
 \midrule
 
\multirow{6}{*}{\colorbox{myorange}{IF}} & \assumptions & 81 & 6 & 25 & 34 & 0 & 16 & 0.93 \\
 & \doless & 100 & 40 & 0 & 22 & 0 & 38 & 0.60 \\
 & \domore & 50 & 7 & 1 & 17 & 0 & 25 & 0.86 \\
 & \format & 99 & 13 & 0 & 56 & 0 & 30 & 0.87 \\
 & \sequence & 49 & 0 & 0 & 49 & 0 & 0 & 1.00 \\
 \rowcolor{myorange}
 & \textbf{\total} & \textbf{379} & \textbf{66} & \textbf{26} & \textbf{178} & \textbf{0} & \textbf{109} & 0.83 \\
 \midrule
 
\multirow{6}{*}{\colorbox{mymagenta}{R}} & \calculation & 149 & 96 & 1 & 18 & 1 & 32 & 0.35 \\
 & \copying & 83 & 58 & 0 & 7 & 1 & 17 & 0.30 \\
 & \final & 97 & 58 & 1 & 6 & 0 & 32 & 0.40 \\
 & \units & 77 & 48 & 0 & 17 & 1 & 11 & 0.38 \\
 & \formula & 88 & 56 & 1 & 15 & 3 & 13 & 0.36 \\
 \rowcolor{mymagenta}
 & \textbf{\total} & \textbf{494} & \textbf{316} & \textbf{3} & \textbf{63} & \textbf{6} & \textbf{105} & 0.36 \\
 \bottomrule
\end{tabular}
\caption{Results from evaluating \fbi~ using the \textbf{Pairwise}$*$ evaluator. An error is said to be detected if the evaluator chooses the Gold Answer. \textbf{G} indicates the number of times the evaluator has chosen the Gold Answer, \textbf{P} for the Perturbed Answer, \textbf{Both \cmark} when both answers are correct, \textbf{Both \xmark} when both are incorrect, and $\neq$ for verdict inconsistencies. }
\label{tab:apx:pair1}
\end{table*}

% f(i, r, gold) -> (explanation, score)
\begin{table*}[]
\centering
\small
\begin{tabular}{clccccccc}
\toprule
\multicolumn{1}{l}{} & \textbf{Perturbation Type} & \multicolumn{1}{c}{\textbf{\begin{tabular}[c]{@{}l@{}}\total\\ Errors\end{tabular}}} & \multicolumn{1}{l}{\textbf{G}} & \multicolumn{1}{l}{\textbf{P}} & \multicolumn{1}{l}{\textbf{Both \cmark}} & \multicolumn{1}{l}{\textbf{Both \xmark}} & \multicolumn{1}{l}{\textbf{$\neq$}} & \multicolumn{1}{c}{\textbf{\begin{tabular}[c]{@{}c@{}}\% Undetected\\ Errors\end{tabular}}} \\
\midrule

\multirow{7}{*}{\colorbox{myblue}{LF}} & \coherence & 91 & 69 & 0 & 2 & 0 & 18 & 0.22 \\
 & \comprehensive & 90 & 25 & 0 & 18 & 0 & 47 & 0.72 \\
 & \consistency & 84 & 10 & 0 & 40 & 0 & 33 & 0.88 \\
 & \grammar & 92 & 12 & 0 & 24 & 0 & 54 & 0.87 \\
 & \chronology& 71 & 0 & 0 & 50 & 0 & 19 & 1.00 \\
 & \spelling & 100 & 5 & 0 & 56 & 0 & 38 & 0.95 \\
 \rowcolor{myblue}
 & \textbf{\total} & \textbf{528} & \textbf{121} & \textbf{0} & \textbf{190} & \textbf{0} & \textbf{209} & 0.77 \\
 \midrule
 
\multirow{7}{*}{\colorbox{mygreen}{F}} & \contextual & 94 & 76 & 0 & 5 & 0 & 13 & 0.19 \\
 & \entity & 87 & 44 & 0 & 11 & 0 & 28 & 0.47 \\
 & \incorrect& 68 & 36 & 0 & 3 & 0 & 27 & 0.45 \\
 & \numerr & 74 & 34 & 0 & 9 & 0 & 28 & 0.52 \\
 & \opposite & 91 & 39 & 0 & 3 & 0 & 48 & 0.57 \\
 & \remove & 69 & 24 & 0 & 16 & 0 & 28 & 0.65 \\
 \rowcolor{mygreen}
 & \textbf{\total} & \textbf{483} & \textbf{253} & \textbf{0} & \textbf{47} & \textbf{0} & \textbf{172} & 0.46 \\
 \midrule
 
\multirow{6}{*}{\colorbox{myorange}{IF}} & \assumptions & 81 & 4 & 43 & 3 & 0 & 31 & 0.95 \\
 & \doless & 100 & 58 & 0 & 11 & 0 & 30 & 0.41 \\
 & \domore & 50 & 24 & 2 & 0 & 0 & 24 & 0.52 \\
 & \format & 99 & 35 & 0 & 27 & 0 & 23 & 0.59 \\
 & \sequence & 49 & 0 & 0 & 23 & 0 & 26 & 1.00 \\
 \rowcolor{myorange}
 & \textbf{\total} & \textbf{379} & \textbf{121} & \textbf{45} & \textbf{64} & \textbf{0} & \textbf{134} & 0.67 \\
 \midrule
 
\multirow{6}{*}{\colorbox{mymagenta}{R}} & \calculation & 149 & 77 & 0 & 6 & 1 & 38 & 0.37 \\
 & \copying & 83 & 40 & 0 & 1 & 1 & 18 & 0.33 \\
 & \final & 97 & 59 & 0 & 0 & 0 & 18 & 0.23 \\
 & \units & 77 & 38 & 0 & 7 & 0 & 20 & 0.42 \\
 & \formula & 88 & 39 & 0 & 4 & 1 & 23 & 0.42 \\
 \rowcolor{mymagenta}
 & \textbf{\total} & \textbf{494} & \textbf{253} & \textbf{0} & \textbf{18} & \textbf{3} & \textbf{117} & 0.35 \\
 \bottomrule
\end{tabular}
\caption{Results from evaluating \fbi~ using the \textbf{Pairwise} evaluator. An error is said to be detected if the evaluator chooses the Gold Answer. \textbf{G} indicates the number of times the evaluator has chosen the Gold Answer, \textbf{P} for the Perturbed Answer, \textbf{Both \cmark} when both answers are correct, \textbf{Both \xmark} when both are incorrect, and $\neq$ for verdict inconsistencies.}
\label{tab:apx:pair2}
\end{table*}

% f(i, r, gold, rules) -> (explanation, score)
\begin{table*}[]
\centering
\small
\begin{tabular}{clccccccc}
\toprule
\multicolumn{1}{l}{} & \textbf{Perturbation Type} & \multicolumn{1}{c}{\textbf{\begin{tabular}[c]{@{}l@{}}\total\\ Errors\end{tabular}}} & \multicolumn{1}{l}{\textbf{G}} & \multicolumn{1}{l}{\textbf{P}} & \multicolumn{1}{l}{\textbf{Both \cmark}} & \multicolumn{1}{l}{\textbf{Both \xmark}} & \multicolumn{1}{l}{\textbf{$\neq$}} & \multicolumn{1}{c}{\textbf{\begin{tabular}[c]{@{}c@{}}\% Undetected\\ Errors\end{tabular}}} \\
\midrule

\multirow{7}{*}{\colorbox{myblue}{LF}} & \coherence & 91 & 82 & 0 & 2 & 0 & 7 & 0.10 \\
 & \comprehensive & 90 & 28 & 0 & 25 & 0 & 37 & 0.69 \\
 & \consistency & 84 & 10 & 0 & 46 & 0 & 28 & 0.88 \\
 & \grammar & 92 & 8 & 0 & 24 & 0 & 60 & 0.91 \\
 & \chronology& 71 & 0 & 0 & 51 & 0 & 20 & 1.00 \\
 & \spelling & 100 & 4 & 0 & 48 & 0 & 48 & 0.96 \\
 \rowcolor{myblue}
 & \textbf{\total} & \textbf{528} & \textbf{132} & \textbf{0} & \textbf{196} & \textbf{0} & \textbf{200} & 0.75 \\
 \midrule
 
\multirow{7}{*}{\colorbox{mygreen}{F}} & \contextual & 94 & 36 & 0 & 9 & 0 & 48 & 0.61 \\
 & \entity & 87 & 37 & 0 & 14 & 0 & 34 & 0.56 \\
 & \incorrect& 68 & 27 & 0 & 4 & 0 & 36 & 0.60 \\
 & \numerr & 74 & 27 & 0 & 13 & 0 & 32 & 0.63 \\
 & \opposite & 91 & 32 & 0 & 6 & 0 & 53 & 0.65 \\
 & \remove & 69 & 19 & 0 & 18 & 0 & 32 & 0.72 \\
 \rowcolor{mygreen}
 & \textbf{\total} & \textbf{483} & \textbf{178} & \textbf{0} & \textbf{64} & \textbf{0} & \textbf{235} & 0.63 \\
 \midrule
 
\multirow{6}{*}{\colorbox{myorange}{IF}} & \assumptions & 81 & 3 & 57 & 5 & 0 & 16 & 0.96 \\
 & \doless & 100 & 60 & 2 & 15 & 0 & 23 & 0.40 \\
 & \domore & 50 & 25 & 3 & 0 & 0 & 22 & 0.50 \\
 & \format & 99 & 33 & 0 & 29 & 0 & 37 & 0.67 \\
 & \sequence & 49 & 1 & 0 & 24 & 0 & 24 & 0.98 \\
 \rowcolor{myorange}
 & \textbf{\total} & \textbf{379} & \textbf{122} & \textbf{62} & \textbf{73} & \textbf{0} & \textbf{122} & 0.68 \\
 \midrule
 
\multirow{6}{*}{\colorbox{mymagenta}{R}} & \calculation & 149 & 82 & 1 & 12 & 0 & 46 & 0.42 \\
 & \copying & 83 & 55 & 0 & 6 & 0 & 18 & 0.30 \\
 & \final & 97 & 47 & 1 & 0 & 0 & 42 & 0.48 \\
 & \units & 77 & 47 & 0 & 10 & 1 & 19 & 0.39 \\
 & \formula & 88 & 46 & 1 & 7 & 0 & 27 & 0.43 \\
 \rowcolor{mymagenta}
 & \textbf{\total} & \textbf{494} & \textbf{277} & \textbf{3} & \textbf{35} & \textbf{1} & \textbf{152} & 0.41 \\
 \bottomrule
\end{tabular}
\caption{Results from evaluating \fbi~ using the \textbf{Rules} evaluator. An error is said to be detected if the evaluator chooses the Gold Answer. \textbf{G} indicates the number of times the evaluator has chosen the Gold Answer, \textbf{P} for the Perturbed Answer, \textbf{Both \cmark} when both answers are correct, \textbf{Both \xmark} when both are incorrect, and $\neq$ for verdict inconsistencies.}
\label{tab:apx:pair3}
\end{table*}

% f(i, r, gold, axes) -> (explanation, score)
\begin{table*}[]
\centering
\small
\begin{tabular}{clccccccc}
\toprule
\multicolumn{1}{l}{} & \textbf{Perturbation Type} & \multicolumn{1}{c}{\textbf{\begin{tabular}[c]{@{}l@{}}\total\\ Errors\end{tabular}}} & \multicolumn{1}{l}{\textbf{G}} & \multicolumn{1}{l}{\textbf{P}} & \multicolumn{1}{l}{\textbf{Both \cmark}} & \multicolumn{1}{l}{\textbf{Both \xmark}} & \multicolumn{1}{l}{\textbf{$\neq$}} & \multicolumn{1}{c}{\textbf{\begin{tabular}[c]{@{}c@{}}\% Undetected\\ Errors\end{tabular}}} \\
\midrule

\multirow{7}{*}{\colorbox{myblue}{LF}} & \coherence & 91 & 82 & 0 & 1 & 0 & 8 & 0.10 \\
 & \comprehensive & 90 & 49 & 0 & 9 & 0 & 32 & 0.46 \\
 & \consistency & 84 & 16 & 0 & 50 & 0 & 18 & 0.81 \\
 & \grammar & 92 & 34 & 0 & 26 & 0 & 32 & 0.63 \\
 & \chronology& 71 & 0 & 0 & 57 & 0 & 14 & 1.00 \\
 & \spelling & 100 & 11 & 0 & 58 & 0 & 31 & 0.89 \\
 \rowcolor{myblue}
 & \textbf{\total} & \textbf{528} & \textbf{192} & \textbf{0} & \textbf{201} & \textbf{0} & \textbf{135} & 0.64 \\
 \midrule
 
\multirow{7}{*}{\colorbox{mygreen}{F}} & \contextual & 94 & 60 & 0 & 8 & 0 & 26 & 0.36 \\
 & \entity & 87 & 60 & 0 & 11 & 0 & 16 & 0.31 \\
 & \incorrect& 68 & 41 & 0 & 4 & 0 & 23 & 0.40 \\
 & \numerr & 74 & 45 & 0 & 10 & 0 & 19 & 0.39 \\
 & \opposite & 91 & 61 & 0 & 7 & 0 & 23 & 0.33 \\
 & \remove & 69 & 5 & 0 & 58 & 0 & 6 & 0.93 \\
 \rowcolor{mygreen}
 & \textbf{\total} & \textbf{483} & \textbf{272} & \textbf{0} & \textbf{98} & \textbf{0} & \textbf{113} & 0.44 \\
 \midrule
 
\multirow{6}{*}{\colorbox{myorange}{IF}} & \assumptions & 81 & 2 & 62 & 4 & 0 & 13 & 0.98 \\
 & \doless & 100 & 57 & 0 & 11 & 0 & 32 & 0.43 \\
 & \domore & 50 & 40 & 2 & 3 & 0 & 5 & 0.20 \\
 & \format & 99 & 53 & 0 & 13 & 0 & 33 & 0.46 \\
 & \sequence & 49 & 5 & 0 & 23 & 0 & 21 & 0.9 \\
 \rowcolor{myorange}
 & \textbf{\total} & \textbf{379} & \textbf{157} & \textbf{64} & \textbf{54} & \textbf{0} & \textbf{104} & 0.59 \\
 \midrule
 
\multirow{6}{*}{\colorbox{mymagenta}{R}} & \calculation & 149 & 108 & 1 & 16 & 0 & 23 & 0.27 \\
 & \copying & 83 & 69 & 1 & 7 & 0 & 6 & 0.17 \\
 & \final & 97 & 75 & 1 & 2 & 0 & 19 & 0.23 \\
 & \units & 77 & 42 & 0 & 20 & 0 & 15 & 0.45 \\
 & \formula & 88 & 64 & 0 & 12 & 0 & 12 & 0.27 \\
 \rowcolor{mymagenta}
 & \textbf{\total} & \textbf{494} & \textbf{358} & \textbf{3} & \textbf{57} & \textbf{0} & \textbf{75} & 0.27 \\
 \bottomrule
\end{tabular}
\caption{Results from evaluating \fbi~ using the \textbf{Axis} evaluator. An error is said to be detected if the evaluator chooses the Gold Answer. \textbf{G} indicates the number of times the evaluator has chosen the Gold Answer, \textbf{P} for the Perturbed Answer, \textbf{Both \cmark} when both answers are correct, \textbf{Both \xmark} when both are incorrect, and $\neq$ for verdict inconsistencies.}
\label{tab:apx:pair4}
\end{table*}

% f(i, r, gold, axes+rules) -> (explanation, score)
\begin{table*}[]
\centering
\small
\begin{tabular}{clccccccc}
\toprule
\multicolumn{1}{l}{} & \textbf{Perturbation Type} & \multicolumn{1}{c}{\textbf{\begin{tabular}[c]{@{}l@{}}\total\\ Errors\end{tabular}}} & \multicolumn{1}{l}{\textbf{G}} & \multicolumn{1}{l}{\textbf{P}} & \multicolumn{1}{l}{\textbf{Both \cmark}} & \multicolumn{1}{l}{\textbf{Both \xmark}} & \multicolumn{1}{l}{\textbf{$\neq$}} & \multicolumn{1}{c}{\textbf{\begin{tabular}[c]{@{}c@{}}\% Undetected\\ Errors\end{tabular}}} \\
\midrule

\multirow{7}{*}{\colorbox{myblue}{LF}} & \coherence & 91 & 84 & 0 & 2 & 0 & 5 & 0.08 \\
 & \comprehensive & 90 & 47 & 0 & 13 & 0 & 29 & 0.47 \\
 & \consistency & 84 & 16 & 0 & 52 & 0 & 16 & 0.81 \\
 & \grammar & 92 & 33 & 0 & 27 & 0 & 32 & 0.64 \\
 & \chronology& 71 & 2 & 0 & 61 & 0 & 8 & 0.97 \\
 & \spelling & 100 & 7 & 0 & 53 & 0 & 40 & 0.93 \\
 \rowcolor{myblue}
 & \textbf{\total} & \textbf{528} & \textbf{189} & \textbf{0} & \textbf{208} & \textbf{0} & \textbf{130} & 0.64 \\
 \midrule
 
\multirow{7}{*}{\colorbox{mygreen}{F}} & \contextual & 94 & 56 & 0 & 8 & 0 & 28 & 0.39 \\
 & \entity & 87 & 61 & 0 & 11 & 0 & 13 & 0.28 \\
 & \incorrect& 68 & 43 & 2 & 2 & 0 & 20 & 0.36 \\
 & \numerr & 74 & 43 & 0 & 8 & 0 & 21 & 0.40 \\
 & \opposite & 91 & 66 & 0 & 5 & 0 & 20 & 0.27 \\
 & \remove & 69 & 9 & 0 & 34 & 0 & 26 & 0.87 \\
 \rowcolor{mygreen}
 & \textbf{\total} & \textbf{483} & \textbf{278} & \textbf{2} & \textbf{68} & \textbf{0} & \textbf{128} & 0.42 \\
 \midrule
 
\multirow{6}{*}{\colorbox{myorange}{IF}} & \assumptions & 81 & 2 & 65 & 2 & 0 & 12 & 0.98 \\
 & \doless & 100 & 59 & 0 & 8 & 0 & 33 & 0.41 \\
 & \domore & 50 & 35 & 2 & 0 & 0 & 13 & 0.30 \\
 & \format & 99 & 51 & 0 & 18 & 0 & 30 & 0.48 \\
 & \sequence & 49 & 1 & 0 & 29 & 0 & 19 & 0.98 \\
 \rowcolor{myorange}
 & \textbf{\total} & \textbf{379} & \textbf{148} & \textbf{67} & \textbf{57} & \textbf{0} & \textbf{107} & 0.61 \\
 \midrule
 
\multirow{6}{*}{\colorbox{mymagenta}{R}} & \calculation & 149 & 93 & 0 & 12 & 0 & 23 & 0.27 \\
 & \copying & 83 & 58 & 0 & 6 & 0 & 12 & 0.24 \\
 & \final & 97 & 57 & 2 & 2 & 0 & 26 & 0.34 \\
 & \units & 77 & 38 & 0 & 19 & 0 & 16 & 0.48 \\
 & \formula & 88 & 54 & 0 & 10 & 0 & 16 & 0.33 \\
 \rowcolor{mymagenta}
 & \textbf{\total} & \textbf{494} & \textbf{300} & \textbf{2} & \textbf{49} & \textbf{0} & \textbf{93} & 0.32 \\
 \bottomrule
\end{tabular}
\caption{Results from evaluating \fbi~ using the \textbf{Axis+Rules} evaluator. An error is said to be detected if the evaluator chooses the Gold Answer. \textbf{G} indicates the number of times the evaluator has chosen the Gold Answer, \textbf{P} for the Perturbed Answer, \textbf{Both \cmark} when both answers are correct, \textbf{Both \xmark} when both are incorrect, and $\neq$ for verdict inconsistencies.}
\label{tab:apx:pair5}
\end{table*}

%----------------------refernce guided----------------------

% reference
\begin{table*}[]
\centering
\small
\begin{tabular}{clcccccc}
\toprule
\multicolumn{1}{l}{} & \textbf{Perturbation Type} & \multicolumn{1}{c}{\textbf{\begin{tabular}[c]{@{}l@{}}\total\\ Errors\end{tabular}}} & \textbf{10} & \textbf{9} & \textbf{8} & \multicolumn{1}{l}{\textbf{$<$8}} & \multicolumn{1}{c}{\textbf{\begin{tabular}[c]{@{}c@{}}\% Undetected\\ Errors\end{tabular}}} \\
\midrule

\multirow{7}{*}{\colorbox{myblue}{LF}} & \coherence & 91 & 2 & 5 & 9 & 75 & 0.02 \\
 & \comprehensive & 90 & 32 & 39 & 6 & 13 & 0.36 \\
 & \consistency & 84 & 31 & 27 & 4 & 22 & 0.37 \\
 & \grammar & 92 & 10 & 51 & 9 & 22 & 0.11 \\
 & \chronology& 71 & 47 & 21 & 2 & 1 & 0.66 \\
 & \spelling & 100 & 14 & 72 & 4 & 10 & 0.14 \\
 \rowcolor{myblue}
 & \textbf{\total} & \textbf{528} & \textbf{136} & \textbf{215} & \textbf{34} & \textbf{143} & 0.26 \\
 \midrule
 
\multirow{7}{*}{\colorbox{mygreen}{F}} & \contextual & 94 & 1 & 27 & 11 & 55 & 0.01 \\
 & \entity & 87 & 8 & 16 & 16 & 47 & 0.09 \\
 & \incorrect& 68 & 2 & 15 & 12 & 39 & 0.03 \\
 & \numerr & 74 & 6 & 21 & 15 & 32 & 0.08 \\
 & \opposite & 91 & 0 & 11 & 4 & 76 & 0.00 \\
 & \remove & 69 & 36 & 18 & 10 & 5 & 0.52 \\
 \rowcolor{mygreen}
 & \textbf{\total} & \textbf{483} & \textbf{53} & \textbf{108} & \textbf{68} & \textbf{254} & 0.11 \\
 \midrule
 
\multirow{6}{*}{\colorbox{myorange}{IF}} & \assumptions & 81 & 50 & 17 & 4 & 10 & 0.62 \\
 & \doless & 100 & 32 & 6 & 15 & 47 & 0.32 \\
 & \domore & 50 & 22 & 10 & 12 & 6 & 0.44 \\
 & \format & 99 & 43 & 18 & 7 & 30 & 0.43 \\
 & \sequence & 49 & 39 & 8 & 2 & 0 & 0.80 \\
 \rowcolor{myorange}
 & \textbf{\total} & \textbf{379} & \textbf{186} & \textbf{59} & \textbf{40} & \textbf{93} & 0.49 \\
 \midrule
 
\multirow{6}{*}{\colorbox{mymagenta}{R}} & \calculation & 149 & 6 & 6 & 6 & 131 & 0.04 \\
 & \copying & 83 & 4 & 4 & 3 & 72 & 0.05 \\
 & \final & 97 & 1 & 2 & 4 & 89 & 0.01 \\
 & \units & 77 & 10 & 10 & 4 & 53 & 0.13 \\
 & \formula & 88 & 1 & 12 & 1 & 74 & 0.01 \\
 \rowcolor{mymagenta}
 & \textbf{\total} & \textbf{494} & \textbf{22} & \textbf{34} & \textbf{18} & \textbf{419} & 0.04 \\
 \bottomrule
\end{tabular}
\caption{Results from evaluating \fbi~ using the \textbf{Reference} evaluator. An error is said to be detected if the evaluator gives a perfect score of 10 to the perturbed answer. \textbf{10} indicates the number of times the evaluator has given the score of 10, \textbf{9} for the score of 9, \textbf{8} for the score of 8 and \textbf{$<$8} for scores less than 8.}
\label{tab:apx:ref1}
\end{table*}

% prometheus
\begingroup
\setlength{\tabcolsep}{4pt}
\begin{table*}[]
\centering
\small
\begin{tabular}{clccccccccc}
\toprule
\multicolumn{1}{l}{} &  & \multicolumn{1}{l}{} & \multicolumn{4}{c}{\textbf{Generic}} & \multicolumn{4}{c}{\textbf{Specific}} \\
\cmidrule(lr){4-7} \cmidrule(lr){8-11}

\multicolumn{1}{l}{} & \textbf{Perturbation Type} & \textbf{\# Errs} & \textbf{5} & \textbf{4} & \multicolumn{1}{l}{\textbf{$<$4}} & \textbf{\% Errors} & \textbf{5} & \textbf{4} & \multicolumn{1}{l}{\textbf{$<$4}} & \textbf{\% Errors} \\
\midrule

\multirow{7}{*}{\colorbox{myblue}{LF}} & \coherence & 91 & 9 & 33 & 49 & 0.10 & 17 & 18 & 56 & 0.19 \\
 & \comprehensive & 90 & 40 & 42 & 8 & 0.44 & 40 & 46 & 4 & 0.44 \\
 & \consistency & 84 & 36 & 36 & 12 & 0.43 & 50 & 26 & 8 & 0.60 \\
 & \grammar & 92 & 44 & 39 & 9 & 0.48 & 56 & 31 & 5 & 0.61 \\
 & \chronology& 71 & 43 & 24 & 4 & 0.61 & 42 & 23 & 6 & 0.59 \\
 & \spelling & 100 & 46 & 49 & 5 & 0.46 & 65 & 28 & 7 & 0.65 \\
 \rowcolor{myblue}
 & \textbf{\total} & \textbf{528} & \textbf{218} & \textbf{223} & \textbf{87} & 0.41 & \textbf{270} & \textbf{172} & \textbf{86} & 0.51 \\
 \midrule
 
\multirow{7}{*}{\colorbox{mygreen}{F}} & \contextual & 94 & 46 & 39 & 9 & 0.49 & 56 & 25 & 13 & 0.60 \\
 & \entity & 87 & 34 & 41 & 12 & 0.39 & 51 & 22 & 14 & 0.59 \\
 & \incorrect& 68 & 29 & 30 & 9 & 0.43 & 45 & 18 & 5 & 0.66 \\
 & \numerr & 74 & 36 & 32 & 6 & 0.49 & 47 & 18 & 9 & 0.64 \\
 & \opposite & 91 & 37 & 41 & 13 & 0.41 & 52 & 28 & 11 & 0.57 \\
 & \remove & 69 & 41 & 27 & 1 & 0.59 & 50 & 18 & 1 & 0.72 \\
 \rowcolor{mygreen}
 & \textbf{\total} & \textbf{483} & \textbf{223} & \textbf{210} & \textbf{50} & 0.46 & \textbf{301} & \textbf{129} & \textbf{53} & 0.62 \\
 \midrule
 
\multirow{6}{*}{\colorbox{myorange}{IF}} & \assumptions & 81 & 38 & 41 & 2 & 0.47 & 56 & 25 & 0 & 0.69 \\
 & \doless & 100 & 53 & 44 & 3 & 0.53 & 54 & 44 & 2 & 0.54 \\
 & \domore & 50 & 17 & 24 & 9 & 0.34 & 16 & 28 & 6 & 0.32 \\
 & \format & 99 & 49 & 43 & 7 & 0.49 & 53 & 35 & 11 & 0.54 \\
 & \sequence & 49 & 18 & 28 & 3 & 0.37 & 21 & 21 & 7 & 0.43 \\
 \rowcolor{myorange}
 & \textbf{\total} & \textbf{379} & \textbf{175} & \textbf{180} & \textbf{24} & 0.46 & \textbf{200} & \textbf{153} & \textbf{26} & 0.53 \\
 \midrule
 
\multirow{6}{*}{\colorbox{mymagenta}{R}} & \calculation & 149 & 23 & 67 & 59 & 0.15 & 14 & 75 & 60 & 0.09 \\
 & \copying & 83 & 11 & 38 & 34 & 0.13 & 9 & 42 & 32 & 0.11 \\
 & \final & 97 & 22 & 46 & 29 & 0.23 & 10 & 54 & 33 & 0.10 \\
 & \units & 77 & 16 & 23 & 38 & 0.21 & 8 & 34 & 35 & 0.10 \\
 & \formula & 88 & 17 & 48 & 23 & 0.19 & 17 & 38 & 33 & 0.19 \\
 \rowcolor{mymagenta}
 & \textbf{\total} & \textbf{494} & \textbf{89} & \textbf{222} & \textbf{183} & 0.18 & \textbf{58} & \textbf{243} & \textbf{193} & 0.12 \\
 \bottomrule
\end{tabular}
\caption{Results from evaluating \fbi~ using the \textbf{Prometheus} evaluator. An error is said to be detected if the evaluator gives a perfect score of 5 to the perturbed answer. \textbf{5} indicates the number of times the evaluator has given the score of 5, \textbf{4} for the score of 4, and \textbf{$<$4} for scores less than 4. \textbf{Generic} indicates evaluating with general scoring rubrics and \textbf{Specific} indicates evaluating with task-specific rubrics. }
\label{tab:apx:ref2}
\end{table*}
\endgroup

% ------------ Reference-based - other models ------------

% f(i, r) -> (explanation, score)	- other models
\begingroup
\setlength{\tabcolsep}{5pt}
\begin{table*}[]
\centering
\small
\begin{tabular}{clcccccccccc}
\toprule
\multicolumn{1}{l}{\multirow{2}{*}{}} & \multirow{2}{*}{} & \multicolumn{1}{l}{\multirow{2}{*}{}} & \multicolumn{3}{c}{\textbf{Llama-3-70B-Instruct}} & \multicolumn{3}{c}{\textbf{Claude-3-Opus}} & \multicolumn{3}{c}{\textbf{Gemini-1.5-Pro}} \\
\cmidrule(lr){4-6} \cmidrule(lr){7-9} \cmidrule(lr){10-12}

\multicolumn{1}{l}{} & \textbf{Perturbation Type} & \multicolumn{1}{l}{\textbf{\# Errs}} & \multicolumn{1}{l}{\textbf{\# DE}} & \multicolumn{1}{l}{\textbf{\# UE}} & \multicolumn{1}{l}{\textbf{\% UE}} & \multicolumn{1}{l}{\textbf{\# DE}} & \multicolumn{1}{l}{\textbf{\# UE}} & \multicolumn{1}{l}{\textbf{\% UE}} & \multicolumn{1}{l}{\textbf{\# DE}} & \multicolumn{1}{l}{\textbf{\# UE}} & \multicolumn{1}{l}{\textbf{\% UE}}  \\
\midrule

\multirow{7}{*}{\colorbox{myblue}{LF}} & \coherence & 91 & 61 & 21 & 0.29 & 72 & 18 & 0.20 & 83 & 8 & 0.09 \\
 & \comprehensive & 90 & 22 & 59 & 0.82 & 19 & 71 & 0.79 & 29 & 60 & 0.67 \\
 & \consistency & 84 & 9 & 65 & 1.00 & 18 & 65 & 0.78 & 29 & 55 & 0.65 \\
 & \grammar & 92 & 8 & 80 & 0.95 & 12 & 80 & 0.87 & 29 & 63 & 0.68 \\
 & \chronology& 71 & 1 & 60 & 1.15 & 6 & 64 & 0.91 & 18 & 53 & 0.75 \\
 & \spelling & 100 & 7 & 85 & 1.01 & 13 & 80 & 0.92 & 18 & 82 & 0.82 \\
 \rowcolor{myblue}
 & \textbf{\total} & \textbf{528} & \textbf{108} & \textbf{370} & 0.86 & \textbf{140} & \textbf{378} & 0.74 & \textbf{206} & \textbf{321} & 0.61 \\
 \midrule
 
\multirow{7}{*}{\colorbox{mygreen}{F}} & \contextual & 94 & 5 & 82 & 1.03 & 14 & 80 & 0.85 & 28 & 66 & 0.70 \\
 & \entity & 87 & 13 & 64 & 0.94 & 22 & 65 & 0.75 & 32 & 55 & 0.63 \\
 & \incorrect& 68 & 6 & 55 & 1.02 & 10 & 58 & 0.85 & 15 & 53 & 0.78 \\
 & \numerr & 74 & 6 & 61 & 1.00 & 8 & 66 & 0.89 & 11 & 63 & 0.85 \\
 & \opposite & 91 & 10 & 74 & 0.96 & 17 & 74 & 0.81 & 32 & 59 & 0.65 \\
 & \remove & 69 & 18 & 49 & 0.74 & 8 & 61 & 0.88 & 13 & 56 & 0.81 \\
 \rowcolor{mygreen}
 & \textbf{\total} & \textbf{483} & \textbf{58} & \textbf{385} & 0.95 & \textbf{79} & \textbf{404} & 0.84 & \textbf{131} & \textbf{352} & 0.73 \\
 \midrule
 
\multirow{6}{*}{\colorbox{myorange}{IF}} & \assumptions & 81 & 10 & 55 & 1.10 & 10 & 71 & 0.88 & 25 & 56 & 0.69 \\
 & \doless & 100 & 34 & 60 & 0.68 & 45 & 54 & 0.55 & 59 & 41 & 0.41 \\
 & \domore & 50 & 11 & 35 & 0.81 & 11 & 39 & 0.78 & 26 & 24 & 0.48 \\
 & \format & 99 & 12 & 53 & 1.71 & 25 & 74 & 0.75 & 49 & 49 & 0.51 \\
 & \sequence & 49 & 16 & 33 & 0.67 & 4 & 45 & 0.92 & 17 & 32 & 0.65 \\
 \rowcolor{myorange}
 & \textbf{\total} & \textbf{379} & \textbf{83} & \textbf{236} & 0.90 & \textbf{95} & \textbf{283} & 0.75 & \textbf{176} & \textbf{202} & 0.54 \\
 \midrule
 
\multirow{6}{*}{\colorbox{mymagenta}{R}} & \calculation & 149 & 55 & 82 & 0.65 & 90 & 59 & 0.40 & 81 & 64 & 0.43 \\
 & \copying & 83 & 27 & 47 & 0.71 & 42 & 41 & 0.49 & 54 & 28 & 0.34 \\
 & \final & 97 & 18 & 70 & 0.88 & 35 & 62 & 0.64 & 36 & 60 & 0.63 \\
 & \units & 77 & 34 & 37 & 0.56 & 50 & 27 & 0.35 & 59 & 17 & 0.22 \\
 & \formula & 88 & 25 & 54 & 0.77 & 43 & 44 & 0.51 & 55 & 32 & 0.37 \\
 \rowcolor{mymagenta}
 & \textbf{\total} & \textbf{494} & \textbf{159} & \textbf{290} & 0.71 & \textbf{260} & \textbf{233} & 0.47 & \textbf{285} & \textbf{201} & 0.41 \\
 \bottomrule
\end{tabular}
\caption{Results from evaluating \fbi~using \textbf{Vanilla}-\llama,\claude~and \gemini~evaluators. An error is said to be detected if the evaluator penalizes the score of the perturbed answer.}
\label{tab:apx}
\end{table*}
\endgroup

\begingroup
\setlength{\tabcolsep}{3pt}
\begin{table*}[]
\centering
\small
\begin{tabular}{clccccccccccccc}
\toprule
\multicolumn{1}{l}{\multirow{2}{*}{}} & \multirow{2}{*}{} & \multicolumn{1}{l}{\multirow{2}{*}{}} & \multicolumn{6}{c}{\textbf{Llama-3-70B-Instruct}} & \multicolumn{6}{c}{\textbf{Gemini-1.5-Pro}} \\
\cmidrule(lr){4-9} \cmidrule(lr){10-15}

\multicolumn{1}{l}{} & \textbf{Perturbation Type} & \multicolumn{1}{l}{\textbf{\# Errs}} & \multicolumn{1}{l}{\textbf{G}} & \multicolumn{1}{l}{\textbf{P}} & \multicolumn{1}{l}{\textbf{Both \cmark}} & \multicolumn{1}{l}{\textbf{Both \xmark}} & \multicolumn{1}{l}{\textbf{$\neq$}} & \multicolumn{1}{l}{\textbf{\% Errs}} & \multicolumn{1}{l}{\textbf{G}} & \multicolumn{1}{l}{\textbf{P}} & \multicolumn{1}{l}{\textbf{Both \cmark}} & \multicolumn{1}{l}{\textbf{Both \xmark}} & \multicolumn{1}{l}{\textbf{$\neq$}} & \multicolumn{1}{l}{\textbf{\% Errs}} \\
\midrule

\multirow{7}{*}{\colorbox{myblue}{LF}} & \coherence & 91 & 59 & 0 & 0 & 0 & 18 & 0.23 & 77 & 0 & 2 & 0 & 12 & 0.15 \\
 & \comprehensive & 90 & 40 & 0 & 0 & 0 & 34 & 0.46 & 40 & 0 & 24 & 0 & 26 & 0.56 \\
 & \consistency & 84 & 8 & 0 & 2 & 0 & 69 & 0.90 & 13 & 0 & 49 & 0 & 22 & 0.85 \\
 & \grammar & 92 & 6 & 0 & 9 & 0 & 66 & 0.93 & 11 & 0 & 42 & 0 & 39 & 0.88 \\
 & \chronology& 71 & 0 & 0 & 1 & 0 & 63 & 1.00 & 3 & 0 & 52 & 0 & 16 & 0.96 \\
 & \spelling & 100 & 4 & 0 & 18 & 0 & 64 & 0.95 & 3 & 0 & 76 & 0 & 21 & 0.97 \\
 \rowcolor{myblue}
 & \textbf{\total} & \textbf{528} & \textbf{117} & \textbf{0} & \textbf{30} & \textbf{0} & \textbf{314} & 0.75 & \textbf{147} & \textbf{0} & \textbf{245} & \textbf{0} & \textbf{136} & 0.72 \\
 \midrule
 
\multirow{7}{*}{\colorbox{mygreen}{F}} & \contextual & 94 & 20 & 0 & 5 & 0 & 40 & 0.69 & 43 & 1 & 8 & 0 & 42 & 0.54 \\
 & \entity & 87 & 24 & 0 & 5 & 0 & 34 & 0.62 & 43 & 0 & 14 & 0 & 30 & 0.51 \\
 & \incorrect& 68 & 11 & 1 & 2 & 0 & 34 & 0.77 & 30 & 0 & 2 & 0 & 36 & 0.56 \\
 & \numerr & 74 & 16 & 0 & 2 & 0 & 38 & 0.71 & 33 & 0 & 10 & 0 & 31 & 0.55 \\
 & \opposite & 91 & 14 & 0 & 4 & 0 & 46 & 0.78 & 41 & 0 & 5 & 0 & 45 & 0.55 \\
 & \remove & 69 & 24 & 0 & 4 & 0 & 33 & 0.61 & 12 & 0 & 35 & 0 & 22 & 0.83 \\
 \rowcolor{mygreen}
 & \textbf{\total} & \textbf{483} & \textbf{109} & \textbf{1} & \textbf{22} & \textbf{0} & \textbf{225} & 0.69 & \textbf{202} & \textbf{1} & \textbf{74} & \textbf{0} & \textbf{206} & 0.58 \\
 \midrule
 
\multirow{6}{*}{\colorbox{myorange}{IF}} & \assumptions & 81 & 2 & 21 & 0 & 0 & 12 & 0.94 & 25 & 20 & 1 & 0 & 35 & 0.69 \\
 & \doless & 100 & 44 & 1 & 1 & 0 & 37 & 0.47 & 38 & 0 & 11 & 2 & 49 & 0.62 \\
 & \domore & 50 & 12 & 9 & 1 & 0 & 14 & 0.67 & 14 & 3 & 0 & 1 & 31 & 0.71 \\
 & \format & 99 & 17 & 0 & 10 & 0 & 28 & 0.69 & 33 & 0 & 24 & 10 & 31 & 0.66 \\
 & \sequence & 49 & 0 & 0 & 0 & 0 & 41 & 1.00 & 4 & 0 & 18 & 0 & 27 & 0.92 \\
 \rowcolor{myorange}
 & \textbf{\total} & \textbf{379} & \textbf{75} & \textbf{31} & \textbf{12} & \textbf{0} & \textbf{132} & 0.70 & \textbf{114} & \textbf{23} & \textbf{54} & \textbf{13} & \textbf{173} & 0.70 \\
 \midrule
 
\multirow{6}{*}{\colorbox{mymagenta}{R}} & \calculation & 149 & 48 & 0 & 30 & 0 & 44 & 0.61 & 89 & 1 & 12 & 0 & 47 & 0.40 \\
 & \copying & 83 & 30 & 0 & 6 & 1 & 28 & 0.54 & 57 & 1 & 5 & 1 & 19 & 0.31 \\
 & \final & 97 & 30 & 0 & 3 & 0 & 41 & 0.59 & 59 & 2 & 0 & 1 & 35 & 0.39 \\
 & \units & 77 & 27 & 0 & 11 & 0 & 25 & 0.57 & 40 & 0 & 12 & 1 & 24 & 0.48 \\
 & \formula & 88 & 25 & 0 & 23 & 0 & 27 & 0.67 & 55 & 1 & 8 & 2 & 22 & 0.38 \\
 \rowcolor{mymagenta}
 & \textbf{\total} & \textbf{494} & \textbf{160} & \textbf{0} & \textbf{73} & \textbf{1} & \textbf{165} & 0.60 & \textbf{300} & \textbf{5} & \textbf{37} & \textbf{5} & \textbf{147} & 0.39 \\
 \bottomrule
\end{tabular}
\caption{Results from evaluating \fbi~ using the \textbf{Axis+Rules}-\llama,\claude~and \gemini~evaluators. An error is said to be detected if the evaluator chooses the Gold Answer. \textbf{G} indicates the number of times the evaluator has chosen the Gold Answer, \textbf{P} for the Perturbed Answer, \textbf{Both \cmark} when both answers are correct, \textbf{Both \xmark} when both are incorrect, and $\neq$ for verdict inconsistencies.}
\label{}
\end{table*}
\endgroup

\begingroup
\setlength{\tabcolsep}{4pt}
\begin{table*}[]
\centering
\small
\begin{tabular}{clccccccccccc}
\toprule
\multicolumn{1}{l}{\multirow{2}{*}{}} & \multirow{2}{*}{} & \multicolumn{1}{l}{\multirow{2}{*}{}} & \multicolumn{5}{c}{\textbf{Llama-3-70B-Instruct}} & \multicolumn{5}{c}{\textbf{Gemini-1.5-Pro}} \\
\cmidrule(lr){4-8} \cmidrule(lr){9-13}

\multicolumn{1}{l}{} & \textbf{Perturbation Type} & \multicolumn{1}{l}{\textbf{\# Errs}} & \textbf{10} & \textbf{9} & \textbf{8} & \multicolumn{1}{l}{\textbf{$<$8}} & \multicolumn{1}{l}{\textbf{\% Errs}} & \textbf{10} & \textbf{9} & \textbf{8} & \multicolumn{1}{l}{\textbf{$<$8}} & \multicolumn{1}{l}{\textbf{\% Errs}} \\

\midrule

\multirow{7}{*}{\colorbox{myblue}{LF}} & \coherence & 91 & 1 & 2 & 3 & 75 & 0.02 & 1 & 1 & 2 & 87 & 0.01 \\
 & \comprehensive & 90 & 1 & 33 & 23 & 20 & 0.01 & 13 & 29 & 26 & 21 & 0.14 \\
 & \consistency & 84 & 2 & 41 & 19 & 6 & 0.03 & 5 & 48 & 18 & 12 & 0.06 \\
 & \grammar & 92 & 1 & 55 & 11 & 5 & 0.01 & 31 & 38 & 17 & 5 & 0.34 \\
 & \chronology& 71 & 3 & 34 & 13 & 1 & 0.06 & 15 & 41 & 12 & 3 & 0.21 \\
 & \spelling & 100 & 4 & 52 & 9 & 4 & 0.06 & 65 & 28 & 5 & 1 & 0.66 \\
 \rowcolor{myblue}
 & \textbf{\total} & \textbf{528} & \textbf{12} & \textbf{217} & \textbf{78} & \textbf{111} & 0.03 & \textbf{130} & \textbf{185} & \textbf{80} & \textbf{129} & 0.25 \\
 \midrule
 
\multirow{7}{*}{\colorbox{mygreen}{F}} & \contextual & 94 & 0 & 49 & 19 & 19 & 0.00 & 4 & 34 & 24 & 29 & 0.04 \\
 & \entity & 87 & 0 & 50 & 17 & 16 & 0.00 & 7 & 29 & 26 & 20 & 0.08 \\
 & \incorrect& 68 & 0 & 38 & 18 & 9 & 0.00 & 2 & 31 & 20 & 13 & 0.03 \\
 & \numerr & 74 & 2 & 53 & 10 & 4 & 0.03 & 3 & 34 & 22 & 9 & 0.04 \\
 & \opposite & 91 & 0 & 37 & 19 & 30 & 0.00 & 4 & 18 & 30 & 38 & 0.04 \\
 & \remove & 69 & 4 & 29 & 22 & 13 & 0.06 & 13 & 18 & 26 & 12 & 0.19 \\
 \rowcolor{mygreen}
 & \textbf{\total} & \textbf{483} & \textbf{6} & \textbf{256} & \textbf{105} & \textbf{91} & 0.01 & \textbf{33} & \textbf{164} & \textbf{148} & \textbf{121} & 0.07 \\
 \midrule
 
\multirow{6}{*}{\colorbox{myorange}{IF}} & \assumptions & 81 & 0 & 31 & 23 & 19 & 0.00 & 0 & 12 & 20 & 48 & 0.00 \\
 & \doless & 100 & 1 & 23 & 35 & 32 & 0.01 & 26 & 15 & 27 & 32 & 0.26 \\
 & \domore & 50 & 1 & 31 & 15 & 1 & 0.02 & 1 & 13 & 28 & 7 & 0.02 \\
 & \format & 99 & 11 & 29 & 14 & 16 & 0.16 & 32 & 17 & 15 & 33 & 0.33 \\
 & \sequence & 49 & 5 & 28 & 13 & 0 & 0.11 & 6 & 26 & 14 & 3 & 0.12 \\
 \rowcolor{myorange}
 & \textbf{\total} & \textbf{379} & \textbf{18} & \textbf{142} & \textbf{100} & \textbf{68} & 0.05 & \textbf{65} & \textbf{83} & \textbf{104} & \textbf{123} & 0.17 \\
 \midrule
 
\multirow{6}{*}{\colorbox{mymagenta}{R}} & \calculation & 149 & 10 & 35 & 41 & 49 & 0.07 & 5 & 17 & 36 & 82 & 0.03 \\
 & \copying & 83 & 2 & 16 & 23 & 36 & 0.03 & 3 & 13 & 14 & 52 & 0.04 \\
 & \final & 97 & 0 & 20 & 56 & 13 & 0.00 & 0 & 28 & 44 & 23 & 0.00 \\
 & \units & 77 & 2 & 22 & 11 & 34 & 0.03 & 3 & 17 & 12 & 45 & 0.04 \\
 & \formula & 88 & 7 & 26 & 25 & 25 & 0.08 & 2 & 12 & 20 & 53 & 0.02 \\
 \rowcolor{mymagenta}
 & \textbf{\total} & \textbf{494} & \textbf{21} & \textbf{119} & \textbf{156} & \textbf{157} & 0.05 & \textbf{13} & \textbf{87} & \textbf{126} & \textbf{255} & 0.03 \\
 \bottomrule
\end{tabular}
\caption{Results from evaluating \fbi~ using the \textbf{Reference}-\llama,\claude~and \gemini~evaluators. An error is said to be detected if the evaluator gives a perfect score of 10 to the perturbed answer. \textbf{10} indicates the number of times the evaluator has given the score of 10, \textbf{9} for the score of 9, \textbf{8} for the score of 8 and \textbf{$<$8} for scores less than 8.}
\label{tab:apx}
\end{table*}
\endgroup

% -------------- justification tables -----------------

\begin{table*}[]
\centering
\small
\begin{tabular}{clccccc}
\toprule
\multicolumn{1}{l}{} & \textbf{Perturbation Type} & \multicolumn{1}{l}{\textbf{\# Errs}} & \multicolumn{1}{c}{\textbf{\begin{tabular}[c]{@{}l@{}}Detected\\ Errors\end{tabular}}} & \multicolumn{1}{c}{\textbf{\begin{tabular}[c]{@{}l@{}}Undetected\\ Errors\end{tabular}}} & \multicolumn{1}{c}{\textbf{\begin{tabular}[c]{@{}l@{}}Detected in \\ Explanation\end{tabular}}} & \multicolumn{1}{c}{\textbf{\begin{tabular}[c]{@{}c@{}}\% Undetected\\ Errors\end{tabular}}} \\
\midrule
\multirow{7}{*}{\colorbox{myblue}{LF}} & \coherence & 91 & 82 & 9 &  1 & \textbf{0.09} \\
 & \comprehensive & 90 & 30 & 60 & 5 & \textbf{0.61} \\
 & \consistency & 84 & 35 & 49 &  7 & \textbf{0.50} \\
 & \grammar & 92 & 40 & 52 & 9 & \textbf{0.47} \\
 & \chronology & 71 & 18 & 53  & 3 & \textbf{0.70} \\
 & \spelling & 100 & 20 & 80  & 11 & \textbf{0.69} \\
 \rowcolor{myblue}
 & \textbf{\total} & \textbf{528} & \textbf{225} & \textbf{303} & \textbf{36} & \textbf{0.51} \\
 \midrule
 
\multirow{7}{*}{\colorbox{mygreen}{F}} & \contextual & 94 & 45 & 48 & 5 & \textbf{0.47} \\
 & \entity & 87 & 43 & 44 &  3 & \textbf{0.47} \\
 & \incorrect & 68 & 29 & 38 &  4 & \textbf{0.51} \\
 & \numerr & 74 & 30 & 44 &  3 & \textbf{0.55} \\
 & \opposite & 91 & 48 & 42 &  6 & \textbf{0.41} \\
 & \remove & 69 & 25 & 44 &  0 & \textbf{0.64} \\
 \rowcolor{mygreen}
 & \textbf{\total} & \textbf{483} & \textbf{220} & \textbf{260} & \textbf{21} & \textbf{0.50} \\
 \midrule
 
\multirow{6}{*}{\colorbox{myorange}{IF}} & \assumptions & 81 & 12 & 69 &  7 & \textbf{0.77} \\
 & \doless & 100 & 57 & 43  & 6 & \textbf{0.37} \\
 & \domore & 50 & 31 & 19 & 12 & \textbf{0.14} \\
 & \format & 99 & 41 & 57 & 10 & \textbf{0.48} \\
 & \sequence & 49 & 20 & 29  & 1 & \textbf{0.57} \\
 \rowcolor{myorange}
 & \textbf{\total} & \textbf{379} & \textbf{161} & \textbf{217} & \textbf{36} & \textbf{0.48} \\
 \midrule
 
\multirow{6}{*}{\colorbox{mymagenta}{R}} & \calculation & 149 & 112 & 34  & 15 & \textbf{0.15} \\
 & \copying & 83 & 69 & 12 & 3 & \textbf{0.13} \\
 & \final & 97 & 53 & 43  & 16 & \textbf{0.29} \\
 & \units & 77 & 60 & 16 & 7 & \textbf{0.13} \\
 & \formula & 88 & 66 & 19  & 6 & \textbf{0.18} \\
 \rowcolor{mymagenta}
 & \textbf{\total} & \textbf{494} & \textbf{360} & \textbf{124}  & \textbf{47} & \textbf{0.18} \\
 \bottomrule
\end{tabular}
\caption{Results from looking at the explanation of the \textbf{Vanilla} evaluator to determine the presence of the error in the response. Detected in Explanation shows the number of ``additional'' errors detected by looking at the explanation in addition to the score. }
\end{table*}

\begin{table*}[]
\centering
\small
\begin{tabular}{clccccc}
\toprule
\multicolumn{1}{l}{} & \textbf{Perturbation Type} & \multicolumn{1}{l}{\textbf{\# Errs}} & \multicolumn{1}{c}{\textbf{\begin{tabular}[c]{@{}l@{}}Detected\\ Errors\end{tabular}}} & \multicolumn{1}{c}{\textbf{\begin{tabular}[c]{@{}l@{}}Undetected\\ Errors\end{tabular}}}  & \multicolumn{1}{c}{\textbf{\begin{tabular}[c]{@{}l@{}}Detected in \\ Justification\end{tabular}}} & \multicolumn{1}{c}{\textbf{\begin{tabular}[c]{@{}c@{}}\% Undetected\\ Errors\end{tabular}}} \\
\midrule

\multirow{7}{*}{\colorbox{myblue}{LF}} & \coherence & 91 & 58 & 33  & 1 & \textbf{0.35} \\
 & \comprehensive & 90 & 1 & 89  & 5 & \textbf{0.93} \\
 & \consistency & 84 & 8 & 76  & 3 & \textbf{0.87} \\
 & \grammar & 92 & 17 & 75  & 7 & \textbf{0.74} \\
 & \chronology & 71 & 0 & 71  & 9 & \textbf{0.87} \\
 & \spelling & 100 & 6 & 94  & 3 & \textbf{0.91} \\
 \rowcolor{myblue}
 & \textbf{\total} & \textbf{528} & \textbf{90} & \textbf{438}  & \textbf{28} & \textbf{0.78} \\
 \midrule
 
\multirow{7}{*}{\colorbox{mygreen}{F}} & \contextual & 94 & 29 & 65  & 23 & \textbf{0.45} \\
 & \entity & 87 & 30 & 57  & 15 & \textbf{0.48} \\
 & \incorrect & 68 & 17 & 51  & 14 & \textbf{0.54} \\
 & \numerr & 74 & 18 & 56  & 16 & \textbf{0.54} \\
 & \opposite & 91 & 32 & 59  & 23 & \textbf{0.40} \\
 & \remove & 69 & 1 & 68  & 20 & \textbf{0.70} \\
 \rowcolor{mygreen}
 & \textbf{\total} & \textbf{483} & \textbf{127} & \textbf{356}  & \textbf{111} & \textbf{0.51} \\
 \midrule
 
\multirow{6}{*}{\colorbox{myorange}{IF}} & \assumptions & 81 & 5 & 76  & 8 & \textbf{0.84} \\
 & \doless & 100 & 20 & 80  & 0 & \textbf{0.80} \\
 & \domore & 50 & 40 & 10  & 6 & \textbf{0.08} \\
 & \format & 99 & 25 & 74  & 12 & \textbf{0.63} \\
 & \sequence & 49 & 5 & 44  & 16 & \textbf{0.57} \\
 \rowcolor{myorange}
 & \textbf{\total} & \textbf{379} & \textbf{95} & \textbf{284}  & \textbf{42} & \textbf{0.64} \\
 \midrule
 
\multirow{6}{*}{\colorbox{mymagenta}{R}} & \calculation & 149 & 100 & 49  & 9 & \textbf{0.27} \\
 & \copying & 83 & 57 & 26  & 9 & \textbf{0.20} \\
 & \final & 97 & 46 & 51  & 7 & \textbf{0.45} \\
 & \units & 77 & 42 & 35  & 7 & \textbf{0.36} \\
 & \formula & 88 & 63 & 25  & 6 & \textbf{0.22} \\
 \rowcolor{mymagenta}
 & \textbf{\total} & \textbf{494} & \textbf{308} & \textbf{186}  & \textbf{38} & \textbf{0.30} \\
 \bottomrule
\end{tabular}
\caption{Results from looking at the explanation of the \textbf{Axis} evaluator to determine the presence of the error in the response. Detected in Explanation shows the number of ``additional'' errors detected by looking at the explanation in addition to the score.}
\end{table*}

\begin{table*}[]
\centering
\small
\begin{tabular}{clccccc}
\toprule
\multicolumn{1}{l}{} & \textbf{Perturbation Type} & \multicolumn{1}{l}{\textbf{\# Errs}} & \multicolumn{1}{c}{\textbf{\begin{tabular}[c]{@{}l@{}}Detected\\ Errors\end{tabular}}} & \multicolumn{1}{c}{\textbf{\begin{tabular}[c]{@{}l@{}}Undetected\\ Errors\end{tabular}}} & \multicolumn{1}{c}{\textbf{\begin{tabular}[c]{@{}l@{}}Detected in \\ Justification\end{tabular}}} & \multicolumn{1}{c}{\textbf{\begin{tabular}[c]{@{}c@{}}\% Undetected\\ Errors\end{tabular}}} \\
\midrule

\multirow{7}{*}{\colorbox{myblue}{LF}} & \coherence & 91 & 47 & 44  & 2 & \textbf{0.46} \\
 & \comprehensive & 90 & 2 & 88  & 5 & \textbf{0.92} \\
 & \consistency & 84 & 11 & 73  & 6 & \textbf{0.80} \\
 & \grammar & 92 & 15 & 77  & 6 & \textbf{0.77} \\
 & \chronology & 71 & 0 & 71  & 5 & \textbf{0.93} \\
 & \spelling & 100 & 4 & 96  & 8 & \textbf{0.88} \\
 \rowcolor{myblue}
 & \textbf{\total} & \textbf{528} & \textbf{79} & \textbf{449}  & \textbf{32} & \textbf{0.79} \\
 \midrule
 
\multirow{7}{*}{\colorbox{mygreen}{F}} & \contextual & 94 & 34 & 60  & 3 & \textbf{0.61} \\
 & \entity & 87 & 29 & 58  & 3 & \textbf{0.63} \\
 & \incorrect & 68 & 18 & 50  & 2 & \textbf{0.71} \\
 & \numerr & 74 & 17 & 57  & 7 & \textbf{0.68} \\
 & \opposite & 91 & 32 & 59  & 6 & \textbf{0.58} \\
 & \remove & 69 & 1 & 68  & 10 & \textbf{0.84} \\
 \rowcolor{mygreen}
 & \textbf{\total} & \textbf{483} & \textbf{131} & \textbf{352}  & \textbf{31} & \textbf{0.66} \\
 \midrule
 
\multirow{6}{*}{\colorbox{myorange}{IF}} & \assumptions & 81 & 1 & 80  & 1 & \textbf{0.98} \\
 & \doless & 100 & 8 & 92  & 8 & \textbf{0.84} \\
 & \domore & 50 & 39 & 11  & 2 & \textbf{0.18} \\
 & \format & 99 & 26 & 73  & 14 & \textbf{0.60} \\
 & \sequence & 49 & 0 & 49  & 5 & \textbf{0.90} \\
 \rowcolor{myorange}
 & \textbf{\total} & \textbf{379} & \textbf{74} & \textbf{305}  & \textbf{30} & \textbf{0.73} \\
 \midrule
 
\multirow{6}{*}{\colorbox{mymagenta}{R}} & \calculation & 149 & 102  & 47 & 10 & \textbf{0.25} \\
 & \copying & 83 & 64 & 19  & 3 & \textbf{0.19} \\
 & \final & 97 & 49 & 48  & 9 & \textbf{0.40} \\
 & \units & 77 & 56 & 21  & 4 & \textbf{0.22} \\
 & \formula & 88 & 61 & 27  & 13 & \textbf{0.16} \\
 \rowcolor{mymagenta}
 & \textbf{\total} & \textbf{494} & \textbf{332} & \textbf{162}  & \textbf{39} & \textbf{0.25} \\
 \bottomrule
\end{tabular}
\caption{Results from looking at the explanation of the \textbf{Rubrics} evaluator to determine the presence of the error in the response. Detected in Explanation shows the number of ``additional'' errors detected by looking at the explanation in addition to the score.}
\end{table*}

\begin{table*}[]
\centering
\small
\begin{tabular}{clccccc}
\toprule
\multicolumn{1}{l}{} & \textbf{Perturbation Type} & \multicolumn{1}{l}{\textbf{\# Errs}} & \multicolumn{1}{c}{\textbf{\begin{tabular}[c]{@{}l@{}}Detected\\ Errors\end{tabular}}} & \multicolumn{1}{c}{\textbf{\begin{tabular}[c]{@{}l@{}}Undetected\\ Errors\end{tabular}}} & \multicolumn{1}{c}{\textbf{\begin{tabular}[c]{@{}l@{}}Detected in \\ Justification\end{tabular}}} & \multicolumn{1}{c}{\textbf{\begin{tabular}[c]{@{}c@{}}\% Undetected\\ Errors\end{tabular}}} \\
\midrule

\multirow{7}{*}{\colorbox{myblue}{LF}} & \coherence & 91 & 45 & 46  & 0 & \textbf{0.51} \\
 & \comprehensive & 90 & 0 & 90  & 11 & \textbf{0.88} \\
 & \consistency & 84 & 6 & 78  & 8 & \textbf{0.83} \\
 & \grammar & 92 & 16 & 76  & 5 & \textbf{0.77} \\
 & \chronology & 71 & 0 & 71  & 12 & \textbf{0.83} \\
 & \spelling & 100 & 7 & 93  & 6 & \textbf{0.87} \\
 \rowcolor{myblue}
 & \textbf{\total} & \textbf{528} & \textbf{74} & \textbf{454}  & \textbf{42} & \textbf{0.78} \\
 \midrule
 
\multirow{7}{*}{\colorbox{mygreen}{F}} & \contextual & 94 & 28 & 66  & 19 & \textbf{0.50} \\
 & \entity & 87 & 27 & 60  & 9 & \textbf{0.59} \\
 & \incorrect & 68 & 15 & 53  & 10 & \textbf{0.63} \\
 & \numerr & 74 & 15 & 59  & 12 & \textbf{0.64} \\
 & \opposite & 91 & 28 & 63  & 12 & \textbf{0.56} \\
 & \remove & 69 & 1 & 68  & 16 & \textbf{0.75} \\
 \rowcolor{mygreen}
 & \textbf{\total} & \textbf{483} & \textbf{114} & \textbf{369}  & \textbf{78} & \textbf{0.60} \\
 \midrule
 
\multirow{6}{*}{\colorbox{myorange}{IF}} & \assumptions & 81 & 2 & 79  & 6 & \textbf{0.90} \\
 & \doless & 100 & 17 & 83  & 9 & \textbf{0.74} \\
 & \domore & 50 & 39 & 11  & 1 & \textbf{0.20} \\
 & \format & 99 & 24 & 75  & 14 & \textbf{0.62} \\
 & \sequence & 49 & 4 & 45  & 5 & \textbf{0.82} \\
 \rowcolor{myorange}
 & \textbf{\total} & \textbf{379} & \textbf{86} & \textbf{293}  & \textbf{35} & \textbf{0.68} \\
 \midrule
 
\multirow{6}{*}{\colorbox{mymagenta}{R}} & \calculation & 149 & 97 & 52  & 14 & \textbf{0.26} \\
 & \copying & 83 & 58 & 25  & 7 & \textbf{0.22} \\
 & \final & 97 & 48 & 49  & 12 & \textbf{0.38} \\
 & \units & 77 & 44 & 33  & 7 & \textbf{0.34} \\
 & \formula & 88 & 63 & 25  & 9 & \textbf{0.18} \\
 \rowcolor{mymagenta}
 & \textbf{\total} & \textbf{494} & \textbf{310} & \textbf{184} & \textbf{49} & \textbf{0.27} \\
 \bottomrule
\end{tabular}
\caption{Results from looking at the explanation of the \textbf{Axis+Rubrics} evaluator to determine the presence of the error in the response. Detected in Explanation shows the number of ``additional'' errors detected by looking at the explanation in addition to the score.}
\end{table*}

%% file: main.bbl
\begin{thebibliography}{52}
\providecommand{\natexlab}[1]{#1}

\bibitem[{Anthropic(2024)}]{claude}
Anthropic. 2024.
\newblock Introducing the next generation of claude.
\newblock \url{https://www.anthropic.com/news/claude-3-family}.
\newblock Accessed: 2024-06-14.

\bibitem[{Chan et~al.(2023)Chan, Chen, Su, Yu, Xue, Zhang, Fu, and Liu}]{chateval}
Chi{-}Min Chan, Weize Chen, Yusheng Su, Jianxuan Yu, Wei Xue, Shanghang Zhang, Jie Fu, and Zhiyuan Liu. 2023.
\newblock \href {https://doi.org/10.48550/ARXIV.2308.07201} {Chateval: Towards better llm-based evaluators through multi-agent debate}.
\newblock \emph{CoRR}, abs/2308.07201.

\bibitem[{Chen et~al.(2023)Chen, Wang, Jiang, Shi, and Xu}]{DBLP:journals/corr/abs-2304-00723}
Yi~Chen, Rui Wang, Haiyun Jiang, Shuming Shi, and Ruifeng Xu. 2023.
\newblock \href {https://doi.org/10.48550/ARXIV.2304.00723} {Exploring the use of large language models for reference-free text quality evaluation: {A} preliminary empirical study}.
\newblock \emph{CoRR}, abs/2304.00723.

\bibitem[{Chen et~al.(2024)Chen, Zhang, Luo, D'Haro, Tan, and Li}]{chen2024unveiling}
Yiming Chen, Chen Zhang, Danqing Luo, Luis~Fernando D'Haro, Robby~T. Tan, and Haizhou Li. 2024.
\newblock Unveiling the achilles' heel of nlg evaluators: A unified adversarial framework driven by large language models.
\newblock \emph{arXiv preprint arXiv: 2405.14646}.

\bibitem[{Chiang and yi~Lee(2023)}]{llms-alt-humans}
Cheng-Han Chiang and Hung yi~Lee. 2023.
\newblock \href {https://doi.org/10.48550/arXiv.2305.01937} {Can large language models be an alternative to human evaluations?}
\newblock \emph{Annual Meeting of the Association for Computational Linguistics}.

\bibitem[{Chiang and Lee(2023)}]{DBLP:conf/acl/ChiangL23}
David~Cheng{-}Han Chiang and Hung{-}yi Lee. 2023.
\newblock \href {https://doi.org/10.18653/V1/2023.ACL-LONG.870} {Can large language models be an alternative to human evaluations?}
\newblock In \emph{Proceedings of the 61st Annual Meeting of the Association for Computational Linguistics (Volume 1: Long Papers), {ACL} 2023, Toronto, Canada, July 9-14, 2023}, pages 15607--15631. Association for Computational Linguistics.

\bibitem[{Cobbe et~al.(2021)Cobbe, Kosaraju, Bavarian, Chen, Jun, Kaiser, Plappert, Tworek, Hilton, Nakano, Hesse, and Schulman}]{gsm8k}
Karl Cobbe, Vineet Kosaraju, Mohammad Bavarian, Mark Chen, Heewoo Jun, Lukasz Kaiser, Matthias Plappert, Jerry Tworek, Jacob Hilton, Reiichiro Nakano, Christopher Hesse, and John Schulman. 2021.
\newblock \href {https://arxiv.org/abs/2110.14168} {Training verifiers to solve math word problems}.
\newblock \emph{CoRR}, abs/2110.14168.

\bibitem[{Devlin et~al.(2019)Devlin, Chang, Lee, and Toutanova}]{devlin-etal-2019-bert}
Jacob Devlin, Ming-Wei Chang, Kenton Lee, and Kristina Toutanova. 2019.
\newblock \href {https://doi.org/10.18653/v1/N19-1423} {{BERT}: Pre-training of deep bidirectional transformers for language understanding}.
\newblock In \emph{Proceedings of the 2019 Conference of the North {A}merican Chapter of the Association for Computational Linguistics: Human Language Technologies, Volume 1 (Long and Short Papers)}, pages 4171--4186, Minneapolis, Minnesota. Association for Computational Linguistics.

\bibitem[{Ding et~al.(2023)Ding, Chen, Xu, Qin, Hu, Liu, Sun, and Zhou}]{ultrachat}
Ning Ding, Yulin Chen, Bokai Xu, Yujia Qin, Shengding Hu, Zhiyuan Liu, Maosong Sun, and Bowen Zhou. 2023.
\newblock \href {https://aclanthology.org/2023.emnlp-main.183} {Enhancing chat language models by scaling high-quality instructional conversations}.
\newblock In \emph{Proceedings of the 2023 Conference on Empirical Methods in Natural Language Processing, {EMNLP} 2023, Singapore, December 6-10, 2023}, pages 3029--3051. Association for Computational Linguistics.

\bibitem[{Dubois et~al.(2023)Dubois, Li, Taori, Zhang, Gulrajani, Ba, Guestrin, Liang, and Hashimoto}]{AlpacaFarm}
Yann Dubois, Chen~Xuechen Li, Rohan Taori, Tianyi Zhang, Ishaan Gulrajani, Jimmy Ba, Carlos Guestrin, Percy Liang, and Tatsunori~B. Hashimoto. 2023.
\newblock \href {http://papers.nips.cc/paper\_files/paper/2023/hash/5fc47800ee5b30b8777fdd30abcaaf3b-Abstract-Conference.html} {Alpacafarm: {A} simulation framework for methods that learn from human feedback}.
\newblock In \emph{Advances in Neural Information Processing Systems 36: Annual Conference on Neural Information Processing Systems 2023, NeurIPS 2023, New Orleans, LA, USA, December 10 - 16, 2023}.

\bibitem[{Fu et~al.(2023)Fu, Ng, Jiang, and Liu}]{fu2023gptscore}
Jinlan Fu, See-Kiong Ng, Zhengbao Jiang, and Pengfei Liu. 2023.
\newblock Gptscore: Evaluate as you desire.
\newblock \emph{arXiv preprint arXiv: 2302.04166}.

\bibitem[{Hada et~al.(2024)Hada, Gumma, Ahmed, Bali, and Sitaram}]{hada2024metal}
Rishav Hada, Varun Gumma, Mohamed Ahmed, Kalika Bali, and Sunayana Sitaram. 2024.
\newblock Metal: Towards multilingual meta-evaluation.
\newblock \emph{arXiv preprint arXiv: 2404.01667}.

\bibitem[{Hada et~al.(2023)Hada, Gumma, de~Wynter, Diddee, Ahmed, Choudhury, Bali, and Sitaram}]{hada2023large}
Rishav Hada, Varun Gumma, Adrian de~Wynter, Harshita Diddee, Mohamed Ahmed, M.~Choudhury, Kalika Bali, and Sunayana Sitaram. 2023.
\newblock \href {https://doi.org/10.48550/arXiv.2309.07462} {Are large language model-based evaluators the solution to scaling up multilingual evaluation?}
\newblock \emph{FINDINGS}.

\bibitem[{Hasanbeig et~al.(2023)Hasanbeig, Sharma, Betthauser, Frujeri, and Momennejad}]{hasanbeig2023allure}
Hosein Hasanbeig, Hiteshi Sharma, Leo Betthauser, Felipe~Vieira Frujeri, and Ida Momennejad. 2023.
\newblock Allure: Auditing and improving llm-based evaluation of text using iterative in-context-learning.
\newblock \emph{arXiv preprint arXiv: 2309.13701}.

\bibitem[{He et~al.(2023)He, Zhang, Wang, Kumar, Cho, Glass, and Tsvetkov}]{he-etal-2023-blind}
Tianxing He, Jingyu Zhang, Tianle Wang, Sachin Kumar, Kyunghyun Cho, James Glass, and Yulia Tsvetkov. 2023.
\newblock \href {https://doi.org/10.18653/v1/2023.acl-long.674} {On the blind spots of model-based evaluation metrics for text generation}.
\newblock In \emph{Proceedings of the 61st Annual Meeting of the Association for Computational Linguistics (Volume 1: Long Papers)}, pages 12067--12097, Toronto, Canada. Association for Computational Linguistics.

\bibitem[{Hendrycks et~al.(2021)Hendrycks, Burns, Kadavath, Arora, Basart, Tang, Song, and Steinhardt}]{math}
Dan Hendrycks, Collin Burns, Saurav Kadavath, Akul Arora, Steven Basart, Eric Tang, Dawn Song, and Jacob Steinhardt. 2021.
\newblock \href {https://datasets-benchmarks-proceedings.neurips.cc/paper/2021/hash/be83ab3ecd0db773eb2dc1b0a17836a1-Abstract-round2.html} {Measuring mathematical problem solving with the {MATH} dataset}.
\newblock In \emph{Proceedings of the Neural Information Processing Systems Track on Datasets and Benchmarks 1, NeurIPS Datasets and Benchmarks 2021, December 2021, virtual}.

\bibitem[{Hu et~al.(2024)Hu, Chen, Li, Guo, Wen, Yu, and Guo}]{hu2024towards}
Xuming Hu, Junzhe Chen, Xiaochuan Li, Yufei Guo, Lijie Wen, Philip~S. Yu, and Zhijiang Guo. 2024.
\newblock \href {https://openreview.net/forum?id=9OevMUdods} {Towards understanding factual knowledge of large language models}.
\newblock In \emph{The Twelfth International Conference on Learning Representations}.

\bibitem[{Kamoi et~al.(2024)Kamoi, Das, Lou, Ahn, Zhao, Lu, Zhang, Zhang, Zhang, Vummanthala, Dave, Qin, Cohan, Yin, and Zhang}]{kamoi2024evaluating}
Ryo Kamoi, Sarkar Snigdha~Sarathi Das, Renze Lou, Jihyun~Janice Ahn, Yilun Zhao, Xiaoxin Lu, Nan Zhang, Yusen Zhang, Ranran~Haoran Zhang, Sujeeth~Reddy Vummanthala, Salika Dave, Shaobo Qin, Arman Cohan, Wenpeng Yin, and Rui Zhang. 2024.
\newblock Evaluating llms at detecting errors in llm responses.
\newblock \emph{arXiv preprint arXiv: 2404.03602}.

\bibitem[{Kim et~al.(2023)Kim, Shin, Cho, Jang, Longpre, Lee, Yun, Shin, Kim, Thorne, and Seo}]{prometheus}
Seungone Kim, Jamin Shin, Yejin Cho, Joel Jang, Shayne Longpre, Hwaran Lee, Sangdoo Yun, Seongjin Shin, Sungdong Kim, James Thorne, and Minjoon Seo. 2023.
\newblock \href {https://doi.org/10.48550/ARXIV.2310.08491} {Prometheus: Inducing fine-grained evaluation capability in language models}.
\newblock \emph{CoRR}, abs/2310.08491.

\bibitem[{Kim et~al.(2024{\natexlab{a}})Kim, Suk, Cho, Longpre, Kim, Yoon, Son, Cho, Shafayat, Baek, Park, Hwang, Jo, Cho, Shin, Lee, Oh, Lee, Ho, Joo, Ko, Lee, Chae, Shin, Jang, Ye, Lin, Welleck, Neubig, Lee, Lee, and Seo}]{Kim2024TheBB}
Seungone Kim, Juyoung Suk, Ji~Yong Cho, Shayne Longpre, Chaeeun Kim, Dongkeun Yoon, Guijin Son, Yejin Cho, Sheikh Shafayat, Jinheon Baek, Sue~Hyun Park, Hyeonbin Hwang, Jinkyung Jo, Hyowon Cho, Haebin Shin, Seongyun Lee, Hanseok Oh, Noah Lee, Namgyu Ho, Se~June Joo, Miyoung Ko, Yoonjoo Lee, Hyungjoo Chae, Jamin Shin, Joel Jang, Seonghyeon Ye, Bill~Yuchen Lin, Sean Welleck, Graham Neubig, Moontae Lee, Kyungjae Lee, and Minjoon Seo. 2024{\natexlab{a}}.
\newblock \href {https://api.semanticscholar.org/CorpusID:270371930} {The biggen bench: A principled benchmark for fine-grained evaluation of language models with language models}.

\bibitem[{Kim et~al.(2024{\natexlab{b}})Kim, Suk, Longpre, Lin, Shin, Welleck, Neubig, Lee, Lee, and Seo}]{kim2024prometheus2}
Seungone Kim, Juyoung Suk, Shayne Longpre, Bill~Yuchen Lin, Jamin Shin, Sean Welleck, Graham Neubig, Moontae Lee, Kyungjae Lee, and Minjoon Seo. 2024{\natexlab{b}}.
\newblock Prometheus 2: An open source language model specialized in evaluating other language models.
\newblock \emph{arXiv preprint arXiv: 2405.01535}.

\bibitem[{Kocmi and Federmann(2023)}]{gemba}
Tom Kocmi and C.~Federmann. 2023.
\newblock \href {https://doi.org/10.48550/arXiv.2302.14520} {Large language models are state-of-the-art evaluators of translation quality}.
\newblock \emph{European Association for Machine Translation Conferences/Workshops}.

\bibitem[{Li et~al.(2023)Li, Peng, He, and Yan}]{li2023evaluating}
Zekun Li, Baolin Peng, Pengcheng He, and Xifeng Yan. 2023.
\newblock Evaluating the instruction-following robustness of large language models to prompt injection.
\newblock \emph{arXiv preprint arXiv: 2308.10819}.

\bibitem[{Liu et~al.(2023)Liu, Iter, Xu, Wang, Xu, and Zhu}]{liu2023geval}
Yang Liu, Dan Iter, Yichong Xu, Shuo Wang, Ruochen Xu, and Chenguang Zhu. 2023.
\newblock \href {https://doi.org/10.48550/arXiv.2303.16634} {G-eval: Nlg evaluation using gpt-4 with better human alignment}.
\newblock \emph{Conference on Empirical Methods in Natural Language Processing}.

\bibitem[{Liusie et~al.(2023)Liusie, Manakul, and Gales}]{liusie2023llm}
Adian Liusie, Potsawee Manakul, and Mark J.~F. Gales. 2023.
\newblock Llm comparative assessment: Zero-shot nlg evaluation through pairwise comparisons using large language models.
\newblock \emph{arXiv preprint arXiv: 2307.07889}.

\bibitem[{Mathur et~al.(2020)Mathur, Baldwin, and Cohn}]{mathur-etal-2020-tangled}
Nitika Mathur, Timothy Baldwin, and Trevor Cohn. 2020.
\newblock \href {https://doi.org/10.18653/v1/2020.acl-main.448} {Tangled up in {BLEU}: Reevaluating the evaluation of automatic machine translation evaluation metrics}.
\newblock In \emph{Proceedings of the 58th Annual Meeting of the Association for Computational Linguistics}, pages 4984--4997, Online. Association for Computational Linguistics.

\bibitem[{Meta(2024)}]{llama3}
Meta. 2024.
\newblock Introducing meta llama 3: The most capable openly available llm to date.
\newblock \url{https://ai.meta.com/blog/meta-llama-3/}.
\newblock Accessed: 2024-06-14.

\bibitem[{Min et~al.(2023)Min, Krishna, Lyu, Lewis, tau Yih, Koh, Iyyer, Zettlemoyer, and Hajishirzi}]{factscore}
Sewon Min, Kalpesh Krishna, Xinxi Lyu, Mike Lewis, Wen tau Yih, Pang~Wei Koh, Mohit Iyyer, Luke Zettlemoyer, and Hannaneh Hajishirzi. 2023.
\newblock Factscore: Fine-grained atomic evaluation of factual precision in long form text generation.
\newblock \emph{arXiv preprint arXiv: 2305.14251}.

\bibitem[{Naismith et~al.(2023)Naismith, Mulcaire, and Burstein}]{naismith-etal-2023-automated}
Ben Naismith, Phoebe Mulcaire, and Jill Burstein. 2023.
\newblock \href {https://doi.org/10.18653/v1/2023.bea-1.32} {Automated evaluation of written discourse coherence using {GPT}-4}.
\newblock In \emph{Proceedings of the 18th Workshop on Innovative Use of NLP for Building Educational Applications (BEA 2023)}, pages 394--403, Toronto, Canada. Association for Computational Linguistics.

\bibitem[{Panickssery et~al.(2024)Panickssery, Bowman, and Feng}]{panickssery2024llm}
Arjun Panickssery, Samuel~R. Bowman, and Shi Feng. 2024.
\newblock Llm evaluators recognize and favor their own generations.
\newblock \emph{arXiv preprint arXiv: 2404.13076}.

\bibitem[{Ribeiro et~al.(2020)Ribeiro, Wu, Guestrin, and Singh}]{ribeiro-etal-2020-beyond}
Marco~Tulio Ribeiro, Tongshuang Wu, Carlos Guestrin, and Sameer Singh. 2020.
\newblock \href {https://doi.org/10.18653/v1/2020.acl-main.442} {Beyond accuracy: Behavioral testing of {NLP} models with {C}heck{L}ist}.
\newblock In \emph{Proceedings of the 58th Annual Meeting of the Association for Computational Linguistics}, pages 4902--4912, Online. Association for Computational Linguistics.

\bibitem[{Saha et~al.(2023)Saha, Levy, Celikyilmaz, Bansal, Weston, and Li}]{bsm}
Swarnadeep Saha, Omer Levy, Asli Celikyilmaz, Mohit Bansal, Jason Weston, and Xian Li. 2023.
\newblock \href {https://doi.org/10.48550/ARXIV.2310.15123} {Branch-solve-merge improves large language model evaluation and generation}.
\newblock \emph{CoRR}, abs/2310.15123.

\bibitem[{Sai et~al.(2021)Sai, Dixit, Sheth, Mohan, and Khapra}]{sai-etal-2021-perturbation}
Ananya~B. Sai, Tanay Dixit, Dev~Yashpal Sheth, Sreyas Mohan, and Mitesh~M. Khapra. 2021.
\newblock \href {https://doi.org/10.18653/v1/2021.emnlp-main.575} {Perturbation {C}heck{L}ists for evaluating {NLG} evaluation metrics}.
\newblock In \emph{Proceedings of the 2021 Conference on Empirical Methods in Natural Language Processing}, pages 7219--7234, Online and Punta Cana, Dominican Republic. Association for Computational Linguistics.

\bibitem[{Sai~B et~al.(2023)Sai~B, Dixit, Nagarajan, Kunchukuttan, Kumar, Khapra, and Dabre}]{indicmteval}
Ananya Sai~B, Tanay Dixit, Vignesh Nagarajan, Anoop Kunchukuttan, Pratyush Kumar, Mitesh~M. Khapra, and Raj Dabre. 2023.
\newblock \href {https://doi.org/10.18653/v1/2023.acl-long.795} {{I}ndic{MT} eval: A dataset to meta-evaluate machine translation metrics for {I}ndian languages}.
\newblock In \emph{Proceedings of the 61st Annual Meeting of the Association for Computational Linguistics (Volume 1: Long Papers)}, pages 14210--14228, Toronto, Canada. Association for Computational Linguistics.

\bibitem[{Shen et~al.(2023)Shen, Cheng, Nguyen, You, and Bing}]{shen-etal-2023-large}
Chenhui Shen, Liying Cheng, Xuan-Phi Nguyen, Yang You, and Lidong Bing. 2023.
\newblock \href {https://doi.org/10.18653/v1/2023.findings-emnlp.278} {Large language models are not yet human-level evaluators for abstractive summarization}.
\newblock In \emph{Findings of the Association for Computational Linguistics: EMNLP 2023}, pages 4215--4233, Singapore. Association for Computational Linguistics.

\bibitem[{Team et~al.(2024)Team, Reid, Savinov, Teplyashin, Dmitry, Lepikhin, Lillicrap, baptiste Alayrac, Soricut, Lazaridou, Firat, Schrittwieser, Antonoglou, Anil, Borgeaud, Dai, Millican, Dyer, Glaese, Sottiaux, Lee, Viola, Reynolds, Xu, Molloy, Chen, Isard, Barham, Hennigan, McIlroy, Johnson, Schalkwyk, Collins, Rutherford, Moreira, Ayoub, Goel, Meyer, Thornton, Yang, Michalewski, Abbas, Schucher, Anand, Ives, Keeling, Lenc, Haykal, Shakeri, Shyam, Chowdhery, Ring, Spencer, Sezener, Vilnis, Chang, Morioka, Tucker, Zheng, Woodman, Attaluri, Kocisky, Eltyshev, Chen, Chung, Selo, Brahma, Georgiev, Slone, Zhu, Lottes, Qiao, Caine, Riedel, Tomala, Chadwick, Love, Choy, Mittal, Houlsby, Tang, Lamm, Bai, Zhang, He, Cheng, Humphreys, Li, Brin, Cassirer, Miao, Zilka, Tobin, Xu, Proleev, Sohn, Magni, Hendricks, Gao, Ontanon, Bunyan, Byrd, Sharma, Zhang, Pinto, Sinha, Mehta, Jia, Caelles, Webson, Morris, Roelofs, Ding, Strudel, Xiong, Ritter, Dehghani, Chaabouni, Karmarkar, Lai, Mentzer, Xu, Li, Zhang, Paine,
  Goldin, Neyshabur, Baumli, Levskaya, Laskin, Jia, Rae, Xiao, He, Giordano, Yagati, Lespiau, Natsev, Ganapathy, Liu, Martins, Chen, Xu, Barnes, May, Vezer, Oh, Franko, Bridgers, Zhao, Wu, Mustafa, Sechrist, Parisotto, Pillai, Larkin, Gu, Sorokin, Krikun, Guseynov, Landon, Datta, Pritzel, Thacker, Yang, Hui, Hauth, Yeh, Barker, Mao-Jones, Austin, Sheahan, Schuh, Svensson, Jain, Ramasesh, Briukhov, Chung, von Glehn, Butterfield, Jhakra, Wiethoff, Frye, Grimstad, Changpinyo, Lan, Bortsova, Wu, Voigtlaender, Sainath, Gu, Smith, Hawkins, Cao, Besley, Srinivasan, Omernick, Gaffney, Surita, Burnell, Damoc, Ahn, Brock, Pajarskas, Petrushkina, Noury, Blanco, Swersky, Ahuja, Avrahami, Misra, de~Liedekerke, Iinuma, Polozov, York, van~den Driessche, Michel, Chiu, Blevins, Gleicher, Recasens, Rrustemi, Gribovskaya, Roy, Gworek, Arnold, Lee, Lee-Thorp, Maggioni, Piqueras, Badola, Vikram, Gonzalez, Baddepudi, Senter, Devlin, Qin, Azzam, Trebacz, Polacek, Krishnakumar, yiin Chang, Tung, Penchev, Joshi, Olszewska, Muir,
  Wirth, Hartman, Newlan, Kashem, Bolina, Dabir, van Amersfoort, Ahmed, Cobon-Kerr, Kamath, Hrafnkelsson, Hou, Mackinnon, Frechette, Noland, Si, Taropa, Li, Crone, Gulati, Cevey, Adler, Ma, Silver, Tokumine, Powell, Lee, Vodrahalli, Hassan, Mincu, Yang, Levine, Brennan, Wang, Hodkinson, Zhao, Lipschultz, Pope, Chang, Li, Shafey, Paganini, Douglas, Bohnet, Pardo, Odoom, Rosca, dos Santos, Soparkar, Guez, Hudson, Hansen, Asawaroengchai, Addanki, Yu, Stokowiec, Khan, Gilmer, Lee, Bostock, Rong, Caton, Pejman, Pavetic, Brown, Sharma, Lučić, Samuel, Djolonga, Mandhane, Sjösund, Buchatskaya, White, Clay, Jiang, Lim, Hemsley, Cankara, Labanowski, Cao, Steiner, Hashemi, Austin, Gergely, Blyth, Stanton, Shivakumar, Siddhant, Andreassen, Araya, Sethi, Shivanna, Hand, Bapna, Khodaei, Miech, Tanzer, Swing, Thakoor, Aroyo, Pan, Nado, Sygnowski, Winkler, Yu, Saleh, Maggiore, Bansal, Garcia, Kazemi, Patil, Dasgupta, Barr, Giang, Kagohara, Danihelka, Marathe, Feinberg, Elhawaty, Ghelani, Horgan, Miller, Walker, Tanburn,
  Tariq, Shrivastava, Xia, Wang, Chiu, Ashwood, Baatarsukh, Samangooei, Kaufman, Alcober, Stjerngren, Komarek, Tsihlas, Boral, Comanescu, Chen, Liu, Welty, Bloxwich, Chen, Sun, Feng, Mauger, Dotiwalla, Hellendoorn, Sharman, Zheng, Haridasan, Barth-Maron, Swanson, Rogozińska, Andreev, Rubenstein, Sang, Hurt, Elsayed, Wang, Lacey, Ilić, Zhao, Iwanicki, Lince, Chen, Lyu, Lebsack, Griffith, Gaba, Sandhu, Chen, Koop, Rajwar, Yeganeh, Chang, Zhu, Radpour, Davoodi, Lei, Xu, Toyama, Segal, Wicke, Lin, Bulanova, Badia, Rakićević, Sprechmann, Filos, Hou, Campos, Kassner, Sachan, Fortunato, Iwuanyanwu, Nikolaev, Lakshminarayanan, Jazayeri, Varadarajan, Tekur, Fritz, Khalman, Reitter, Dasgupta, Sarcar, Ornduff, Snaider, Huot, Jia, Kemp, Trdin, Vijayakumar, Kim, Angermueller, Lao, Liu, Zhang, Engel, Greene, White, Austin, Taylor, Ashraf, Liu, Georgaki, Cai, Kulizhskaya, Goenka, Saeta, Xu, Frank, de~Cesare, Robenek, Richardson, Alnahlawi, Yew, Ponnapalli, Tagliasacchi, Korchemniy, Kim, Li, Rosgen, Levin, Wiesner,
  Banzal, Srinivasan, Yu, Çağlar Ünlü, Reid, Tung, Finchelstein, Kumar, Elisseeff, Huang, Zhang, Aguilar, Giménez, Xia, Dousse, Gierke, Yates, Jalan, Li, Latorre-Chimoto, Nguyen, Durden, Kallakuri, Liu, Johnson, Tsai, Talbert, Liu, Neitz, Elkind, Selvi, Jasarevic, Soares, Cui, Wang, Wang, Ye, Kallarackal, Loher, Lam, Broder, Holtmann-Rice, Martin, Ramadhana, Shukla, Basu, Mohan, Fernando, Fiedel, Paterson, Li, Garg, Park, Choi, Wu, Singh, Zhang, Globerson, Yu, Carpenter, de~Chaumont~Quitry, Radebaugh, Lin, Tudor, Shroff, Garmon, Du, Vats, Lu, Iqbal, Yakubovich, Tripuraneni, Manyika, Qureshi, Hua, Ngani, Raad, Forbes, Stanway, Sundararajan, Ungureanu, Bishop, Li, Venkatraman, Li, Thornton, Scellato, Gupta, Wang, Tenney, Wu, Shenoy, Carvajal, Wright, Bariach, Xiao, Hawkins, Dalmia, Farabet, Valenzuela, Yuan, Agarwal, Chen, Kim, Hulse, Dukkipati, Paszke, Bolt, Choo, Beattie, Prendki, Vashisht, Santamaria-Fernandez, Cobo, Wilkiewicz, Madras, Elqursh, Uy, Ramirez, Harvey, Liechty, Zen, Seibert, Hu, Khorlin,
  Le, Aharoni, Li, Wang, Kumar, Casagrande, Hoover, Badawy, Soergel, Vnukov, Miecnikowski, Simsa, Kumar, Sellam, Vlasic, Daruki, Shabat, Zhang, Su, Zhang, Liu, Sun, Palmer, Ghaffarkhah, Xiong, Cotruta, Fink, Dixon, Sreevatsa, Goedeckemeyer, Dimitriev, Jafari, Crocker, FitzGerald, Kumar, Ghemawat, Philips, Liu, Liang, Sterneck, Repina, Wu, Knight, Georgiev, Lee, Askham, Chakladar, Louis, Crous, Cate, Petrova, Quinn, Owusu-Afriyie, Singhal, Wei, Kim, Vincent, Nasr, Choquette-Choo, Tojo, Lu, de~Las~Casas, Cheng, Bolukbasi, Lee, Fatehi, Ananthanarayanan, Patel, Kaed, Li, Belle, Chen, Konzelmann, Põder, Garg, Koverkathu, Brown, Dyer, Liu, Nova, Xu, Walton, Parrish, Epstein, McCarthy, Petrov, Hassabis, Kavukcuoglu, Dean, and Vinyals}]{team2024gemini}
Gemini Team, Machel Reid, Nikolay Savinov, Denis Teplyashin, Dmitry, Lepikhin, Timothy Lillicrap, Jean baptiste Alayrac, Radu Soricut, Angeliki Lazaridou, Orhan Firat, Julian Schrittwieser, Ioannis Antonoglou, Rohan Anil, Sebastian Borgeaud, Andrew Dai, Katie Millican, Ethan Dyer, Mia Glaese, Thibault Sottiaux, Benjamin Lee, Fabio Viola, Malcolm Reynolds, Yuanzhong Xu, James Molloy, Jilin Chen, Michael Isard, Paul Barham, Tom Hennigan, Ross McIlroy, Melvin Johnson, Johan Schalkwyk, Eli Collins, Eliza Rutherford, Erica Moreira, Kareem Ayoub, Megha Goel, Clemens Meyer, Gregory Thornton, Zhen Yang, Henryk Michalewski, Zaheer Abbas, Nathan Schucher, Ankesh Anand, Richard Ives, James Keeling, Karel Lenc, Salem Haykal, Siamak Shakeri, Pranav Shyam, Aakanksha Chowdhery, Roman Ring, Stephen Spencer, Eren Sezener, Luke Vilnis, Oscar Chang, Nobuyuki Morioka, George Tucker, Ce~Zheng, Oliver Woodman, Nithya Attaluri, Tomas Kocisky, Evgenii Eltyshev, Xi~Chen, Timothy Chung, Vittorio Selo, Siddhartha Brahma, Petko
  Georgiev, Ambrose Slone, Zhenkai Zhu, James Lottes, Siyuan Qiao, Ben Caine, Sebastian Riedel, Alex Tomala, Martin Chadwick, Juliette Love, Peter Choy, Sid Mittal, Neil Houlsby, Yunhao Tang, Matthew Lamm, Libin Bai, Qiao Zhang, Luheng He, Yong Cheng, Peter Humphreys, Yujia Li, Sergey Brin, Albin Cassirer, Yingjie Miao, Lukas Zilka, Taylor Tobin, Kelvin Xu, Lev Proleev, Daniel Sohn, Alberto Magni, Lisa~Anne Hendricks, Isabel Gao, Santiago Ontanon, Oskar Bunyan, Nathan Byrd, Abhanshu Sharma, Biao Zhang, Mario Pinto, Rishika Sinha, Harsh Mehta, Dawei Jia, Sergi Caelles, Albert Webson, Alex Morris, Becca Roelofs, Yifan Ding, Robin Strudel, Xuehan Xiong, Marvin Ritter, Mostafa Dehghani, Rahma Chaabouni, Abhijit Karmarkar, Guangda Lai, Fabian Mentzer, Bibo Xu, YaGuang Li, Yujing Zhang, Tom~Le Paine, Alex Goldin, Behnam Neyshabur, Kate Baumli, Anselm Levskaya, Michael Laskin, Wenhao Jia, Jack~W. Rae, Kefan Xiao, Antoine He, Skye Giordano, Lakshman Yagati, Jean-Baptiste Lespiau, Paul Natsev, Sanjay Ganapathy, Fangyu
  Liu, Danilo Martins, Nanxin Chen, Yunhan Xu, Megan Barnes, Rhys May, Arpi Vezer, Junhyuk Oh, Ken Franko, Sophie Bridgers, Ruizhe Zhao, Boxi Wu, Basil Mustafa, Sean Sechrist, Emilio Parisotto, Thanumalayan~Sankaranarayana Pillai, Chris Larkin, Chenjie Gu, Christina Sorokin, Maxim Krikun, Alexey Guseynov, Jessica Landon, Romina Datta, Alexander Pritzel, Phoebe Thacker, Fan Yang, Kevin Hui, Anja Hauth, Chih-Kuan Yeh, David Barker, Justin Mao-Jones, Sophia Austin, Hannah Sheahan, Parker Schuh, James Svensson, Rohan Jain, Vinay Ramasesh, Anton Briukhov, Da-Woon Chung, Tamara von Glehn, Christina Butterfield, Priya Jhakra, Matthew Wiethoff, Justin Frye, Jordan Grimstad, Beer Changpinyo, Charline~Le Lan, Anna Bortsova, Yonghui Wu, Paul Voigtlaender, Tara Sainath, Shane Gu, Charlotte Smith, Will Hawkins, Kris Cao, James Besley, Srivatsan Srinivasan, Mark Omernick, Colin Gaffney, Gabriela Surita, Ryan Burnell, Bogdan Damoc, Junwhan Ahn, Andrew Brock, Mantas Pajarskas, Anastasia Petrushkina, Seb Noury, Lorenzo
  Blanco, Kevin Swersky, Arun Ahuja, Thi Avrahami, Vedant Misra, Raoul de~Liedekerke, Mariko Iinuma, Alex Polozov, Sarah York, George van~den Driessche, Paul Michel, Justin Chiu, Rory Blevins, Zach Gleicher, Adrià Recasens, Alban Rrustemi, Elena Gribovskaya, Aurko Roy, Wiktor Gworek, Sébastien M.~R. Arnold, Lisa Lee, James Lee-Thorp, Marcello Maggioni, Enrique Piqueras, Kartikeya Badola, Sharad Vikram, Lucas Gonzalez, Anirudh Baddepudi, Evan Senter, Jacob Devlin, James Qin, Michael Azzam, Maja Trebacz, Martin Polacek, Kashyap Krishnakumar, Shuo yiin Chang, Matthew Tung, Ivo Penchev, Rishabh Joshi, Kate Olszewska, Carrie Muir, Mateo Wirth, Ale~Jakse Hartman, Josh Newlan, Sheleem Kashem, Vijay Bolina, Elahe Dabir, Joost van Amersfoort, Zafarali Ahmed, James Cobon-Kerr, Aishwarya Kamath, Arnar~Mar Hrafnkelsson, Le~Hou, Ian Mackinnon, Alexandre Frechette, Eric Noland, Xiance Si, Emanuel Taropa, Dong Li, Phil Crone, Anmol Gulati, Sébastien Cevey, Jonas Adler, Ada Ma, David Silver, Simon Tokumine, Richard
  Powell, Stephan Lee, Kiran Vodrahalli, Samer Hassan, Diana Mincu, Antoine Yang, Nir Levine, Jenny Brennan, Mingqiu Wang, Sarah Hodkinson, Jeffrey Zhao, Josh Lipschultz, Aedan Pope, Michael~B. Chang, Cheng Li, Laurent~El Shafey, Michela Paganini, Sholto Douglas, Bernd Bohnet, Fabio Pardo, Seth Odoom, Mihaela Rosca, Cicero~Nogueira dos Santos, Kedar Soparkar, Arthur Guez, Tom Hudson, Steven Hansen, Chulayuth Asawaroengchai, Ravi Addanki, Tianhe Yu, Wojciech Stokowiec, Mina Khan, Justin Gilmer, Jaehoon Lee, Carrie~Grimes Bostock, Keran Rong, Jonathan Caton, Pedram Pejman, Filip Pavetic, Geoff Brown, Vivek Sharma, Mario Lučić, Rajkumar Samuel, Josip Djolonga, Amol Mandhane, Lars~Lowe Sjösund, Elena Buchatskaya, Elspeth White, Natalie Clay, Jiepu Jiang, Hyeontaek Lim, Ross Hemsley, Zeyncep Cankara, Jane Labanowski, Nicola~De Cao, David Steiner, Sayed~Hadi Hashemi, Jacob Austin, Anita Gergely, Tim Blyth, Joe Stanton, Kaushik Shivakumar, Aditya Siddhant, Anders Andreassen, Carlos Araya, Nikhil Sethi, Rakesh
  Shivanna, Steven Hand, Ankur Bapna, Ali Khodaei, Antoine Miech, Garrett Tanzer, Andy Swing, Shantanu Thakoor, Lora Aroyo, Zhufeng Pan, Zachary Nado, Jakub Sygnowski, Stephanie Winkler, Dian Yu, Mohammad Saleh, Loren Maggiore, Yamini Bansal, Xavier Garcia, Mehran Kazemi, Piyush Patil, Ishita Dasgupta, Iain Barr, Minh Giang, Thais Kagohara, Ivo Danihelka, Amit Marathe, Vladimir Feinberg, Mohamed Elhawaty, Nimesh Ghelani, Dan Horgan, Helen Miller, Lexi Walker, Richard Tanburn, Mukarram Tariq, Disha Shrivastava, Fei Xia, Qingze Wang, Chung-Cheng Chiu, Zoe Ashwood, Khuslen Baatarsukh, Sina Samangooei, Raphaël~Lopez Kaufman, Fred Alcober, Axel Stjerngren, Paul Komarek, Katerina Tsihlas, Anudhyan Boral, Ramona Comanescu, Jeremy Chen, Ruibo Liu, Chris Welty, Dawn Bloxwich, Charlie Chen, Yanhua Sun, Fangxiaoyu Feng, Matthew Mauger, Xerxes Dotiwalla, Vincent Hellendoorn, Michael Sharman, Ivy Zheng, Krishna Haridasan, Gabe Barth-Maron, Craig Swanson, Dominika Rogozińska, Alek Andreev, Paul~Kishan Rubenstein, Ruoxin
  Sang, Dan Hurt, Gamaleldin Elsayed, Renshen Wang, Dave Lacey, Anastasija Ilić, Yao Zhao, Adam Iwanicki, Alejandro Lince, Alexander Chen, Christina Lyu, Carl Lebsack, Jordan Griffith, Meenu Gaba, Paramjit Sandhu, Phil Chen, Anna Koop, Ravi Rajwar, Soheil~Hassas Yeganeh, Solomon Chang, Rui Zhu, Soroush Radpour, Elnaz Davoodi, Ving~Ian Lei, Yang Xu, Daniel Toyama, Constant Segal, Martin Wicke, Hanzhao Lin, Anna Bulanova, Adrià~Puigdomènech Badia, Nemanja Rakićević, Pablo Sprechmann, Angelos Filos, Shaobo Hou, Víctor Campos, Nora Kassner, Devendra Sachan, Meire Fortunato, Chimezie Iwuanyanwu, Vitaly Nikolaev, Balaji Lakshminarayanan, Sadegh Jazayeri, Mani Varadarajan, Chetan Tekur, Doug Fritz, Misha Khalman, David Reitter, Kingshuk Dasgupta, Shourya Sarcar, Tina Ornduff, Javier Snaider, Fantine Huot, Johnson Jia, Rupert Kemp, Nejc Trdin, Anitha Vijayakumar, Lucy Kim, Christof Angermueller, Li~Lao, Tianqi Liu, Haibin Zhang, David Engel, Somer Greene, Anaïs White, Jessica Austin, Lilly Taylor, Shereen
  Ashraf, Dangyi Liu, Maria Georgaki, Irene Cai, Yana Kulizhskaya, Sonam Goenka, Brennan Saeta, Ying Xu, Christian Frank, Dario de~Cesare, Brona Robenek, Harry Richardson, Mahmoud Alnahlawi, Christopher Yew, Priya Ponnapalli, Marco Tagliasacchi, Alex Korchemniy, Yelin Kim, Dinghua Li, Bill Rosgen, Kyle Levin, Jeremy Wiesner, Praseem Banzal, Praveen Srinivasan, Hongkun Yu, Çağlar Ünlü, David Reid, Zora Tung, Daniel Finchelstein, Ravin Kumar, Andre Elisseeff, Jin Huang, Ming Zhang, Ricardo Aguilar, Mai Giménez, Jiawei Xia, Olivier Dousse, Willi Gierke, Damion Yates, Komal Jalan, Lu~Li, Eri Latorre-Chimoto, Duc~Dung Nguyen, Ken Durden, Praveen Kallakuri, Yaxin Liu, Matthew Johnson, Tomy Tsai, Alice Talbert, Jasmine Liu, Alexander Neitz, Chen Elkind, Marco Selvi, Mimi Jasarevic, Livio~Baldini Soares, Albert Cui, Pidong Wang, Alek~Wenjiao Wang, Xinyu Ye, Krystal Kallarackal, Lucia Loher, Hoi Lam, Josef Broder, Dan Holtmann-Rice, Nina Martin, Bramandia Ramadhana, Mrinal Shukla, Sujoy Basu, Abhi Mohan, Nick
  Fernando, Noah Fiedel, Kim Paterson, Hui Li, Ankush Garg, Jane Park, DongHyun Choi, Diane Wu, Sankalp Singh, Zhishuai Zhang, Amir Globerson, Lily Yu, John Carpenter, Félix de~Chaumont~Quitry, Carey Radebaugh, Chu-Cheng Lin, Alex Tudor, Prakash Shroff, Drew Garmon, Dayou Du, Neera Vats, Han Lu, Shariq Iqbal, Alex Yakubovich, Nilesh Tripuraneni, James Manyika, Haroon Qureshi, Nan Hua, Christel Ngani, Maria~Abi Raad, Hannah Forbes, Jeff Stanway, Mukund Sundararajan, Victor Ungureanu, Colton Bishop, Yunjie Li, Balaji Venkatraman, Bo~Li, Chloe Thornton, Salvatore Scellato, Nishesh Gupta, Yicheng Wang, Ian Tenney, Xihui Wu, Ashish Shenoy, Gabriel Carvajal, Diana~Gage Wright, Ben Bariach, Zhuyun Xiao, Peter Hawkins, Sid Dalmia, Clement Farabet, Pedro Valenzuela, Quan Yuan, Ananth Agarwal, Mia Chen, Wooyeol Kim, Brice Hulse, Nandita Dukkipati, Adam Paszke, Andrew Bolt, Kiam Choo, Jennifer Beattie, Jennifer Prendki, Harsha Vashisht, Rebeca Santamaria-Fernandez, Luis~C. Cobo, Jarek Wilkiewicz, David Madras, Ali
  Elqursh, Grant Uy, Kevin Ramirez, Matt Harvey, Tyler Liechty, Heiga Zen, Jeff Seibert, Clara~Huiyi Hu, Andrey Khorlin, Maigo Le, Asaf Aharoni, Megan Li, Lily Wang, Sandeep Kumar, Norman Casagrande, Jay Hoover, Dalia~El Badawy, David Soergel, Denis Vnukov, Matt Miecnikowski, Jiri Simsa, Praveen Kumar, Thibault Sellam, Daniel Vlasic, Samira Daruki, Nir Shabat, John Zhang, Guolong Su, Jiageng Zhang, Jeremiah Liu, Yi~Sun, Evan Palmer, Alireza Ghaffarkhah, Xi~Xiong, Victor Cotruta, Michael Fink, Lucas Dixon, Ashwin Sreevatsa, Adrian Goedeckemeyer, Alek Dimitriev, Mohsen Jafari, Remi Crocker, Nicholas FitzGerald, Aviral Kumar, Sanjay Ghemawat, Ivan Philips, Frederick Liu, Yannie Liang, Rachel Sterneck, Alena Repina, Marcus Wu, Laura Knight, Marin Georgiev, Hyo Lee, Harry Askham, Abhishek Chakladar, Annie Louis, Carl Crous, Hardie Cate, Dessie Petrova, Michael Quinn, Denese Owusu-Afriyie, Achintya Singhal, Nan Wei, Solomon Kim, Damien Vincent, Milad Nasr, Christopher~A. Choquette-Choo, Reiko Tojo, Shawn Lu, Diego
  de~Las~Casas, Yuchung Cheng, Tolga Bolukbasi, Katherine Lee, Saaber Fatehi, Rajagopal Ananthanarayanan, Miteyan Patel, Charbel Kaed, Jing Li, Shreyas~Rammohan Belle, Zhe Chen, Jaclyn Konzelmann, Siim Põder, Roopal Garg, Vinod Koverkathu, Adam Brown, Chris Dyer, Rosanne Liu, Azade Nova, Jun Xu, Alanna Walton, Alicia Parrish, Mark Epstein, Sara McCarthy, Slav Petrov, Demis Hassabis, Koray Kavukcuoglu, Jeffrey Dean, and Oriol Vinyals. 2024.
\newblock Gemini 1.5: Unlocking multimodal understanding across millions of tokens of context.
\newblock \emph{arXiv preprint arXiv: 2403.05530}.

\bibitem[{Wang et~al.(2023{\natexlab{a}})Wang, Liang, Meng, Sun, Shi, Li, Xu, Qu, and Zhou}]{gpt-good-evaluator}
Jiaan Wang, Yunlong Liang, Fandong Meng, Zengkui Sun, Haoxiang Shi, Zhixu Li, Jinan Xu, Jianfeng Qu, and Jie Zhou. 2023{\natexlab{a}}.
\newblock Is chatgpt a good nlg evaluator? a preliminary study.
\newblock \emph{arXiv preprint arXiv: 2303.04048}.

\bibitem[{Wang et~al.(2023{\natexlab{b}})Wang, Li, Chen, Cai, Zhu, Lin, Cao, Liu, Liu, and Sui}]{faireval}
Peiyi Wang, Lei Li, Liang Chen, Zefan Cai, Dawei Zhu, Binghuai Lin, Yunbo Cao, Qi~Liu, Tianyu Liu, and Zhifang Sui. 2023{\natexlab{b}}.
\newblock Large language models are not fair evaluators.
\newblock \emph{arXiv preprint arXiv: 2305.17926}.

\bibitem[{Wang et~al.(2023{\natexlab{c}})Wang, Li, Chen, Zhu, Lin, Cao, Liu, Liu, and Sui}]{llm-not-fair-eval}
Peiyi Wang, Lei Li, Liang Chen, Dawei Zhu, Binghuai Lin, Yunbo Cao, Qi~Liu, Tianyu Liu, and Zhifang Sui. 2023{\natexlab{c}}.
\newblock \href {https://doi.org/10.48550/ARXIV.2305.17926} {Large language models are not fair evaluators}.
\newblock \emph{CoRR}, abs/2305.17926.

\bibitem[{Wang et~al.(2023{\natexlab{d}})Wang, Yu, Zeng, Yang, Wang, Chen, Jiang, Xie, Wang, Xie, Ye, Zhang, and Zhang}]{PandaLM}
Yidong Wang, Zhuohao Yu, Zhengran Zeng, Linyi Yang, Cunxiang Wang, Hao Chen, Chaoya Jiang, Rui Xie, Jindong Wang, Xing Xie, Wei Ye, Shikun Zhang, and Yue Zhang. 2023{\natexlab{d}}.
\newblock \href {https://doi.org/10.48550/ARXIV.2306.05087} {Pandalm: An automatic evaluation benchmark for {LLM} instruction tuning optimization}.
\newblock \emph{CoRR}, abs/2306.05087.

\bibitem[{Watts et~al.(2024)Watts, Gumma, Yadavalli, Seshadri, Manohar, and Sitaram}]{watts2024pariksha}
Ishaan Watts, Varun Gumma, Aditya Yadavalli, Vivek Seshadri, Swami Manohar, and Sunayana Sitaram. 2024.
\newblock \href {https://www.microsoft.com/en-us/research/publication/pariksha-a-scalable-democratic-transparent-evaluation-platform-for-assessing-indic-large-language-models/} {Pariksha: A scalable, democratic, transparent evaluation platform for assessing indic large language models}.

\bibitem[{Wei et~al.(2022)Wei, Wang, Schuurmans, Bosma, hsin Chi, Xia, Le, and Zhou}]{Wei2022ChainOT}
Jason Wei, Xuezhi Wang, Dale Schuurmans, Maarten Bosma, Ed~Huai hsin Chi, F.~Xia, Quoc Le, and Denny Zhou. 2022.
\newblock \href {https://api.semanticscholar.org/CorpusID:246411621} {Chain of thought prompting elicits reasoning in large language models}.
\newblock \emph{ArXiv}, abs/2201.11903.

\bibitem[{Wu and Aji(2023)}]{wu2023style}
Minghao Wu and Alham~Fikri Aji. 2023.
\newblock Style over substance: Evaluation biases for large language models.
\newblock \emph{arXiv preprint arXiv: 2307.03025}.

\bibitem[{Wu et~al.(2023)Wu, Qiu, Ross, Akyürek, Chen, Wang, Kim, Andreas, and Kim}]{wu2023reasoning}
Zhaofeng Wu, Linlu Qiu, Alexis Ross, Ekin Akyürek, Boyuan Chen, Bailin Wang, Najoung Kim, Jacob Andreas, and Yoon Kim. 2023.
\newblock Reasoning or reciting? exploring the capabilities and limitations of language models through counterfactual tasks.
\newblock \emph{arXiv preprint arXiv: 2307.02477}.

\bibitem[{Xu et~al.(2023)Xu, Sun, Zheng, Geng, Zhao, Feng, Tao, and Jiang}]{wizardlm}
Can Xu, Qingfeng Sun, Kai Zheng, Xiubo Geng, Pu~Zhao, Jiazhan Feng, Chongyang Tao, and Daxin Jiang. 2023.
\newblock \href {https://doi.org/10.48550/ARXIV.2304.12244} {Wizardlm: Empowering large language models to follow complex instructions}.
\newblock \emph{CoRR}, abs/2304.12244.

\bibitem[{Ye et~al.(2023)Ye, Kim, Kim, Hwang, Kim, Jo, Thorne, Kim, and Seo}]{DBLP:journals/corr/abs-2307-10928}
Seonghyeon Ye, Doyoung Kim, Sungdong Kim, Hyeonbin Hwang, Seungone Kim, Yongrae Jo, James Thorne, Juho Kim, and Minjoon Seo. 2023.
\newblock \href {https://doi.org/10.48550/ARXIV.2307.10928} {{FLASK:} fine-grained language model evaluation based on alignment skill sets}.
\newblock \emph{CoRR}, abs/2307.10928.

\bibitem[{Zeng et~al.(2023)Zeng, Yu, Gao, Meng, Goyal, and Chen}]{llm-bar}
Zhiyuan Zeng, Jiatong Yu, Tianyu Gao, Yu~Meng, Tanya Goyal, and Danqi Chen. 2023.
\newblock \href {https://doi.org/10.48550/ARXIV.2310.07641} {Evaluating large language models at evaluating instruction following}.
\newblock \emph{CoRR}, abs/2310.07641.

\bibitem[{Zhang et~al.(2023)Zhang, Yu, Yu, Lv, Liu, Huang, Xu, and Li}]{wider-and-deeper}
Xinghua Zhang, Bowen Yu, Haiyang Yu, Yangyu Lv, Tingwen Liu, Fei Huang, Hongbo Xu, and Yongbin Li. 2023.
\newblock Wider and deeper llm networks are fairer llm evaluators.
\newblock \emph{arXiv preprint arXiv: 2308.01862}.

\bibitem[{Zheng et~al.(2023)Zheng, Chiang, Sheng, Zhuang, Wu, Zhuang, Lin, Li, Li, Xing, Zhang, Gonzalez, and Stoica}]{llm-judge}
Lianmin Zheng, Wei{-}Lin Chiang, Ying Sheng, Siyuan Zhuang, Zhanghao Wu, Yonghao Zhuang, Zi~Lin, Zhuohan Li, Dacheng Li, Eric~P. Xing, Hao Zhang, Joseph~E. Gonzalez, and Ion Stoica. 2023.
\newblock \href {http://papers.nips.cc/paper\_files/paper/2023/hash/91f18a1287b398d378ef22505bf41832-Abstract-Datasets\_and\_Benchmarks.html} {Judging llm-as-a-judge with mt-bench and chatbot arena}.
\newblock In \emph{Advances in Neural Information Processing Systems 36: Annual Conference on Neural Information Processing Systems 2023, NeurIPS 2023, New Orleans, LA, USA, December 10 - 16, 2023}.

\bibitem[{Zhou et~al.(2023{\natexlab{a}})Zhou, Liu, Xu, Iyer, Sun, Mao, Ma, Efrat, Yu, Yu, Zhang, Ghosh, Lewis, Zettlemoyer, and Levy}]{lima}
Chunting Zhou, Pengfei Liu, Puxin Xu, Srini Iyer, Jiao Sun, Yuning Mao, Xuezhe Ma, Avia Efrat, Ping Yu, Lili Yu, Susan Zhang, Gargi Ghosh, Mike Lewis, Luke Zettlemoyer, and Omer Levy. 2023{\natexlab{a}}.
\newblock \href {https://doi.org/10.48550/ARXIV.2305.11206} {{LIMA:} less is more for alignment}.
\newblock \emph{CoRR}, abs/2305.11206.

\bibitem[{Zhou et~al.(2023{\natexlab{b}})Zhou, Lu, Mishra, Brahma, Basu, Luan, Zhou, and Hou}]{ifeval}
Jeffrey Zhou, Tianjian Lu, Swaroop Mishra, Siddhartha Brahma, Sujoy Basu, Yi~Luan, Denny Zhou, and Le~Hou. 2023{\natexlab{b}}.
\newblock \href {https://doi.org/10.48550/ARXIV.2311.07911} {Instruction-following evaluation for large language models}.
\newblock \emph{CoRR}, abs/2311.07911.

\bibitem[{Zhu et~al.(2023)Zhu, Wang, and Wang}]{zhu2023judgelm}
Lianghui Zhu, Xinggang Wang, and Xinlong Wang. 2023.
\newblock Judgelm: Fine-tuned large language models are scalable judges.
\newblock \emph{arXiv preprint arXiv: 2310.17631}.

\end{thebibliography}
